\documentclass[11pt]{article}

\usepackage[draft]{fixme}
\fxusetheme{color}
\fxsetup{targetlayout=color}

\usepackage{apacite}
\usepackage[utf8]{inputenc}
\usepackage{a4wide, amsmath, amssymb, amsthm}
\usepackage[ruled, commentsnumbered, linesnumbered, vlined]{algorithm2e}
\usepackage{enumerate}
\usepackage{latexsym, epsfig, lscape, graphicx}
\usepackage{scalefnt}
\usepackage[normalem]{ulem}
\usepackage[title]{appendix}
\usepackage{fullpage}
\usepackage[dvipsnames]{xcolor}
\newcommand\Tstrut{\rule{0pt}{2.2ex}}         % = `top' strut
\newcommand\Bstrut{\rule[-1.5ex]{0pt}{0pt}}   % = `bottom' strut

\usepackage{booktabs, multirow} % for borders and merged ranges
\usepackage{changepage,threeparttable} % for wide tables
\usepackage[inline]{enumitem}
\usepackage{caption}
\usepackage{subcaption}
\usepackage{microtype}
\usepackage{float}
\usepackage{ulem}
\newcommand{\floor}[1]{\left\lfloor#1\right\rfloor}

\SetKw{KwAnd}{and}
\SetKw{KwReach}{not reached}

\newcommand{\rafaelC}[1]{\textcolor{black}{#1}}

\begin{document}

\title{An enhanced simulation-based iterated local search metaheuristic for gravity fed water distribution network design optimization}

\author{
Willian C. S. Martinho {\thanks{Universidade Federal da Bahia, Departamento de Ci\^{e}ncia da Computa\c{c}\~{a}o, Computational Intelligence and Optimization Research Lab (CInO), Salvador, Brazil.  ({\tt willianmartinho@gmail.com}). }}
\and 
Rafael A. Melo {\thanks{Universidade Federal da Bahia, Departamento de Ci\^{e}ncia da Computa\c{c}\~{a}o, Computational Intelligence and Optimization Research Lab (CInO), Salvador, Brazil.  ({\tt melo@dcc.ufba.br}).}}
\and 
Kenneth S\"orensen{\thanks{University of Antwerp, Department of Engineering Management, ANT/OR - Operations Research Group, Belgium. ({\tt kenneth.sorensen@uantwerpen.be}).}}
}

\maketitle

\begin{abstract}
The gravity fed water distribution network design (WDND) optimization problem consists in determining the pipe diameters of a water network such that hydraulic constraints are satisfied and the total cost is minimized. Traditionally, such design decisions are made on the basis of expert experience. When networks increase in size, however, rules of thumb will rarely lead to near optimal decisions.
Over the past thirty years, a large number of techniques have been developed to tackle the problem of optimally designing a water distribution network.
In this paper, we tackle the NP-hard WDND optimization problem in a multi-period setting where time varying demand patterns occur. We propose a new enhanced simulation-based iterated local search (ILS) metaheuristic which further explores the structure of the problem in an attempt to obtain high quality solutions.
More specifically, four novelties are proposed: (a) a local search strategy to smartly dimension pipes in the shortest paths between the reservoirs and the nodes with highest demands; (b) a technique to speed up convergence based on an aggressive pipe diameter reduction scheme; (c) a novel concentrated perturbation mechanism to allow escaping from very restrained local optima solutions; and (d) a pool of solutions to achieve a good compromise between intensification and diversification.
Computational experiments show that our approach is able to improve over a state-of-the-art metaheuristic for most of the performed tests. Furthermore, it converges much faster to low cost solutions and demonstrates a more robust performance in that it obtains smaller deviations from the best known solutions.
\newline 

\end{abstract}

\section{Introduction}
\label{sec:intro}

A safe and adequate access to drinking water is one of the basic necessities of human beings.
The most efficient and effective way to transport drinking water is through a network of pipelines.
Consequently, water distribution networks are amongst the most vital elements of a society's infrastructure, providing people with high-quality drinking water in sufficient quantities and at adequate pressures.
A water distribution system is an infrastructure that consists of different elements (such as pumps, reservoirs, pipes, valves, among others) connected together to convey quantities of water from one or more sources to multiple consumers (which can be domestic, commercial, or industrial).
The distribution system is designed to reliably distribute water in sufficient quantities and provide to demand patterns, pressure and velocity limitations, quality assurance, and maintenance issues.
Therefore, different combinations of reservoir storage, network layout, water mains, and pumps are used, depending on the system service area, topography, and size. 
Very often, the reliability of these networks requires major investments and thus an efficient network design is of crucial importance.

Water distribution networks require decisions in three different phases with distinct time horizons, namely, layout, design, and planning~\cite{DeCSor16a}.
The \emph{layout} phase consists of strategic decisions based on which the structure of the network is defined. 
In this phase, decisions taken include where the different pipes will be constructed as well as the locations to build pumps, valves, water tanks, reservoirs, and other components of the network.
Traditionally, the pipe network layout is designed by the engineers and taken as fixed during the process of component sizing.
The \emph{design} phase also consists of strategical decisions which will define the type and size of each pipe, pump, valve, tank, and all other components of the network.
All these decisions take into consideration the water demands, the adequacy of water facilities, and the water storage necessary to meet peak demands. Furthermore, other reliability conditions of the network should be met with provision for the estimated requirements in the future.
The \emph{planning} phase groups all tactical decisions which are taken on a daily basis and are concerned with the functioning of valves and operational pump levels to ensure that sufficient water is available in all nodes of the network.

In this paper, we study the gravity fed water distribution network design (WDND) optimization problem in a multi-period setting in which time varying demand patterns occur (each time period representing a scenario to be satisfied). 
Informally, this problem can be defined as follows. Given a network topology (i.e., a set of water supply points, a set of demand points, and a set of junction points, together with a set of potential pipes connecting them), associated demand patterns, a collection of hydraulic constraints, and a set of available pipe types, each of which with a cost per length, the problem consists in assigning a particular type to each pipe in the network such that all requirements are met and the total design cost is minimized. Note that pumps are not considered in these networks as it is assumed that the pressure suffices to supply all the demands. In terms of the different phases mentioned before, the WDND can be thought of as combining the \emph{design} phase (determining the type of pipe for each potential pipeline) with aspects of the \emph{layout} phase (determinining whether a potential pipeline should be built or not).
The WDND optimization problem is challenging for several reasons. Firstly, it is known to be NP-hard \cite{YatTemBof84}. Secondly, the hydraulic conditions of the network are affected by every pipe and therefore changes in one of them can influence the circumstances as a whole. 
For these reasons, this work proposes a robust simulation-based iterated local search metaheuristic.

% Linear Programming
Originally, the proposed methods dealt with water distribution system management problems using linear/nonlinear optimization techniques. These methods were limited by the system size, the number of constraints, and the number of loading conditions.
\citeA{AlpSha77} proposed a linear programming (LP) based approach, denoted linear programming gradient (LPG), in which the pipes were considered to be composed of segments of different sizes and the sum of these segments, each one with different diameters, was equal to the length of the pipe. 
The method decomposes the optimization problem in two levels. In the first level, given a set of flows, the corresponding optimal cost of the network is obtained using linear programming. 
At the second level, this solution is used to modify the flow of the pipes according to the gradient of the objective function with the aim of improving the cost. 
The authors proposed to repeat the procedure using different starting points to increase the probability of finding a global optimum.
\citeA{QuiJonDow81} extended the LPG of~\citeA{AlpSha77}, but instead of the flows they considered the head (internal energy per unit weight of fluid) of the nodes as values in the formulation of the linear program. 
The authors also adjusted the original expression of the gradient derivation to express the interaction between paths, which enhanced the performance of the method.
\citeA{FujJenEdi87} suggested an improvement of the original LPG in which the adjustments in the flows of the pipes are performed by a quasi-Newton method rather than a simple gradient search heuristic.
A concise and complete study of the LPG has been adopted and restated in \citeA{KesShaUri89}, where the authors successfully improved the method.

% Nonlinear Programming
\citeA{Sha74} proposed a nonlinear programming (NLP) approach that uses a combination of generalized reduced gradient and penalty methods. 
The approach constructs a Lagrangian function by penalizing the objective function for violation of the constraints. 
The flow values are obtained by a Newton-Raphson method employing sparse matrix techniques.
\citeA{FujKha90} proposed a two-phase decomposition method called nonlinear programming gradient (NLPG), which extends the method of \citeA{AlpSha77} to nonlinear modeling. 
The method first specifies the flow distribution and pumping heads. Then, it solves the resulting convex program to get the pipes' head losses. 
The Lagrange multipliers of the obtained optimal solution are used to modify the flow distribution and pumping heads to achieve a reduction in the cost of the system.
In the second phase the obtained pipes' head losses are fixed, and the resulting concave program is solved for the flow distribution and pumping heads. These two phases are repeated until no further improvement can be achieved.
Recently, \citeA{CabRav19} proposed a mixed integer non-linear programming (MINLP) model considering the flow directions and the pipes diameters as optimization variables. The authors used techniques to reduce the number of nonlinear equations in the model, which was solved using a global optimization solver. The approach was used to solve to optimality network instances with up to 34 pipes.

% Local search metaheuristics Methods
Different metaheuristic frameworks have been applied to deal with the WDND optimization problem. 
\citeA{LogGreAhn95} proposed an outer flow search-inner optimization procedure.
In their approach, each pipe in the network is subject to an external search scheme based on simulated annealing (SA) and a multi-start local search that selects alternative flow configurations. 
Next, an inner linear program is used for the design of least-cost diameters through selected flow configurations.
\citeA{CunSou99} also proposed an SA based metaheuristic, in which the neighborhood of a solution is any configuration having all the pipes but one with the same diameter as in the current configuration. 
Starting with an initial configuration of pipes with their respective diameters, at each step of the algorithm, a new configuration is randomly selected from the neighborhood of the current configuration, and then its cost is evaluated. 
If it is accepted, the configuration will be used as the starting point for the next step. 
If not, the original configuration will play that role. 
Each generated configuration is evaluated using the Newton-Raphson method to solve the hydraulic equilibrium equation set and determine the flows and hydraulic heads of the system.

Evolutionary algorithms represent an alternative strategy for tackling complex problems. 
\citeA{SimDanMur94} proposed an approach using a genetic algorithm (GA) that uses a binary encoding in which each tube diameter is assigned to a binary code. 
A Newton-Raphson method was used for hydraulic network analysis at each function evaluation to determine the flows and hydraulic heads of the system.
The method assigns a penalty cost for every constraint violation, which is added to the network's total cost.
The algorithm proceeds until a given total number of generations is evaluated. Results were compared with the techniques of complete enumeration and nonlinear programming.
Simulation-based heuristics are presented as an alternative for problems that have difficult solution equations. 
In this way, it is possible to work on improving heuristic methods for the problem, without spending too much effort on solving the equations. \citeA{SavWal97} used a genetic algorithm which adopts a Gray coding instead of the traditionally used binary coding.
The hydraulic simulator EPANET~\cite{Ros93} was used to solve the equations of the system and determine the flows and hydraulic heads.
\citeA{Guptal99} developed an alternative GA implementation which represents the set of solutions as the discrete pipe sizes and not in the binary alphabet as usual. 
Therefore, the decoding or coding required during the execution of the algorithm for each new set of solutions is avoided. 
The hydraulic simulator ANALIS~\cite{BasGup92} was used to compute the hydraulic equations.
\citeA{CheJiaZhaLuoJiaZha19} presented a cooperative co-evolutionary algorithm (CCEA)
using the hydraulic simulator EPANET. The authors proposed an iterative trace-based decomposition method to divide a large-scale water distribution network into sub-networks. After decomposition, the sub-components are handled by an equal number of cooperative evolutionary algorithms. 

\citeA{Maital03} proposed a simulation based ant colony optimization (ACO) approach in which every decision point is associated with a pipe in the network. At each decision point, there are a number of options, corresponding to the available pipe diameters. The cost corresponding to a particular option is the product of the unit cost per meter length of each of the pipe diameters and the length of the pipe segment under consideration. The algorithm was linked with the hydraulic solver WADISO~\cite{GesWal85} in order to calculate the maximum pressure deficit for each generated candidate solution. \citeA{Zecetal05} presented a similar strategy but additionally conducted an extensive study to determine the guidelines for assigning values to the various parameters in the ACO algorithm specifically for WDND optimization. 
Other approaches such as particle swarm optimization (PSO)~\cite{MonIzq10} and harmony search (HS)~\cite{Geem06} were also applied in the context of WDND optimization.

\citeA{DeCSor16a} presented an approach using iterated local search (ILS) to handle the WDND optimization problem. The algorithm constructs an initial solution and then attempts to iteratively improve this solution by performing small diameter reduction moves. 
Therefore, neighboring solutions have configurations in which all pipes but one have the same diameter as the current solution. 
The decision of which pipe will be decreased in diameter depends on a greedy randomized strategy. 
When a local optimum is reached, a perturbation move is used to try escaping from this local optimum. 
The hydraulic simulator EPANET 2~\cite{Ros00} was used to evaluate the problem constraints for each obtained candidate solution. Later, \citeA{DeCSor16b} extended the problem to a multi-period setting in which time-varying demand patterns occur, which is the variation we consider in our work.

A detailed critical review on gravity-fed water distribution network design optimization is given in \citeA{DeCSor13}.
\citeA{DamLodWieBra15} presented a survey of mathematical programming approaches widely applied to mixed integer non-linear programming models and introduced them to the context of water distribution network optimization. The authors discussed a class of challenging optimization problems related to the nonlinear network flow model and then attempted to solve them using the described techniques.
More recently, \citeA{MalSulSav18} presented a systematic review and discussed limits, trends, and future research directions in the field of water distribution network design optimization. 

\subsection{Main contribution and organization}

The main contribution of our work is a robust simulation-based iterated local search metaheuristic to tackle the gravity fed water distribution network design optimization problem focused on a multi-period setting with time-varying demand patterns. The approach improves over the metaheuristic of \citeA{DeCSor16b} by considering relevant concepts on the structure of the problem to allow enhancements in performance. The approach presents four main novelties.
Firstly, observing that low cost water distribution networks tend to have in common a structure pattern in which larger diameter pipes create paths between the nodes with highest demands and the source nodes, we attempt to smartly dimension the pipes in these paths.
Secondly, it uses an aggressive pipe diameter reduction scheme based on different factors to speed up convergence. Thirdly, it introduces a concentrated perturbation mechanism to allow escaping from very restrained local optima solutions. Fourtly, it encompasses a pool of solutions to achieve %attempt achieving 
a good compromise between intensification and diversification.
Extensive computational experiments carried out using a set of benchmark instances show that the introduced novelties allow our new approach to outperform a state-of-the-art metaheuristic for most of the instances. Furthermore, our new approach presents much faster convergence to good quality solutions and more robustness, as the obtained solutions show much less deviation from the best ones.

The remainder of the paper is organized as follows. Section~\ref{sec:wdnd} formalizes the water distribution network design optimization problem in a multi-period setting with time-varying demand patterns. Section~\ref{sec:approaches} describes the proposed simulation-based iterated local search metaheuristic. Section~\ref{sec:experiments} summarizes the computational experiments marking the improvements achieved with the newly proposed approach.
Section~\ref{sec:final} discusses final comments.

\section{The water distribution network  design optimization problem}
\label{sec:wdnd}

\begin{figure}[H]
     \centering
     \begin{subfigure}[b]{0.6\textwidth}
        \centering
         \includegraphics[width=\textwidth]{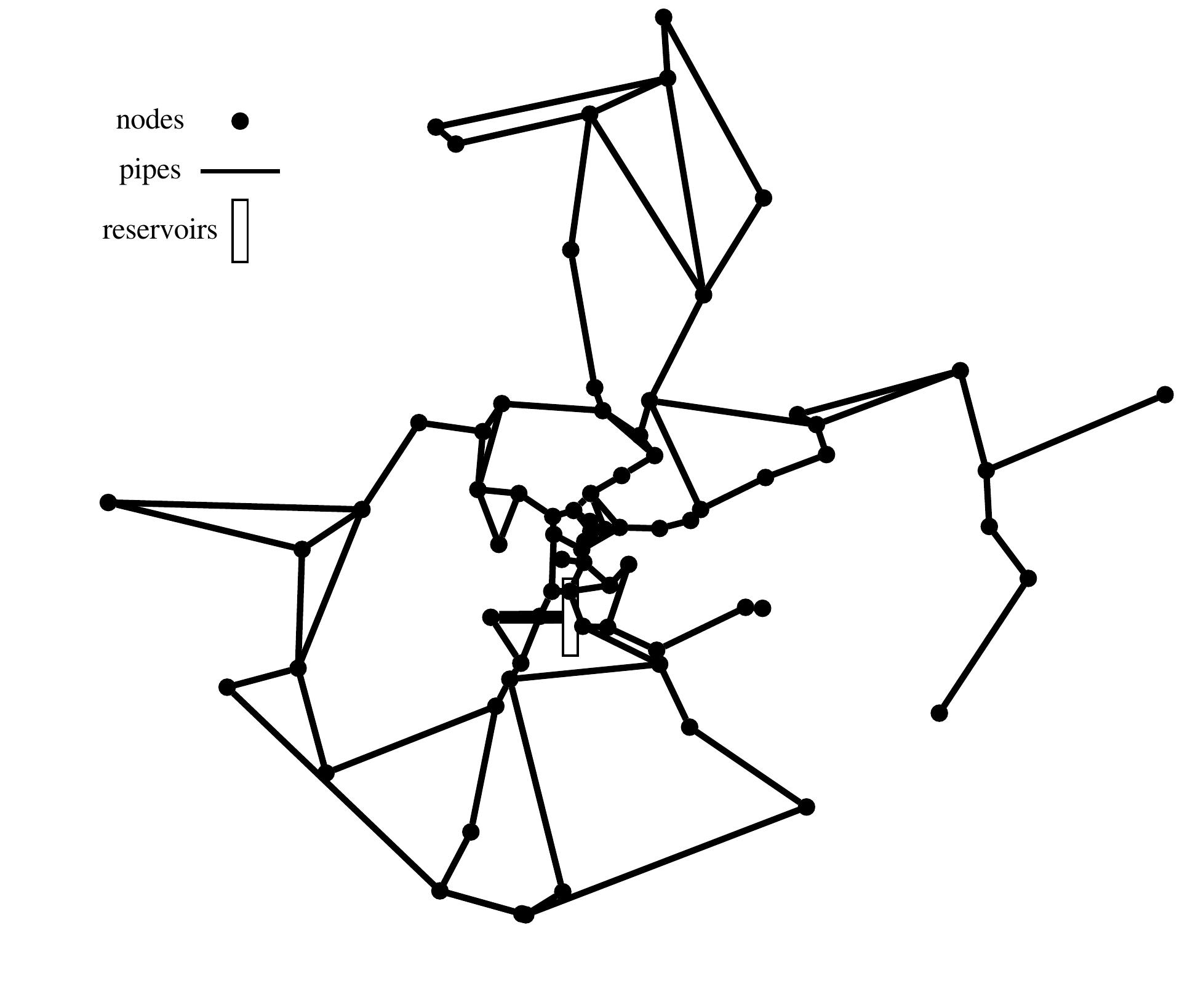}
         \caption{}
         \label{fig:netOpt}
     \end{subfigure}
     %\hfill
     \hspace{1cm}
     \begin{subfigure}[b]{0.6\textwidth}
         \centering
         \includegraphics[width=\textwidth]{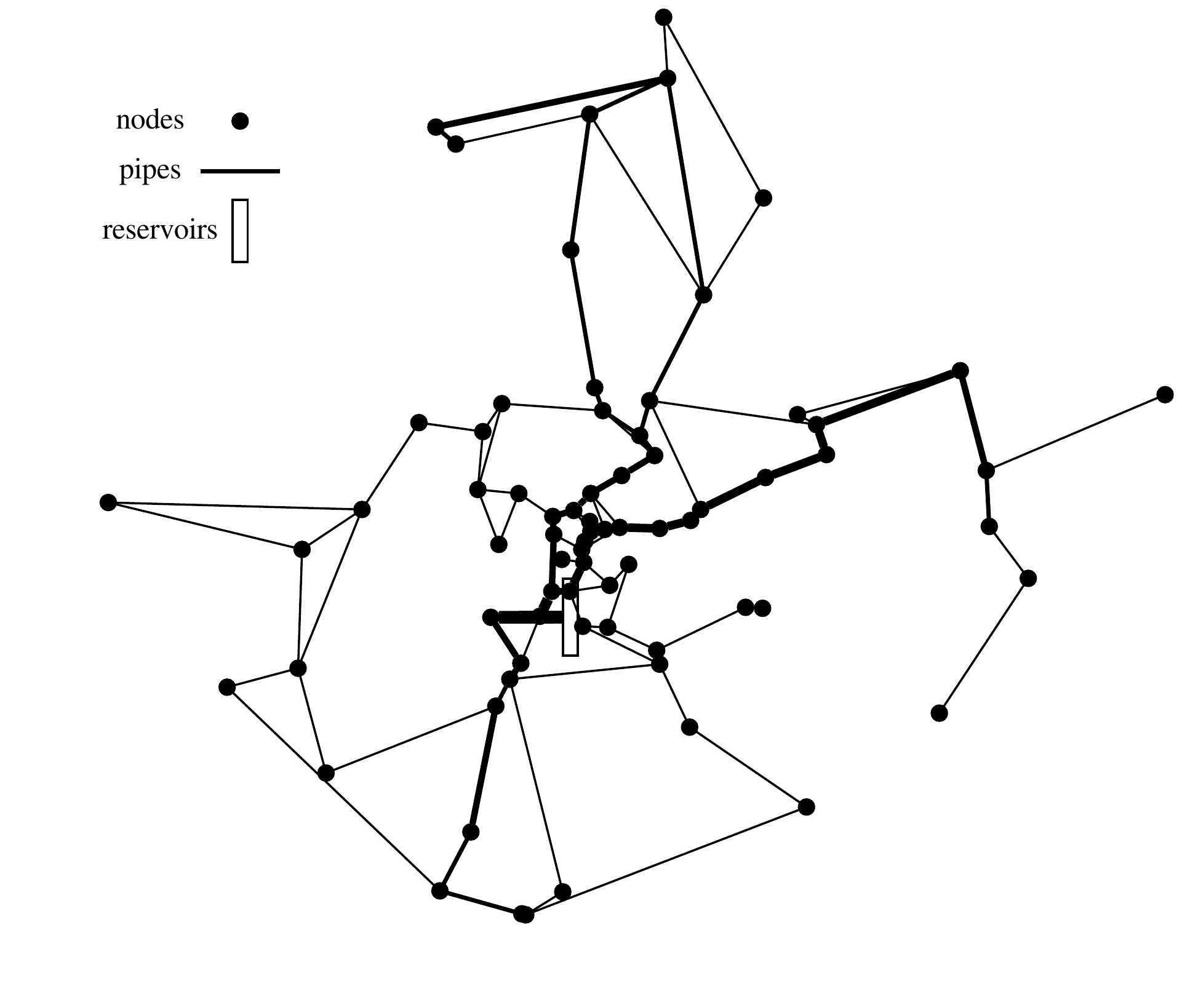}
         \caption{}
         \label{fig:net!Opt}
     \end{subfigure}
        \caption{Example of a water distribution network. Demand nodes are represented as dots, reservoirs as rectangles and pipes as lines. The thicker lines represent larger diameter pipes. A network which is not optimized is exemplified in (a), whereas an optimized network is illustrated in (b).}
        \label{fig:network}
\end{figure}

Consider a set of nodes $N = \{1, 2, \ldots, |N|\}$ representing points of water demand and supply (such as a reservoir) as well as junctions (points with both demand and supply equal to zero), and a set of water distribution pipes $P = \{1, 2, \ldots, |P|\}$. 
Note that a water distribution network can be represented as a graph $G = (V, E)$ in which there is a vertex for each node and an edge for each pipe of the network.
A water distribution network is illustrated in Figure~\ref{fig:network}.

Define a planning horizon $\mathcal{T} = \{1, 2, \ldots, |\mathcal{T}|\}$ which describes a typical network operation cycle, with each $\tau \in \mathcal{T}$ representing a different scenario.
Let $T = \{1, 2, ..., |T|\}$ be the set of available pipe types in which each type $t \in T$ has an associated cost $c_t \in \mathbb{R_+}$ per unit of length, roughness $r_t \in \mathbb{R_+}$ and diameter $d_t \in \mathbb{R_+}$ (in $mm$). In this work, it is assumed that the costs of the types increase as their diameters increase. It is also assumed that the elements in $T$ are organized such that $d_1 \leq d_2 \leq \ldots \leq d_{|T|}$. In addition, consider $l_p \in \mathbb{R_+}$ (in $m$) to be the length of pipe $p \in P$.
Consider, for every node $n \in N$, the expected water demand $D_{n, \tau} \in \mathbb{R_+}$ (in $m^3/s$) and the water supply $R_{n, \tau} \in \mathbb{R_+}$ (in $m^3/s$) that has to be satisfied for every period $\tau \in \mathcal{T}$.
Consider $h^{min} \in \mathbb{R_+}$ (in $m$) to be the minimum pressure assumed for every node $n \in N$ and $v^{max} \in \mathbb{R_+}$ (in $m/s$) to be the maximum velocity for each pipe $p \in P$ to be satisfied for every period $\tau \in \mathcal{T}$.
The WDND optimization problem consists in the selection of a type in $T = \{1, 2, \ldots, |T|\}$ for each pipe in the network so that the total cost of the network is minimized without violating any hydraulic constraints. A summary of the input data and decision variables is given in Table~\ref{tab:symbols}.

\begin{table}[H]
\small
    \caption{Summary of used symbols.}
    \centering
    \begin{tabular}{cp{10cm}} 
        \hline
	\textbf{Symbols} & \textbf{Descriptions} \\
	\hline\hline
	$N$ & Set of nodes \\
	$P$ & Set of pipes \\
	$\mathcal{T}$ & Set representing the planning horizon \\
    $T$ & Set of available pipe types \\
    $\Xi$ & Set of closed loops in the network \\
    $\xi$ & Set of pipes in a closed loop \\
    $c_t $ & Cost of pipe type $t \in T$, per unit of length \\
    $r_t $ & Roughness of pipe type $t \in T$ \\
    $d_t $ & Diameter of pipe type $t \in T$ (in $mm$) \\
    $l_p $ & Length of pipe $p \in P$ (in $m$) \\
    $D_{n, \tau}$ & Water demand (in $m^3/s$) of node $n \in N$ in period $\tau \in \mathcal{T}$ \\
    $R_{n, \tau}$ & Water supply (in $m^3/s$)  of node $n \in N$ in period $\tau \in \mathcal{T}$ \\
    $h^{min} $ & Minimum pressure (in $m$) allowed in node $n \in N$ \\
    $v^{max} $ & Maximum velocity (in $m/s$) allowed in pipe $p \in P$   \\ \hline
    $h_{n, \tau} $ & Variable indicating the pressure (in $m$) in node $n\in N$ in period $\tau \in \mathcal{T}$ \\
    $v_{p, \tau} $ & Variable indicating the velocity (in $m/s$) of the water flowing through pipe $p \in P$ in period $\tau \in \mathcal{T}$ \\
    $x_{p,t}$ & Binary variable indicating whether pipe $p\in P$ is of type $t\in T$ \\
    $Q_{(i,j),\tau}$ & Variable indicating the water flow from node $i\in N$ to node $j\in N$ in period $\tau \in \mathcal{T}$ (nonnegative)\\
    $Q_{p,\tau}$ & Variable indicating the water flow through pipe $p \in \mathcal{P}$ in period $\tau \in \mathcal{T}$ (nonnegative) \\
    $y_{p,\tau}$ & Variable representing the sign that incorporates changes in the direction of the water flow in pipe $p \in P$ in period $\tau \in \mathcal{T}$ \\
    $\Delta H_{p,\tau}$ & Variable indicating the head loss (in $m$) of pipe $p\in P$, connecting nodes $n_1$ and $n_2$, in period $\tau \in \mathcal{T}$ \\
    \hline
    \end{tabular}
    \label{tab:symbols}
\end{table}

Let $x_{p,t}$ be a binary decision variable that determines whether pipe $p$ is of type $t$ ($x_{p,t} = 1$) or not ($x_{p,t} = 0$). The objective function of the WDND optimization problem can be written as
\begin{equation}\label{eq:objfunc}
   z = \min \sum_{p \in P}\sum_{t\in T} c_t\cdot l_p\cdot x_{p,t}.
\end{equation}
A solution minimizing the objective function is conditioned to the principles of conservation of mass and energy, along with the constraints of minimum pressure head on the nodes and maximum water velocity in the pipes, for each period $\tau \in \mathcal{T}$.

Conservation of mass states that the volume of water flowing into a node in the network must be equal to the volume of water flowing out of this node. 
Consider variable $Q_{(i,j),\tau} \in \mathbb{R}_+$ to represent the water flowing from node $i$ to node $j$ in period $\tau$. The law of conservation of mass can be represented as
\begin{equation} \label{eq:mass}
    \sum_{i \in N \setminus \{n\}} Q_{(i, n),\tau} - \sum_{j \in N \setminus \{n\}} Q_{(n, j),\tau} = D_{n,\tau} - R_{n,\tau}, \ \ \ {\rm \text{for} \ } n \in N , \ \tau \in \mathcal{T}.
\end{equation}

The head loss in a piping system can be generalized as the energy used in overcoming friction caused by the walls of the pipe.
Let the set $\Xi = \{\xi_1, \xi_2, ..., \xi_{|\Xi|}\}$ represent the closed loops in the network, where each element $\xi \in \Xi$ represents a set of pipes belonging to that loop.
Therefore, the energy conservation law states that the sum of head losses for every closed loop $\xi \in \Xi$ in the network must be equal to zero.
Let variable $h_{n, \tau} \in \mathbb{R_+}$ (in $m$) be the pressure in node $n$ at period $\tau$ and let $\Delta H_{p,\tau} = h_{n{_1}, \tau} - h_{n_{2}, \tau}$ (in $m$) represent the head loss of a pipe $p$ in period $\tau$ that connects $n_1$ and $n_2$ with the direction defined by a closed loop that the nodes belong. The energy conservation law can be asserted as
\begin{equation} \label{eq:energy}
    \sum_{p \in \xi}\Delta H_{p,\tau} = 0, \ \ \ {\rm for \ } \xi \in \Xi, \ \tau \in \mathcal{T}.
\end{equation}
The head losses in the pipes of a network are approximated using the \rafaelC{empirical} Hazen-Williams equation. 
%%%%%%%%%%%%%%%%%%%%%%%%%%%%%%%%%%%%%%%%%%%%%
That equation, for a pipe $p$ at period $\tau$, with the parameters set as \rafaelC{predefined by} the hydraulic solver EPANET 2 \cite{Ros00} is described as  
\begin{equation} \label{eq:hw}
    \Delta H_{p,\tau} = \frac{10.6744 \cdot y_{p,\tau}\cdot Q_{p,\tau}^{1.852}\cdot l_p}{\sum_{t\in T}(x_{p,t}\cdot r_t^{1.852}\cdot d_t^{4.871})}.
\end{equation}
Therefore, Equation (\ref{eq:energy}) can be rewritten as
\begin{equation}
    \sum_{p \in \xi} \frac{10.6744 \cdot y_{p,\tau} \cdot Q_{p,\tau}^{1.852} \cdot l_p}{\sum_{t\in T}(x_{p,t}\cdot r_t^{1.852}\cdot d_t^{4.871})}  = 0, \ \ \ {\rm for \ } \xi \in \Xi, \ \tau \in \mathcal{T}.
\end{equation}
This work considers the numerical conversion constant (which depends on the units used) as $10.6744$ in Equation~\eqref{eq:hw}, as used by the simulator EPANET 2. 
Other values, such as $10.6750$, are also possible but it is known that small variations in this parameter can lead to different results \cite{SavWal97, PerAriOst09}. Besides, the two other constants are intrinsic to the formula and related to how it was obtained. The constant exponent 1.852 used for $Q_{p,\tau}$ and $r_t$ is a unitless coefficient originated from the fraction $\frac{1}{0.54}$, while the constant exponent $4.871$ used for $d_t$ comes from the fraction $\frac{2.63}{0.54}$. Small variations in these last constants are possible but they are just numerical, related to the number of considered decimal places.

Remark that, $Q_{p, \tau}$ represents an alternative formulation of $Q_{(i,j), \tau}$ if the pipe $p$ connects the nodes $i$ and $j$. However, that notation requires the definition of a variable $y_{p,\tau}$, which is the sign that incorporates changes in the direction of the water flow of pipe $p$ relative to the defined flow directions.

The minimum pressure constraint is described as
\begin{equation} \label{eq:pressure}
    h_{n,\tau} \geq h^{min}, \ \ \ {\rm for \ } n \in N, \ \tau\in \mathcal{T}.
\end{equation}

Assume variable $v_{p, \tau} \in \mathbb{R_+}$ (in $m/s$) to be the velocity of the water flowing through pipe $p$ in time period $\tau$, with  
\begin{equation}
    v_{p,\tau} = \frac{1.27 \cdot Q_{p,\tau}}{\sum_{t\in T}(x_{p,t}\cdot d_t^2)}, \ \ \ {\rm for \ } p \in P, \ \tau\in \mathcal{T}.
\end{equation}
The maximum water velocity constraint is thus represented by
\begin{equation} \label{eq:velocity}
    v_{p,\tau} \leq v^{max}, \ \ \ {\rm for \ } p \in P, \ \tau\in \mathcal{T}.
\end{equation}

Finally, every pipe $p$ in the network should get exactly one type $t$ assigned, which is ensured by
\begin{equation}
    \sum_{t\in T}x_{p,t} = 1, \ \ \ {\rm for \ } p \in P.  
\end{equation}
The integrality  requirements  of  the variables are defined as
\begin{align}
    \ \ & x_{p,t} \in \{0,1\},  \ \ \ {\rm for \ } p \in P, \ t \in T, \\
    \ \ & y_{p,\tau} \in \{-1,1\},  \ \ \ {\rm for \ } p \in P, \ \tau \in \mathcal{T}.
\end{align}

\section{The simulation-based iterated local search metaheuristic}
\label{sec:approaches}

As mentioned in the introduction, the optimization of water distribution networks has received considerable attention from the research community. 
Classical exact methods often use complex formulations and can be time-consuming for solving large real-world networks, as can be observed in \citeA{DamLodWieBra15}.
Metaheuristic approaches have successfully produced good-quality solutions in reasonable computational time, but most of them are black-box procedures that seldom use problem-specific structure information \cite{DeCSor13}. 

In this section, we describe a new robust simulation-based metaheuristic for the WDND optimization problem that overcomes some of the drawbacks of existing methods, both exact and heuristic. 
Similary to \citeA{DeCSor16b}, the metaheuristic combines an iterated local search (ILS) approach with the hydraulic simulator EPANET~2, but improves the former in several ways by taking into consideration the structure of the problem.

Firstly, we observed that low cost water distribution networks seem to have in common a structure pattern in which larger diameter pipes create paths between the nodes with highest demands and the source nodes. Hence, we attempt to dimension the pipes in some potential paths in a smart way. In this direction, we take into consideration the pipes in the shortest paths between nodes with high demands and the source nodes in the corresponding graph and postpone their reduction. 
This strategy can potentially lead to solutions with a reduced amount (in meters) of more expensive pipes with large diameters from which ramifications with cheaper pipes having small diameters would be originated.
Secondly, we apply an aggressive reduction in the pipes' diameters, attempting to reduce the number of simulations necessary to converge to low cost solutions. 
Thirdly, changing a pipe diameter usually becomes harder as its neighbors reach lower diameters, leading to hard to leave local optima regions. We attempt to overcome this issue by creating a perturbation which is concentrated on the neighbors of a given pipe, allowing disturbances which can permit alternative decreases in the diameters of the involved pipes in later iterations. 
Last but not least, as the metaheuristic looks for fast convergence, we attempt to diversify its search using a pool of solutions. In this way, it may perform local searches in the neighborhood of different solutions, potentially allowing to encounter better solutions.

For simplicity, we assume that the water network with its sets of nodes $N$ and pipes $P$ is represented as a graph $G = (V, E)$, with $V=N$ and $E=P$. Note that this is a slight abuse of notation, as we do not explicitly define the sets of vertices and edges, but rather perform a direct mapping from nodes and pipes of the network to the vertices and edges, correspondingly.
We define a solution $S$ of the problem to be an association of a type $t \in T$ to each pipe $p \in P$ of the network and $S(p)$ to be the type attributed to pipe $p$ in solution $S$. 

The remainder of this section is organized as follows. The procedure to perform the hydraulic simulations and evaluate every candidate solution is presented in Subsection \ref{sec:simulation}. The heuristic to generate an initial solution is described in Subsection \ref{sec:constructive}. The local search procedure is described in Subsection \ref{sec:localsearch}. Subsection \ref{sec:perturbation} describes the perturbation procedures to try escaping from local optima.
The criterion to accept candidates solutions is defined in Subsection \ref{sec:acceptance}. 
The complete simulation-based iterated local search metaheuristic is described in Subsection \ref{sec:ils}.

%%%%%%%%%%%%%%%%%%%%%%%%%%%%%%%%%%%%%%%%%%%%%%%%%
%%%%%%%%%%%%%%%%%%%%%%%%%%%%%%%%%%%%%%%%%%%%%%%%%
%%%%%%%%%%%%%%%%%%%%%%%%%%%%%%%%%%%%%%%%%%%%%%%%%

\subsection{Solution validation using hydraulic simulation} \label{sec:simulation}

As in \citeA{DeCSor16b}, our algorithm uses the hydraulic simulator EPANET~2 \cite{Ros00}, which performs simulation of hydraulic and water quality behavior within pressurized pipe networks. 
\rafaelC{EPANET 2's simulation consists of solving a set of hydraulic equations to determine feasible flows, the hydraulic pressure, and the velocity in each pipe and junction. To this end, it uses a heuristic procedure based on the global gradient algorithm (GGA)~\cite{TodPil87}, which generally yields near-optimal solutions.}
%EPANET 2's implementations are based on the global gradient algorithm (GGA)~\cite{TodPil87}, and the simulations consist of solving the hydraulic equations that rule the WDND optimization problem.
%\rafaelC{As in \citeA{DeCSor16b}, our algorithm uses the hydraulic simulator EPANET~2 \cite{Ros00}, which performs simulation of hydraulic and water quality behavior within pressurized pipe networks. EPANET 2's implementations are based on the global gradient algorithm (GGA)~\cite{TodPil87}, which that linearizes the energy equations. and the simulations consist of solving the hydraulic equations that rule the WDND optimization problem.}
The use of EPANET 2 allows us to concentrate our efforts on the search mechanism for good quality solutions without having to deal with the specific details of the feasibility checking of each individual solution.
The validation of a solution succeeds when all its nodes and pipes comply with the problem %'s 
constraints. Therefore, this method first uses the hydraulic simulator EPANET 2 and afterwards checks the outcome, i.e., the hydraulic parameters in  
each node and pipe in the solution, to verify whether a constraint violation occurs.

The \textsc{Hydraulic-Simulation} procedure takes as input the water distribution network $G$, the demands $D$, the supplies $R$, the planning horizon $\mathcal{T}$, the minimum pressure $h^{min}$ allowed at each node, the maximum velocity $v^{max}$ permitted at each pipe, and a candidate solution $S$ which is not necessarily feasible. The pseudocode of the procedure is described in Algorithm \ref{algo:simu}.

\begin{algorithm}[H]
\label{algo:simu}
\caption {\textsc{Solution-Validation}($G$, $D$, $R$, $\mathcal{T}$, $h^{min}$, $v^{max}$, $S$)}
    %\textit{Epanet}($S$)\; \label{algo:simu:s}
    \For{$\tau \in \mathcal{T}$}{\label{algo:simu:for:period:init}
        \textsc{Hydraulic-Simulation}($G$, $D$, $R$, $\tau$, $S$)\; \label{algo:simu:period}
        \For{$n \in V$}{ \label{algo:simu:for:node:init}
            $h_{n,\tau} \leftarrow$ \textsc{Hydraulic-Pressure}($n$)\; \label{algo:simu:head}
            \If{$h_{n,\tau} < h^{min}$}{\label{algo:simu:if:head}
                \Return \textsc{false}\;\label{algo:simu:if:head:false}
            } 
        }\label{algo:simu:for:node:end}
        \For{$p \in E$}{\label{algo:simu:for:pipe:init}
            $v_{p,\tau} \leftarrow$ \textsc{Hydraulic-Velocity}($p$)\; \label{algo:simu:velocity}
            \If{$v_{p,\tau} > v^{max}$}{\label{algo:simu:if:velocity}
                \Return \textsc{false}\;\label{algo:simu:if:velocity:false}
            }
        }\label{algo:simu:for:pipe:end}
    }\label{algo:simu:for:period:end}
    \Return \textsc{true}\; \label{algo:simu:returntrue}
\end{algorithm}

In Algorithm~\ref{algo:simu}, the for loop of lines \ref{algo:simu:for:period:init}-\ref{algo:simu:for:period:end} iterates over all periods $\tau \in \mathcal{T}$. 
The simulation in period $\tau$ for a network with the pipe types set as solution $S$ occurs in line \ref{algo:simu:period} with a call to \textsc{Hydraulic-Simulation}($G$, $D$, $R$, $\tau$, $S$).  
\textsc{Hydraulic-Simulation} uses the hydraulic simulator which computes the hydraulic head of the nodes and the flow in the pipes for a fixed set of reservoir levels and water demands over a succession of points in time.
Determining the values for heads and flows at a particular point in  time  involves  solving  simultaneously  the  conservation  of  flow  equation  for  each junction and the head loss relationship across each link of the network.
It determines the heads and flows in period $\tau$ with pipe configuration $S$ by solving simultaneously the conservation of flow given by constraints (\ref{eq:mass}) for each node and the head loss relationship determined by constraints (\ref{eq:energy}) across each pipe of the network. 
The for loop in lines \ref{algo:simu:for:node:init}-\ref{algo:simu:for:node:end} iterates over all nodes $n \in V$ and verifies whether constraints (\ref{eq:pressure}) are satisfied. 
Procedure \textsc{Hydraulic-Pressure}($n$) invoked in line \ref{algo:simu:head} retrieves the pressure on node $n$ based on the latest performed simulation. Whenever (\ref{eq:pressure}) is violated, the algorithm returns \textsc{false} in line~\ref{algo:simu:if:head:false}.
Next, the for loop in lines \ref{algo:simu:for:pipe:init}-\ref{algo:simu:for:pipe:end} iterates for all pipes $p \in E$ and verifies if constraints (\ref{eq:velocity}) are satisfied.  Procedure \textsc{Hydraulic-Velocity}($p$) invoked in line \ref{algo:simu:if:head} retrieves the velocity on pipe $p$ based on the latest performed simulation. Whenever (\ref{eq:velocity}) is violated, the algorithm returns \textsc{false} in line~\ref{algo:simu:if:velocity:false}.
Algorithm~\ref{algo:simu} returns \textsc{true} in line~\ref{algo:simu:returntrue} if solution $S$ is feasible.

\subsection{Initial solution} \label{sec:constructive}
As proposed by \citeA{DeCSor16a}, the initial solution procedure attempts to construct a low-cost solution in which all pipes have the same diameter. 
The main idea of the approach is to start with all pipes having the largest diameter possible, and iteratively reduce the diameter of all pipes in the network while a feasible candidate solution can be obtained.

The procedure to build an initial feasible solution takes as input the water distribution network $G$, the demands $D$, the supplies $R$, the planning horizon $\mathcal{T}$ representing the different scenarios to be satisfied, the minimum pressure $h^{min}$ allowed at each node, the maximum velocity $v^{max}$ permitted at each pipe, as well as the set of available pipe types $T$.
The pseudocode of the \textsc{Initial-Solution} procedure is presented in Algorithm~\ref{algo:initsol}.

\begin{algorithm}[H]
\label{algo:initsol}
\caption {\textsc{Initial-Solution}($G$, $D$, $R$, $\mathcal{T}$, $h^{min}$, $v^{max}$, $T$)}
    \For{$t = T,\ldots,1$ %in nonincreasing order %of diameter $d_t$
    }{\label{algo:initsol:for:type:init}
        $S' \leftarrow$ solution obtained with $S'(p) = t$ for all $p \in E$\; \label{algo:initsol:type} 
        
        \If{\textsc{Solution-Validation}($G$, $D$, $R$, $\mathcal{T}$, $h^{min}$, $v^{max}$, $S'$)}{\label{algo:initsol:if:simu}
            $S \leftarrow S'$\; \label{algo:initsol:sol}
        }
        \ElseIf{$t \neq |T|$}{\label{algo:initsol:if:type}
            %$f \leftarrow \floor{\frac{t - 1}{2}}$\; \label{algo:initsol:f1}
            %\Return $S$, $f$\;
            \Return $S$\;
            \label{algo:initsol:s}
        }
        \Else{
            \textbf{terminate} ``No feasible solution exists using the same type for all pipes.''\; \label{algo:initsol:else:type}
        }
    }\label{algo:initsol:for:type:end}
    %$f \leftarrow 1$\; \label{algo:initsol:f2}
    \Return $S$\; \label{algo:inisol:opt}
\end{algorithm}

{As exposed in Equation~(\ref{eq:hw}), the head loss in a pipe is inversely proportional to the diameter of the pipe. As a consequence, the constraint of minimum pressure allowed at each node is more likely to be satisfied by large diameter pipes.}
For this reason, the procedure starts with the for loop in lines \ref{algo:initsol:for:type:init}-\ref{algo:initsol:for:type:end} that iterates over the pipe types from the largest to the smallest diameter.
A candidate solution $S'$, which is not necessarily feasible, is created in line \ref{algo:initsol:type} with all pipes set as type $t$. 
Solution $S'$ is evaluated in line \ref{algo:initsol:if:simu} and in case it is feasible, it is taken as current solution, represented by $S$, in line \ref{algo:initsol:sol}. 
A solution $S$ from a previous iteration is returned in line \ref{algo:initsol:s} if at least one decrease of pipe type was performed. 
Line \ref{algo:initsol:else:type} deals with the case in which it is not possible to construct a feasible solution with all pipes set as the largest pipe diameter available.
In case a feasible candidate solution can be obtained with all pipes set as the smallest type, i.e., the cheapest pipe with the smallest diameter, that solution is returned in line \ref{algo:inisol:opt}.

\subsection{Local search} \label{sec:localsearch} 

{The local search} attempts to achieve improvements by iteratively performing small changes in the obtained solutions.
At each iteration of the local search, a pipe is selected using a greedy randomized~\cite{ResRib16} mechanism to have its diameter reduced. Whenever reducing the diameter of the selected pipe leads to an infeasible candidate solution, such move is inserted into a tabu list \cite{GloLag98} to prevent the algorithm to waste computational effort trying to perform changes which were attempted without success. 

Two important features differentiate our approach from that of \citeA{DeCSor16a}. Firstly, the selection of pipes to have the diameter reduced considers a list of pipes to be avoided, which are those in shortest paths between source nodes and those with highest demands. This intents to keep these pipes with larger diameters while the other pipes are further reduced.
The intuition behind this idea is to have a smaller quantity (in meters) of pipes with larger diameters from which pipes with smaller diameters would branch to serve other nodes. Therefore, our approach uses this information about shortest paths to increase the chances of achieving such type of low cost solutions.
Secondly, the convergence of the algorithm was also improved by allowing a more aggressive diameter reduction, which may also permit a diversification in the obtained candidate solutions.

Given an incumbent solution $S$, its neighborhood {$\mathcal{N}(S)$} is composed of all solutions having configurations in which all pipes but one, which has its diameter reduced, have the same diameter as $S$.
The \textsc{Local-Search} procedure takes as input the water distribution network $G$, the demands $D$, the supplies $R$, the planning horizon $\mathcal{T}$, the minimum pressure $h^{min}$ allowed at each node, the maximum velocity $v^{max}$ permitted at each pipe, 
the greediness factor $\alpha \in [0,1]$,
the type reduction factor $f$, as well as the current solution $S$.
The pseudocode of the \textsc{Local-Search} procedure is presented in Algorithm \ref{algo:ls}. 

\begin{algorithm}[H]
\label{algo:ls}
\caption {\textsc{Local-Search}($G$, $D$, $R$, $\mathcal{T}$, $h^{min}$, $v^{max}$, $\alpha$, $f$, $S$)}
    
    $L^{tabu} \leftarrow \emptyset$\; \label{algo:ls:tabu}
    
    $L^{path} \leftarrow $ \textsc{Path-List}$(\text{SPT}, \mathcal{T}, D, \alpha)$\; \label{algo:ls:spath}
    
    $\textit{improve} \leftarrow \textsc{true}$\; \label{algo:ls:improving}
    
    \While{$\textit{improve}$}{\label{algo:ls:while:improving:init}
        $\textit{improve} \leftarrow \textsc{false}$\; \label{algo:ls:improving:false}
        
        %$fails \leftarrow 0$\; \label{algo:ls:cfails}
        
        $L^P \leftarrow  E \setminus L^{tabu}$\;
        \label{algo:ls:list}
        
        \While{$L^P \neq \emptyset$}{\label{algo:ls:while:pipe:init}
            %$RCL \leftarrow \{p \in L^P: p \leq \alpha \lfloor |L^P| \rfloor$ in nonincreasing order of the length $l_p$\}\; \label{algo:ls:rcl}
            Create $\text{RCL}(L^P, \alpha)$ with the best elements of $L^P$\; \label{algo:ls:rcl}
            
            %Randomly choose $p \in RCL(L^P, L^{path} \text{, } \alpha)$\; \label{algo:ls:random}
            %$p \leftarrow$ \textit{Randmoly-Criterion}$(RCL(L^P, \alpha), L^{path})$\; \label{algo:ls:random}
            
            \If{$\text{RCL}(L^P, \alpha) \setminus L^{path} \neq \emptyset$}{ \label{algo:ls:if:rcl}
                Randomly choose $p \in \text{RCL}(L^P, \alpha) \setminus \{L^{path}\}$\; \label{algo:ls:if:rcl:p}
            }
            \Else{
                Randomly choose $p \in \text{RCL}(L^P, \alpha)$\; \label{algo:ls:else:rcl:p}
            }
            \If{$S(p) > f$}{\label{algo:ls:if:tabu}
                
                %$S' \leftarrow$ solution  with assigned type $S'(p) = S(p) -f$\; 
                $S' \leftarrow \text{ solution obtained with }\bigg\{$%
                \begin{tabular}{l}%
                $S'(p) = S(p) - f \text{ and}$\\%
                $S'(p') = S(p') \text{ for all } p' \in E \setminus \{p\}$\;%
                \end{tabular} \label{algo:ls:new}%
                
                \If{\textsc{Solution-Validation}($G$, $D$, $R$, $\mathcal{T}$, $h^{min}$, $v^{max}$, $S'$)}{\label{algo:ls:simulation}
                    $S \leftarrow S'$\; \label{algo:ls:s}
                    $\textit{improve} \leftarrow \textsc{true}$\; \label{algo:ls:true}
                }
                \Else{
                    $L^{tabu} \leftarrow L^{tabu} \cup \{p\}$\; \label{algo:ls:tabu:p}
                    %$fails \leftarrow fails + 1$\; \label{algo:ls:fails}
                }
            }
        $L^P \leftarrow L^P \setminus \{p\}$\; \label{algo:ls:remove}
        } \label{algo:ls:while:pipe:end}
        
        %\If{$fails > (1 - \alpha)|E|$ \KwAnd $f > 1$}{
        %    $f \leftarrow f - 1$\;
        %} 
    } \label{algo:ls:while:improving:end}
    \Return $S$\;
\end{algorithm}

In Algorithm \ref{algo:ls}, 
the memory set $L^{tabu}$ that stores %indicates 
moves which led to infeasible candidate solutions is created as empty in line \ref{algo:ls:tabu} .
Consider the base demand of node $n$, $d_n = \underset{\tau \in \mathcal{T}}{\mathrm{min}} \: \{D_{n, \tau}\} \text{ for all } n \in V$, to be its smallest demand that occurs during the planning horizon.
Define $d^{max} = \underset{n \in V}{\mathrm{max}}\{d_n\}$ and $d^{min} = \underset{n \in V}{\mathrm{min}}\{d_n\}$ to be, respectively, the maximum and minimum base demands of the network~$G$ in the planning horizon.
Consider a shortest path tree (SPT) obtained when finding the shortest path from the reservoirs to all other nodes in $G$, considering that the distance between nodes $u$ and $v$ is measured by the length $l_e$ of the pipe $e=uv$ connecting them. Note that SPT can be obtained using a variation of Dijkstra's Algorithm \cite{Dij59} which can be implemented to run in $O(|E| + |V|\log|V|)$.

A set $L^{path}$ is obtained in line \ref{algo:ls:spath} with a call to \textsc{Path-List}$(SPT, \mathcal{T}, D, \alpha)$, which retrieves from SPT a set containing the pipes that are in the shortest path from the nodes that have a base demand $d_n$ in the interval $\big[d^{max} - \alpha(d^{max} - d^{min}), d^{max}\big]$ to their nearest reservoirs.
We remark that \textsc{Path-List} only needs to be invoked once before the call to \textsc{Local-Search} as the information does not change throughout the algorithm, and it is shown in \textsc{Local-Search} just for easiness of explanation.  

The variable $\textit{improve}$, which is set as $\textsc{true}$ in line \ref{algo:ls:improving}, keeps the local search procedure running as long as improvements in the current solution $S$ are obtained.
The outer while loop of lines \ref{algo:ls:while:improving:init}-\ref{algo:ls:while:improving:end} iterates while the flag $\textit{improve}$ is set as $\textsc{true}$, that is, the solution $S$ was improved in the previous iteration  which indicates opportunity for further improvements.
For every iteration of the outer loop, the flag $improve$ is set to $\textsc{false}$ in line \ref{algo:ls:improving:false}.
The set $L^P$ containing all pipes of the network which are not tabu is created in line \ref{algo:ls:list}.
The inner while loop of lines \ref{algo:ls:while:pipe:init}-\ref{algo:ls:while:improving:end} iterates while the set $L^P$ contains %has 
pipes that did not experience a type reduction attempt. 
Consider $l^{max} = \underset{p \in L^P}{\mathrm{max}}(l_p)$ and $l^{min} = \underset{p \in L^P}{\mathrm{min}}(l_p)$ to be, respectively, the maximum and minimum lengths of pipes in $L^P$.
A restricted candidate list $\text{RCL}(L^P, \alpha)$ is created in line \ref{algo:ls:rcl} containing pipes $p \in L^P$ with lengths $l_p$ in the interval $\big[l^{max} - \alpha(l^{max} - l^{min}), l^{max}\big]$. 
Whenever $\text{RCL}(L^P, \alpha) \setminus L^{path} \neq \emptyset$ a pipe $p$ is randomly chosen from $\text{RCL}(L^P, \alpha) \setminus L^{path}$ as shown in line \ref{algo:ls:if:rcl:p}. 
Otherwise, a pipe $p$ is randomly chosen from $\text{RCL}(L^P, \alpha)$ as represented in line \ref{algo:ls:else:rcl:p}.
Whenever the selected pipe $p$ is not in the memory set $L^{tabu}$ and its type identification is greater than the factor size $f$, the new candidate solution $S'$, which is not necessarily feasible, is created as the solution $S$ but with the type of pipe $p$ reduced with a factor $f$ in line \ref{algo:ls:new}.
Then, a hydraulic simulation is performed in line \ref{algo:ls:simulation}. 
Whenever the solution $S'$ does not violate any constraints, solution $S$ is replaced by $S'$ as a new current solution in line \ref{algo:ls:s} and the flag $improve$ is set to $\textsc{true}$ in line \ref{algo:ls:true}.  
Otherwise, the attempt to reduce the type of pipe $p$ is added to the tabu memory set $L^{tabu}$ in line \ref{algo:ls:tabu:p}.
Finally, pipe $p$ is removed from the set $L^P$ in line \ref{algo:ls:remove}, restarting the inner loop. 
Algorithm \ref{algo:ls} returns a possibly new incumbent solution $S$.

\subsection{Perturbations} \label{sec:perturbation}
Perturbations are used to allow the search to escape from locally optimal solutions. 
Even though this can be achieved by restarting from a random solution (so-called \emph{multistart} local search), using perturbations is a more efficient approach, since it does not completely destroy the solution built during the previous local search iteration. \rafaelC{Some limited preliminary experiments indeed demonstrate that multi-start local search is outperformed by iterated local search in terms of solution quality.} Two different perturbations are employed in our metaheuristic. 
The perturbation used in \citeA{DeCSor16a} randomly selects a pipe and increases its diameter by one size type until a certain percentage of pipes of the water distribution network has been disturbed.
In this work, we improve this concept by attempting to increase the diameter by one size type for a certain percentage of the pipes in the water distribution network at once, which provides savings in the number of simulations required during the procedure. We call this a ``dispersed'' perturbation.
Additionally, a new ``concentrated'' perturbation is added.
This procedure randomly selects a pipe amongst the most expensive pipes of the network and increases a certain amount of its neighboring pipes by one type size.
The intuition of the concentrated perturbation is to focus the changes in a specific region of the network that contains a high cost pipe. The aim is to allow the next local search to find some improvements in this region that could not be found without the concentrated perturbation.

\subsubsection{Dispersed perturbation}
The \textsc{Dispersed-Perturbation} procedure receives as input the water distribution network $G$, the demands $D$, the supplies $R$, the planning horizon $\mathcal{T}$, the minimum pressure $h^{min}$ allowed at each node, the maximum velocity $v^{max}$ permitted in each pipe, the current solution $S$, as well as a greediness factor $\alpha \in [0,1]$. 
The pseudocode of the \textsc{Dispersed-Perturbation} is presented in Algorithm \ref{algo:disp:pert}.

\begin{algorithm}[H]
\label{algo:disp:pert}
\caption{\textsc{Dispersed-Perturbation}($G$, $D$, $R$, $\mathcal{T}$, $h^{min}$, $v^{max}$, $S$, $\alpha$)}
    $m \leftarrow \floor{\alpha|E|}$\; \label{algo:pert:k}
    \While{$m > 0$}{\label{algo:pert:while:out:init}
        $L^P \leftarrow E$\; \label{algo:pert:list}
        \While{$L^P \neq \emptyset$}{\label{algo:pert:while:init}
            $P' \leftarrow $ subset with $\min\{m,|L^P|\}$ pipes randomly chosen from $L^P$\; \label{algo:pert:p}
            %$S' \leftarrow$ solution  with assigned type $S'(p) = S(p) +1$\; \label{algo:pert:s'}
            $S' \leftarrow$ solution obtained from $S$ with $\bigg\{$%
            \begin{tabular}{l}%
            $S'(p) = S(p) + 1, \text{ for all } p \in P', \text{ and}$\\%
            $S'(p) = S(p), \text{ for all } p \in E \setminus \{P'\}$\;%
            \end{tabular} \label{algo:pert:s'}%
            
            \If{\textsc{Solution-Validation}($G$, $D$, $R$, $\mathcal{T}$, $h^{min}$, $v^{max}$, $S'$)}{\label{algo:pert:if:simu}
                %$S \leftarrow S'$\; \label{algo:pert:s}
                \Return $S'$\;
                %$n_p \leftarrow n_p + 1$\; \label{algo:pert:n1}
            }
            Randomly choose $p' \in P'$\; \label{algo:pert:p'}
            %$L^P \leftarrow L^P \setminus \{P'\}$\; \label{algo:pert:remove}
            $L^P \leftarrow L^P \setminus \{p'\}$\; \label{algo:pert:remove}
        }\label{algo:pert:while:end}
        $m \leftarrow \floor{\frac{m}{2}}$\; \label{algo:pert:max}
        %$it = it + 1$\;
    }\label{algo:pert:while:out:end}
    \Return S\; \label{algo:pert:return}
\end{algorithm}

The procedure begins initializing the maximum amount $m$ of pipes to have their types changed in line \ref{algo:pert:k} based on a percentage $\alpha$ of the number of pipes in the network.
The outer while loop in lines \ref{algo:pert:while:out:init}-\ref{algo:pert:while:out:end} proceeds until $m$ reaches zero or a feasible solution is obtained.
The set $L^P$ is created with all pipes of the network in line \ref{algo:pert:list}. 
The inner while loop in lines \ref{algo:pert:while:init}-\ref{algo:pert:while:end} iterates while there are still pipes remaining in $L^P$ and a feasible solution is not found.
A subset $P'$ is populated with $min\{m, |L^P|\}$ pipes randomly chosen from $L^P$ as detailed in line \ref{algo:pert:p}. 
A candidate solution $S'$, which is not necessarily a feasible solution, is created based on the current solution $S$ but for every pipe $p \in P'$ with its type incremented by one as presented in line \ref{algo:pert:s'}.
The candidate solution $S'$ is evaluated in line \ref{algo:pert:if:simu}. 
In case $S'$ does not violate any constraints, it is returned as feasible perturbed solution. 
Otherwise, a pipe $p'$ is randomly chosen from $P'$ in line \ref{algo:pert:p'} and removed from $L^P$ in line \ref{algo:pert:remove}.
Therefore, in the next iteration of the inner while loop it is guaranteed that the new set of pipes chosen for $P'$ from $L^P$ is different from the previous iteration.
For the next iteration of the outer while loop, the value of $m$ is rounded down with a geometric progression to the nearest integer of $\floor{\frac{m}{2}}$.
Therefore, at each iteration $m$ tends to zero. 
Lastly, if a feasible perturbation could not be performed, the current solution $S$ is returned in line \ref{algo:pert:return}.
Algorithm \ref{algo:disp:pert} returns a new candidate solution $S'$ whenever one is obtained, otherwise it returns the input solution $S$.

\subsubsection{Concentrated perturbation}

For the concentrated perturbation, consider the well-known and textbook knowledge breadth-first search (BFS) algorithm that traverses a graph $G$ in layers exploring the neighboring nodes from a given source of nodes with a runtime complexity $O(|V| + |E|)$ \cite{Cor09}.
Assume the sequence $K = \langle 1, 2, \ldots, |K| \rangle$ as the distance levels found by BFS.
Therefore, each pipe $p \in E$ can be classified with a distance level $k \in K$ from the source nodes based on the number of pipes in the path.
Let $w: L^P \rightarrow K$ be a function which assigns each pipe $p \in L^P$ to a distance level $k \in K$.
Consider an auxiliary procedure, denoted \textsc{Selection-Criterion}, whose purpose is to choose a fixed number of elements from a given set of pipes, while giving priority to pipes that are at low distance levels.

The pseudocode of \textsc{Selection-Criterion} is detailed in Algorithm~\ref{algo:slct}.
It takes as input a set of pipes $L^P$, a sequence of distance levels $K$ and a variable $m$ that gives the maximum number of pipes to be selected.

\begin{algorithm}[H]
\label{algo:slct}
\caption{\textsc{Selection-Criterion}$(L^P, K, m)$}
$r_1 \leftarrow min\{m, |L^P|\}$\; \label{algo:slct:r}
$P' \leftarrow \emptyset$\; \label{algo:slct:p'}

\ForEach{$k \in K$}{
    $C_k \leftarrow \{p \in L^P \mid w(p) = k\}$\; \label{algo:slct:ck}
    $r_2 \leftarrow |C_k|$\; \label{algo:slct:rl}
    \ForEach{$p \in C_k$}{
        \If{$r_1 > 0$}{
            % $rnd \leftarrow random(0,1)$\; \label{algo:slct:rnd}
            \If{$random(0,1) < {}^{r_1}/_{r_2}$}{
                $P' \leftarrow P' \cup \{p\}$\; \label{algo:slct:p:p'}
                $r_1 \leftarrow r_1 - 1$\;  \label{algo:slct:r:1}
            }
            $r_2 \leftarrow r_2 - 1$\;  \label{algo:slct:rl:1}
        }
        \Else{
            \Return $P'$\; \label{algo:slct:return} 
        }
    }
}
\end{algorithm}

\textsc{Selection-Criterion} begins setting variable $r_1$, representing the amount of elements yet to be selected by the procedure, to $min\{m, |L^P|\}$ in line \ref{algo:slct:r} . Also, a set $P'$ containing the pipes to be returned is initialized as empty in line \ref{algo:slct:p'}.
Next, for each element $k$ of the sequence $K$, a set $C_k$ is created with the pipes $p \in L^P$ such that $w(p) = k$, as shown in line \ref{algo:slct:ck}.
Let the variable $r_2$, which is set in line \ref{algo:slct:rl}, be the number of pipes in $C_k$ that have not been evaluated by the procedure.
Afterwards, for each element $p \in C_k$, whenever there are still elements to be selected, i.e. $r_1 > 0$, a pipe $p$ is selected with a probability $Pr(random(0,1) < {}^{r_1}/_{r_2})$. 
Remark that, $random(0,1)$ is a real-valued random variable defined on the sample space of the uniform distribution in the interval $[0,1]$.
Whenever pipe $p$ is selected with a probability $Pr(rnd < {}^{r_1}/_{r_2})$, it is inserted into $P'$ as shown in line \ref{algo:slct:p:p'} and $r_1$ is decremented by one in line \ref{algo:slct:r:1}.
After each evaluation, $r_2$ is decremented by one in line \ref{algo:slct:rl:1}.
Otherwise, with $r_1$ pipes selected, the set $P'$ is returned in line \ref{algo:slct:return}.
Observe that a call to \textsc{Selection-Criterion} always returns a nonempty set $P'$ in line \ref{algo:slct:return} assuming that the input $L^P$ is not empty.

The \textsc{Concentrated-Perturbation} procedure receives as input the water distribution network $G$, the demands $D$, the supplies $R$, the planning horizon $\mathcal{T}$, the minimum pressure $h^{min}$ allowed at each node, the maximum velocity $v^{max}$ permitted in each pipe, the current solution $S$, as well as a greediness factor $\alpha \in [0,1]$. 
The pseudocode of the \textsc{Concentrated-Perturbation} is detailed in Algorithm~\ref{algo:conc:pert}. 

\begin{algorithm}[H]
\label{algo:conc:pert}
\caption{\textsc{Concentrated-Perturbation}$(G, D, R, \mathcal{T}, h^{min}, v^{max}, S, \alpha)$}
    Create $\text{RCL}(E, S, \alpha)$ with the best elements of $S$\; \label{algo:conc:pert:rcl}
    Randomly choose $p \in RCL(E, S, \alpha)$\;
    \label{algo:conc:pert:p}
    % $I_p \leftarrow \{u, v \mid p = uv\}$\;
    $K \leftarrow BFS(G, \{u, v \mid p = uv\})$\; \label{algo:conc:pert:bfs}
    $m \leftarrow \floor{\alpha|E|}$\; \label{algo:conc:pert:k}
    \While{$m > 0$}{\label{algo:conc:pert:outer:whie:init}
        %$L^P \leftarrow$ list containing all elements of $p' \in E \setminus \{p\}$\; \label{algo:conc:pert:list}
        $L^P \leftarrow E \setminus \{p\}$\; \label{algo:conc:pert:list}
        \While{$L^P \neq \emptyset$}{\label{algo:conc:pert:while:init}
            $P' \leftarrow$ \textsc{Selection-Criterion}$(L^P, K, m)$\; \label{algo:conc:pert:P'}
            $S' \leftarrow$ solution obtained from $S$ with $\bigg\{$%
            \begin{tabular}{l}%
            $S'(p) = S(p) + 1 \text{ for all } p \in P', \text{ and}$\\%
            $S'(p) = S(p) \text{ for all } p \in E \setminus \{P'\}$\;%
            \end{tabular} \label{algo:conc:pert:s'}%
            
            \If{\textsc{Solution-Validation}$(G, D, R, \mathcal{T}, h^{min}, v^{max}, S')$}{\label{algo:conc:pert:if:simu}
                \Return $S'$\; \label{algo:conc:pert:return:s'}
            }
            Randomly choose $p' \in P'$\; \label{algo:conc:pert:p'}
            $L^P \leftarrow L^P \setminus \{p'\}$\; \label{algo:conc:pert:remove}
        }\label{algo:conc:pert:while:end}
        $m \leftarrow \floor{\frac{m}{2}}$\; \label{algo:conc:pert:max}
    }\label{algo:conc:pert:outer:whie:end}
    \Return $S$\; \label{algo:conc:pert:return}
\end{algorithm}

Given a solution $S$, consider $\psi_S(p)$ to be the cost of a pipe $p$ obtained as the cost of its type per unit of length $c_{S(p)}$ multiplied by its length $l_p$, $\psi_S(p) = c_{S(p)} \cdot l_p$. 
Remark that, the type $t$ of a pipe $p$ is determined by solution $S$, i.e., $t = S(p)$. Also, let  $\psi^{max}_S = \underset{p \in E}{\mathrm{max}}\{\psi_S(p)\}$ and $\psi^{min}_S = \underset{p \in E}{\mathrm{min}}\{\psi_S(p)\}$ be, respectively, the costs of the most expensive and of the cheapest pipes of the network in a solution $S$.
A restricted candidate list $RCL(E, S, \alpha)$ containing every pipe $p \in E$ whose cost lies in the interval $\big[ \psi^{max}_S - \alpha(\psi^{max}_S - \psi^{min}_S), \psi^{max}_S \big]$ is defined in line \ref{algo:conc:pert:rcl}. In order to avoid an excessively restricted list, we always ensure that at least the five best elements are included.
The pipe $p$ is randomly chosen from the list $RCL(E, S, \alpha)$ as shown in line \ref{algo:conc:pert:p}.
Consider as source nodes those which are incident to $p$.
Procedure $BFS(G, \{ u, v \mid p = uv\})$ invoked in line \ref{algo:conc:pert:bfs} traverses a graph $G$ starting from nodes $u,v$ and returns the distance levels as a sequence $K = \langle 1, 2, \ldots, |K| \rangle$.
The maximum amount $m$ of pipes to have their types changed is defined as a percentage $\alpha$ of the number of pipes in the network in line \ref{algo:conc:pert:k}.
The outer while loop in lines \ref{algo:conc:pert:outer:whie:init}-\ref{algo:conc:pert:outer:whie:end} proceeds until $m$ reaches zero or a feasible solution is obtained.
The set $L^P$ is created with all pipes of the network except $p$ in line \ref{algo:conc:pert:list}.
The inner while loop in lines \ref{algo:conc:pert:while:init}-\ref{algo:conc:pert:while:end} iterates while there are still pipes in the set $L^P$ and a feasible solution is not found.
The \textsc{Selection-Criterion}($K$, $L^P$, $m$) procedure invoked in line \ref{algo:conc:pert:p'} returns $min\{m, |L^P|\}$ pipes from the set $L^P$.
The candidate solution $S'$, which is not necessarily feasible, is created based on the current solution $S$ but for every pipe $p \in P'$ with its type incremented by one as presented in line \ref{algo:conc:pert:s'}.
The candidate solution $S'$ is evaluated in line \ref{algo:conc:pert:if:simu}.
In case $S'$ does not violate any constraints, it is returned as current solution in line \ref{algo:conc:pert:return:s'}.
Otherwise, a pipe $p'$ is randomly chosen from $P'$ in line \ref{algo:conc:pert:p'} and removed from $L^P$ in line \ref{algo:conc:pert:remove}.
Therefore, in the next iteration of the inner while loop it is guaranteed that the new set of pipes chosen for $P'$ is different from the previous iteration.
For the next iteration of the outer while loop, the value of $m$ is rounded down with a geometric progression to the nearest integer of $\floor{\frac{m}{2}}$ as shown in line \ref{algo:conc:pert:max}.
Therefore, at each iteration $m$ tends to zero. 
Lastly, if a feasible perturbation could not be performed, the current solution $S$ is returned in line \ref{algo:conc:pert:return}.

\subsection{Acceptance criterion} \label{sec:acceptance}

As the local search algorithm converges, it may reach a local optimum caused by the intensification in certain neighborhoods.
The acceptance criterion determines whether a newly proposed candidate solution should replace the current one and can be used to control the balance between intensification and diversification during the search and potentially lead to better solutions \cite{LouMarStu03}.

Differently from \citeA{DeCSor16a}, where the authors use a first improvement paradigm which accepts a new solution whenever the current one is improved, we propose an acceptance criterion approach which uses a \emph{pool} of solutions. The aim is to %attempt achieving 
achieve a compromise between diversification and intensification. The diversification is achieved by using solutions with possibly different characteristics, i.e., with the potential to allow the search to move into unexplored regions of the search space. 
On the other hand, intensification is produced by the prioritization of solutions that are more likely to improve the best known.
These two concepts are achieved by allowing to select solutions which are either in the pool or are the best encountered ones.

%Denote 
Define $z(S)$ to be the cost of a solution $S$ defined according to the objective function in Equation (\ref{eq:objfunc}).
The algorithm works with two solutions indicating its current state, namely,
$S^{\text{cur}}$ represents an incumbent solution from which the algorithm iteratively generates a new candidate solution from its neighborhood
$\mathcal{N}(S^{\text{cur}})$,  denoted by $S^{\text{cand}}$.
Let $S^{\text{best}}$ be the solution with the lowest cost found so far. Consider the pool of solutions $\Pi^{\nu}$ to be a memory structure which stores $\nu$ solutions and $S^{\text{worst}}_{\Pi^{\nu}}$ to be the highest cost solution in $\Pi^{\nu}$. This memory structure is maintained through decisions that determine when a solution is replaced and guarantee its diversity.

Procedure \textsc{Acceptance-Criterion} takes as inputs the best solution found so far $S^{\text{best}}$, the candidate solution $S^{\text{cand}}$, the incumbent solution $S^{\text{cur}}$, and the memory structure $\Pi^{\nu}$.
The pseudocode of the \textsc{Acceptance-Criterion} procedure is presented in Algorithm \ref{algo:accp}. 

\begin{algorithm}[H]
\label{algo:accp}
\caption{\textsc{Acceptance-Criterion}$(S^{\text{best}}, S^{\text{cand}}, S^{\text{cur}}, \Pi^{\nu})$}
    \If{$z(S^{\text{cand}}) < z(S^{\text{cur}})$
    \label{algo:accp:if}}{
        \If{$z(S^{\text{cand}}) < z(S^{\text{best}})$}{
            $S^{\text{best}} \leftarrow S^{\text{cand}}$\; \label{algo:accp:best}
        }
        \Else {
            $S^{\text{worst}}_{\Pi^{\nu}} \leftarrow S^{\text{cand}}$\; \label{algo:accp:worst}
        }
        \Return $S^{\text{cand}}$\; \label{algo:accp:rcand}
    }
    \Else{
    \Return Randomly choose an element from $\Pi^{\nu} \cup \{S^{\text{best}}\}$\; \label{algo:accp:rrandom}
    }
\end{algorithm}

Algorithm \ref{algo:accp} starts testing whether the candidate solution $S^{\text{cand}}$ improves the incumbent one $S^{\text{cur}}$. 
Whenever this happens, in case it also improves the best solution $S^{\text{best}}$, $S^{\text{best}}$ is replaced by $S^{\text{cand}}$ as shown in line~\ref{algo:accp:best}. 
Otherwise, $S^{\text{cand}}$ replaces the worst solution in the memory $S^{\text{worst}}_{\Pi^{\nu}}$ as shown in line \ref{algo:accp:worst}.
Later, the procedure returns the candidate solution $S^{\text{cand}}$ in line \ref{algo:accp:rcand}.
In case the candidate solution $S^{\text{cand}}$ does not improve the incumbent one $S^{\text{cur}}$, the procedure returns a solution randomly chosen from $\Pi^{\nu} \cup S^{\text{best}}$ as shown in line \ref{algo:accp:rrandom}.
Observe that Algorithm \ref{algo:accp} returns either the candidate solution received as input or a solution selected from the pool of solutions or the best solution found so far.

\subsection{Simulation-based iterated local search metaheuristic} \label{sec:ils}

The proposed simulation-based iterated local search (ILS) metaheuristic embeds all the components presented previously in this section. The main idea consists of, starting from an initial solution, performing a sequence of local searches and perturbations until a time limit is reached.
The simulation-based ILS receives as input the water distribution network $G$, the demands $D$, the supplies $R$, the planning horizon $\mathcal{T}$, the available pipe types $T$, the minimum pressure $h^{min}$ at each node, the maximum velocity $v^{max}$ permitted in each pipes, the time limit $time^{max}$, the size of the memory of solutions $\nu$, the greediness factor $\alpha \in [0,1]$, and the type reduction factor $f$.
The pseudocode of the simulation-based ILS metaheuristic is described in Algorithm \ref{algo:ils}.

\begin{algorithm}[H]
\label{algo:ils}
\caption {\textsc{\small{Simulation-Based-ILS}}($G$, $D$, $R$, $\mathcal{T}$, $T$, $h^{min}$, $v^{max}$, $time^{max}$, $\nu$, $\alpha$, $f$)}
    $\Pi^{\nu} \leftarrow$ Populate with $\nu$ copies of \textsc{Initial-Solution}($G$, $D$, $R$, $\mathcal{T}$, $h^{min}$, $v^{max}$, $T$)\; \label{algo:ils:s}
    $S^{\text{cur}}, S^{\text{best}} \leftarrow S^{\text{worst}}_{\Pi^{\nu}}$\; \label{algo:ils:ccurbest}
    %\textsc{Local-Search}($G$, $D$, $R$, $\mathcal{T}$, $h^{min}$, $v^{max}$, $\alpha$, $f$, $S_0$)\; \label{algo:ils:local1}
    
    \While{$time^{max}$ \KwReach}{\label{algo:ils:while:init}
    
        $S^{\text{cand}} \leftarrow$ \textsc{Local-Search}($G$, $D$, $R$, $\mathcal{T}$, $h^{min}$, $v^{max}$, $\alpha$, $f$, $S^{\text{cur}}$)\; \label{algo:ils:local1}
        \If{$f > 1$}{
        	$f \leftarrow \floor{{}^{f}/_{2}}$ \; \label{algo:ils:factor}
        }
        
        $S^{\text{cand}} \leftarrow$ \textsc{Acceptance-Criterion}$(S^{\text{best}}, S^{\text{cand}}, S^{\text{cur}}, \Pi^{\nu})$\; \label{algo:ils:accp}

        % $rnd \leftarrow random(0 , 1)$\; \label{algo:ils:rnd}
        \If{$random(0 , 1) \leq (1 - \alpha)$}{\label{algo:ils:pert}
             $S^{\text{cand}} \leftarrow$ \textsc{Dispersed-Perturbation}$(G, D, R, \mathcal{T}, h^{min}, v^{max}, S^{\text{cand}}, \alpha)$\; \label{algo:ils:disp:pert}
        }
        \Else{
             $S^{\text{cand}} \leftarrow$ \textsc{Concentrated-Perturbation}$(G, D, R, \mathcal{T}, h^{min}, v^{max}, S^{\text{cand}}, \alpha)$\; \label{algo:ils:conc:pert}
        }
    } \label{algo:ils:while:end}
    \Return $S^{\text{best}}$\;
\end{algorithm}

\textsc{Simulation-Based-ILS} starts with the creation of the memory structure $\Pi^{\nu}$ (see Section~\ref{sec:acceptance}) containing $\nu$ copies of the initial solution returned by the call to \textsc{Initial-Solution}($G$, $D$, $R$, $\mathcal{T}$, $h^{min}$, $v^{max}$, $T$), as shown in line \ref{algo:ils:s}.
The current and best solutions, $S^{\text{cur}}$ and $S^{\text{best}}$, are defined according to $S^{\text{worst}}_{\Pi^{\nu}}$ in line \ref{algo:ils:ccurbest}, i.e., they all correspond to the initial solution. Note that such solution is both the worst and the best one in $\Pi^{\nu}$ at this moment, \rafaelC{as it is the only one obtained so far}.
The approach attempts to improve the current solution $S^{\text{cur}}$ by a call to \textsc{Local-Search}($G$, $D$, $R$, $\mathcal{T}$, $h^{min}$, $v^{max}$, $\alpha$, $f$, $S^{\text{cur}}$) in line \ref{algo:ils:local1}, and the resulting solution is stored into $S^{\text{cand}}$.
The main while loop in lines \ref{algo:ils:while:init}-\ref{algo:ils:while:end} repeats while the time limit is not reached.
Right after the \textsc{Local-Search}, whenever the factor \textit{f} is greater than one, it is divided by two as shown in line \ref{algo:ils:factor}.
Thus,  we are more likely to decrease the chance of getting infeasible solutions in the next attempts to perform type reductions in the local search.
In line \ref{algo:ils:accp} the approach determines if the recently generated $S^{\text{cand}}$ will be accepted for further iterations via a call to \textsc{Acceptance-Criterion}($S^{\text{best}}$, $S^{\text{cand}}$, $S^{\text{cur}}$, $\Pi^{\nu}$).
In line \ref{algo:ils:pert}, the approach determines which perturbation will be executed. 
For each iteration the probabilities to perform a dispersed perturbation and a concentrated perturbation are, respectively, $\alpha$ and $1 - \alpha$.
The procedure \textsc{Dispersed-Perturbation}$(G, D, R, \mathcal{T}, h^{min}, v^{max}, \alpha, S^{\text{cur}})$ is invoked in line \ref{algo:ils:disp:pert} and \textsc{Concentrated-Perturbation}($G$, $D$, $R$, $\mathcal{T}$, $h^{min}$, $v^{max}$, $\alpha$, $S^{\text{cur}}$) is performed in line \ref{algo:ils:conc:pert}.
Both methods return a perturbed solution $S^{\text{cand}}$ starting from the current solution $S^{\text{cur}}$.
Lastly, Algorithm \ref{algo:ils} returns the best feasible solution found so far $S^{\text{best}}$.

\section{Computational experiments}
\label{sec:experiments}

This section summarizes the performed computational experiments.
All the tests were carried out on a machine running under Ubuntu x86-64 GNU/Linux, with an Intel Core i7-8700 Hexa-Core 3.20GHz processor and 16Gb of RAM. 
The simulation based iterated local search metaheuristic was implemented in C++, and its interaction with the hydraulic simulator EPANET was established via the Extended EPANET Toolkit 1.5~\cite{LopezIbanezPhD, LopezIbanezTool}. 
Subsection \ref{sec:benchmark} describes the benchmark set of instances. 
Subsection \ref{sec:test} lists the tested approaches and reports the parameter settings.
Subsection \ref{sec:results} reports the results comparing our newly proposed approach with a state-of-the-art metaheuristic presented in~\citeA{DeCSor16b}.
Subsection \ref{sec:overviewresult} presents a graphical summary of the results.
Subsection~\ref{sec:performanceindividualimprovements} analyzes the impacts of each individual novelty to the solution's quality using a subset of the instances. Subsection~\ref{sec:experimentsadditionalindicators} compares some additional indicators achieved by the approach of~\citeA{DeCSor16b} and by our new improved simulation based iterated local search metaheuristic.

\subsection{Benchmark set}
\label{sec:benchmark}
 
The developed algorithm was tested using a set of challenging benchmark instances generated using HydroGen~\cite{DeCSor14}, which is a tool that generates virtual water distribution networks. 
The input files of these networks are available via~\citeA{DeCSor}. 
The benchmark instances are divided into 30 instance groups. Each instance group contains five instances with similar size, expressed by the number of pipes, and characteristics, reflected by the meshedness coefficient.
The meshedness coefficient, $M = \frac{|\Xi|}{2|N|-5}$, is a concept in graph theory which evaluates the number of loops against the maximum number of loops in a planar graph~\cite{Buhtal06}.
Remark that $|\Xi|$ and $|N|$ represent, respectively, the numbers of independent loops and of nodes in the network.
The sizes and characteristics of each instance group are given in Table~\ref{tab:instances}.

\begin{table}[H]
    \caption{Information about the instance groups.}
    \centering
    \resizebox{\textwidth}{!}{
    \begin{tabular}{p{2cm}p{2.1cm}p{2cm}p{2cm}p{2.1cm}|p{2cm}p{2.1cm}p{2cm}p{2cm}p{2.1cm}} 
        \hline
	\textbf{Network}	&	\textbf{Meshedness Coefficient}	&	\textbf{Pipes}	&	\textbf{Demand Nodes}	&	\textbf{Water Reservoirs}	& \textbf{Network}	&	\textbf{Meshedness Coefficient}	&	\textbf{Pipes}	&	\textbf{Demand Nodes}	&	\textbf{Water Reservoirs}   \\
	\hline\hline
	HG-MP-1	&	0.20	&	100	&	73	&	1 &	HG-MP-16	&	0.20	&	606	&	431	&	4	\\
	HG-MP-2	&	0.15	&	100	&	78	&	1 &	HG-MP-17	&	0.15	&	607	&	465	&	4	\\
	HG-MP-3	&	0.10	&	99	&	83	&	1 &	HG-MP-18	&	0.10	&	608	&	503	&	5	\\
	HG-MP-4	&	0.20	&	198	&	143	&	1 &	HG-MP-19	&	0.20	&	708	&	503	&	5	\\
	HG-MP-5	&	0.15	&	200	&	155	&	1 &	HG-MP-20	&	0.15	&	703	&	538	&	5	\\
	HG-MP-6	&	0.10	&	198	&	166	&	1 &	HG-MP-21	&	0.10	&	707	&	586	&	5	\\
	HG-MP-7	&	0.20	&	299	&	215	&	2 &	HG-MP-22	&	0.20	&	805	&	572	&	6	\\
	HG-MP-8	&	0.15	&	300	&	232	&	2 &	HG-MP-23	&	0.15	&	804	&	615	&	6	\\
	HG-MP-9	&	0.10	&	295	&	247	&	2 &	HG-MP-24	&	0.10	&	808	&	669	&	6	\\
	HG-MP-10	&	0.20	&	397	&	285	&	2 &	HG-MP-25	&	0.20	&	906	&	644	&	6	\\
	HG-MP-11	&	0.15	&	399	&	308	&	2 &	HG-MP-26	&	0.15	&	905	&	692	&	7	\\
	HG-MP-12	&	0.10	&	395	&	330	&	3 &	HG-MP-27	&	0.10	&	908	&	752	&	7	\\
	HG-MP-13	&	0.20	&	498	&	357	&	2 &	HG-MP-28	&	0.20	&	1008	&	716	&	7	\\
	HG-MP-14	&	0.15	&	499	&	385	&	3 &	HG-MP-29	&	0.15	&	1007	&	770	&	7	\\
	HG-MP-15	&	0.10	&	495	&	413	&	3 &	HG-MP-30	&	0.10	&	1009	&	835	&	8	\\
    \hline
    \end{tabular}}
    \label{tab:instances}
\end{table}

The set of available pipe types and their corresponding costs can be found in Table~\ref{tab:pipetypes}.
The planning horizon $\mathcal{T}$ is composed of 24 time periods (corresponding to the hourly scenarios in a day).
Additionally, a minimum pressure head of $20 m$ is required in every demand node at every time period, and a maximum velocity of $2 m/s$ is set for the water flow through every pipe in the network, at every time period. 
Demand nodes are divided into five categories (domestic, industrial, energy, public services, and commercial demand nodes), each with a corresponding base load and demand pattern. 
The base loads and demand patterns can be found in the EPANET input files of the instances~\cite{DeCSor}.
We remark that we do not use multi-period counterparts for the classic single-period instances available in the literature since these are relatively small (considering the number of pipes). 
Besides, \citeA{DeCSor16a} tested the algorithm with the classical instances and were able to reach the best-known solutions. Furthermore, the proposed approaches are directed to improve the capability of obtaining good quality solutions for more challenging larger instances.

\begin{table}[H]
    \caption{Information about the available pipe types.}
    \centering
    \resizebox{\textwidth}{!}{
    \begin{tabular}{p{1.8cm}p{2cm}p{2cm}p{2.8cm}|p{1.8cm}p{2cm}p{2cm}p{2.8cm}} 
        \hline
	\textbf{Number}	&	\textbf{Diameter (in mm)}	&	\textbf{Roughness (unitless)}	&	\textbf{Cost (in EUR per m)} & \textbf{Number}	&	\textbf{Diameter (in mm)}	&	\textbf{Roughness (unitless)}	&	\textbf{Cost (in EUR per m)}\\
	\hline\hline
	1 & 20 & 130 & 9 & 9 & 200 & 130 & 116 \\
	2 & 30 & 130 & 20 & 10 & 250 & 130 & 150 \\
	3 & 40 & 130 & 25 & 11 & 300 & 130 & 201 \\
	4 & 50 & 130 & 30 & 12 & 400 & 130 & 290 \\
	5 & 60 & 130 & 35 & 13 & 400 & 130 & 290 \\
	6 & 80 & 130 & 48 & 14 & 500 & 130 & 351 \\
	7 & 100 & 130 & 50 & 15 & 600 & 130 & 528 \\
	8 & 150 & 130 & 61 & 16 & 1000 & 130 & 628 \\
    \hline
    \end{tabular}}
    \label{tab:pipetypes}
\end{table}

\subsection{Tested approaches and settings}
\label{sec:test}
In this subsection we present the tested approaches and the preliminary experiments carried out to determine the parameters of the proposed techniques. 
We compared the following approaches:
\begin{enumerate}
    \item[(a)] the simulation based iterated local search metaheuristic of \citeA{DeCSor16b} (ILS);
    \item[(b)] our newly proposed enhanced simulation based iterated local search metaheuristic (ILS+), described in Section~\ref{sec:approaches}.
\end{enumerate}

In the computational tests, each approach was executed ten times on each test instance for each of the time limits $time^{max} \in \{60, 180, 300, 600\}$ (in seconds). The best and average costs over these ten runs are reported in the following sections.

ILS was executed with the same parameter settings used in the authors' original work.
The settings for ILS+ were determined based on preliminary computational experiments, which
%These preliminary experiments 
took into consideration around 10\% of the instances, randomly chosen, with varying sizes and characteristics.

The following values were tested for each parameter:
\begin{itemize}
    \item greediness factor: $\alpha \in \{{0.05}, 0.08, 0.10\}$;
    \item type reduction factor: $f \in \{1, {2}, 4\}$;
    \item size of the pool of solutions: $\nu \in \{1, 2, {3}\}$.
\end{itemize}
Based on the preliminary experiments, the selected configurations for the parameters were $\alpha = 0.05$, $f = 4$ and $\nu = 3$.

\subsection{Computational results}
\label{sec:results}

Tables~\ref{tab:overview1}~and~\ref{tab:overview2} summarize the performed computational results. We remind that the reported values for each combination of instance and time limit take into consideration ten runs of the corresponding approach. For each instance group, the reported best and average values are the average over the five instances in the group.

Table~\ref{tab:overview1} reports the results for the instance groups with less than 500 pipes while Table~\ref{tab:overview2} presents the results for those with more than 500 pipes. In each of these tables, the first two columns identify the instance group and the time limit (in seconds). The next two columns give, respectively, the best and average solutions achieved by ILS for the corresponding instance group. Due to their magnitude, the best and average values are presented using the scientific notation $value \times 10^6$, with four significant digits. The next six columns present the results for ILS+ and indicate how it compares against ILS. More specifically, the fifth and sixth columns indicate the best and average solutions for the corresponding instance group. The lowest values for best and average solutions are shown in boldface. The seventh and eighth columns indicate, respectively, the numbers of instances (out of the five in the instance group) the best and average values obtained by ILS+ were strictly better than those obtained by ILS. The last two columns represent the average gain, calculated for each instance as $100 \times \frac{z_{ILS+} - z_{ILS}}{z_{ILS}}$, achieved by ILS+ over ILS for best and average values, correspondingly.

Table~\ref{tab:overview1} shows that for the instances with less than 500 pipes, ILS and ILS+ present quite similar results when it comes to the best encountered solutions. As it can be seen in the two last lines of the table, ILS+ presented improved best solutions for 156 out of the 300 combinations and an average gain of only 0.5\%. However, it is remarkable that ILS+ is more robust for these instances. To be more specific, ILS+ achieved improved average results for 227 out of the 300 combinations and an average improvement of 3.6\%. As expected, considerably large gains were achieved for the more restricted time limits (e.g., with a time limit of 60 seconds, average gains of 12.4\% and 8.1\% were achieved, respectively, for instance groups HG-MP-2 and HG-MP-13).     

Table~\ref{tab:overview2} exhibits that ILS+ outperforms ILS for the instances with more than 500 pipes. The best and average results achieved by ILS+ are considerably better than those obtained by ILS for most of the instance groups. As it can be seen in the two last lines of the table, ILS+ presented improved best solutions for 228 out of the 300 combinations and an average gain of 8.4\%. Additionally, it was able to obtain improved average solutions for 240 out of the 300 combinations and an average gain of 9.3\%. 
The results evidence the fact that noticeable gains were achieved by ILS+ for all instance groups when considering the more restricted time limits.

%\begin{adjustwidth}{-2.5 cm}{-2.5 cm}\centering\begin{threeparttable}[!htb]
\begin{table}[!htp]\centering
\caption{Results obtained by ILS and ILS+ for instances with less than 500 pipes.}\label{tab:overview1}
%\scriptsize
\tiny
\resizebox{\textwidth}{!}{%
\begin{tabular}{lc|cc|ccccccc}\hline
& &\multicolumn{2}{c|}{ILS} &\multicolumn{6}{c}{ILS+}  \Tstrut\\
Instance group & Time limit  & Best & Avg & Best & Avg & \#impr & \#impr & Gain (\%) & Gain (\%) \\
 &  (seconds) & ({${10^{6}}$})  & ({${10^{6}}$}) & ({${10^{6}}$}) & ({${10^{6}}$}) & (Best) & (Avg) & (Best) & (Avg) \Bstrut\\\hline
HG-MP-1 & 60 & 0.345 & 0.358 & \textbf{0.336} & \textbf{0.341} & 3 & 3 & 2.7 & 5.0 \Tstrut\\
& 180 & 0.344 & 0.357 & \textbf{0.336} & \textbf{0.339} & 3 & 3 & 2.6 & 5.1 \\
& 300 & 0.343 & 0.357 & \textbf{0.335} & 0.\textbf{339} & 3 & 3 & 2.5 & 5.2 \\
& 600 & 0.343 & 0.357 & \textbf{0.335} & \textbf{0.339} & 3 & 3 & 2.3 & 5.1 \\\hline
HG-MP-2 & 60 & \textbf{0.301} & 0.355 & 0.303 & \textbf{0.308} & 3 & 5 & -0.9 & 12.4 \Tstrut\\
& 180 & \textbf{0.300} & 0.351 & 0.301 & \textbf{0.306} & 3 & 5 & -0.3 & 12.1 \\
& 300 & \textbf{0.300} & 0.349 & 0.301 & \textbf{0.305} & 3 & 5 & -0.5 & 11.9 \\
& 600 & \textbf{0.299} & 0.348 & 0.301 & \textbf{0.304} & 3 & 5 & -0.6 & 11.8 \\\hline
HG-MP-3 & 60 & 0.388 & 0.413 & \textbf{0.387} & \textbf{0.389} & 2 & 5 & 0.3 & 5.6 \Tstrut\\
& 180 & 0.388 & 0.408 & \textbf{0.386} & 0.\textbf{389} & 2 & 5 & 0.3 & 4.6 \\
& 300 & 0.388 & 0.407 & \textbf{0.386} & \textbf{0.389} & 2 & 5 & 0.3 & 4.6 \\
& 600 & 0.387 & 0.406 & \textbf{0.386} & \textbf{0.388} & 3 & 5 & 0.3 & 4.6 \\\hline
HG-MP-4 & 60 & \textbf{0.699} & 0.728 & 0.706 & \textbf{0.723} & 2 & 3 & -1.0 & 0.6 \Tstrut\\
& 180 & \textbf{0.695} & 0.723 & 0.702 & \textbf{0.717} & 2 & 3 & -1.1 & 0.6 \\
& 300 & \textbf{0.695} & 0.722 & 0.701 & \textbf{0.715} & 2 & 3 & -0.9 & 0.7 \\
& 600 & \textbf{0.694} & 0.721 & 0.700 & \textbf{0.713} & 2 & 3 & -0.9 & 0.9 \\\hline
HG-MP-5 & 60 & 0.748 & 0.810 & \textbf{0.740} & \textbf{0.759} & 3 & 4 & 1.0 & 6.4 \Tstrut\\
& 180 & 0.744 & 0.800 & \textbf{0.738} & \textbf{0.753} & 3 & 3 & 0.8 & 6.0 \\
& 300 & 0.742 & 0.796 & \textbf{0.738} & \textbf{0.751} & 3 & 3 & 0.5 & 5.7 \\
& 600 & 0.739 & 0.791 & \textbf{0.738} & \textbf{0.749} & 3 & 3 & 0.2 & 5.4 \\\hline
HG-MP-6 & 60 & \textbf{0.753} & \textbf{0.779} & 0.755 & 0.780 & 1 & 2 & -0.4 & -0.2 \Tstrut\\
& 180 & \textbf{0.750} & 0.775 & 0.752 & \textbf{0.773} & 1 & 2 & -0.4 & 0.0 \\
& 300 & \textbf{0.749} & 0.774 & 0.751 & \textbf{0.771} & 1 & 2 & -0.4 & 0.2 \\
& 600 & \textbf{0.741} & 0.771 & 0.749 & \textbf{0.766} & 0 & 2 & -1.1 & 0.4 \\\hline
HG-MP-7 & 60 & 0.847 & 0.905 & \textbf{0.818} & \textbf{0.856} & 3 & 5 & 3.1 & 5.0 \Tstrut\\
& 180 & 0.823 & 0.883 & \textbf{0.799} & \textbf{0.839} & 4 & 5 & 2.6 & 4.6 \\
& 300 & 0.815 & 0.880 & \textbf{0.789} & \textbf{0.834} & 4 & 5 & 2.9 & 4.9 \\
& 600 & 0.811 & 0.875 & \textbf{0.787} & \textbf{0.830} & 4 & 5 & 2.7 & 4.9 \\\hline
HG-MP-8 & 60 & 0.871 & 0.954 & \textbf{0.858} & \textbf{0.904} & 3 & 3 & 1.3 & 4.9 \Tstrut\\
& 180 & 0.862 & 0.934 & \textbf{0.846} & \textbf{0.886} & 3 & 4 & 1.7 & 5.0 \\
& 300 & 0.859 & 0.931 & \textbf{0.843} & \textbf{0.880} & 3 & 4 & 1.7 & 5.3 \\
& 600 & 0.856 & 0.927 & \textbf{0.834} & \textbf{0.873} & 4 & 4 & 2.4 & 5.6 \\\hline
HG-MP-9 & 60 & \textbf{0.844} & \textbf{0.881} & 0.861 & 0.892 & 1 & 3 & -2.1 & -1.3 \Tstrut\\
& 180 & \textbf{0.833} & \textbf{0.863} & 0.840 & 0.869 & 1 & 3 & -0.9 & -0.6 \\
& 300 & \textbf{0.828} & \textbf{0.858} & 0.836 & 0.861 & 1 & 3 & -1.0 & -0.4 \\
& 600 & 0.826 & \textbf{0.854} & \textbf{0.825} & \textbf{0.854} & 3 & 4 & -0.0 & -0.1 \\\hline
HG-MP-10 & 60 & 0.827 & 0.893 & \textbf{0.802} & \textbf{0.845} & 5 & 5 & 2.9 & 5.5 \Tstrut\\
& 180 & 0.797 & 0.855 & \textbf{0.786} & \textbf{0.812} & 4 & 5 & 1.4 & 5.0 \\
& 300 & 0.793 & 0.849 & \textbf{0.782} & \textbf{0.807} & 4 & 5 & 1.3 & 5.0 \\
& 600 & 0.791 & 0.843 & \textbf{0.778} & \textbf{0.800} & 5 & 5 & 1.7 & 5.0 \\\hline
HG-MP-11 & 60 & \textbf{0.932} & 0.998 & 0.942 & \textbf{0.991} & 1 & 4 & -1.1 & 0.6 \Tstrut\\
& 180 & \textbf{0.907} & \textbf{0.955} & 0.920 & 0.959 & 2 & 3 & -1.5 & -0.4 \\
& 300 & \textbf{0.904} & 0.948 & 0.912 & \textbf{0.945} & 2 & 3 & -1.0 & 0.4 \\
& 600 & \textbf{0.896} & 0.941 & 0.905 & \textbf{0.932} & 2 & 4 & -1.0 & 0.9 \\\hline
HG-MP-12 & 60 & \textbf{1.050} & \textbf{1.104} & 1.071 & 1.107 & 2 & 2 & -1.8 & -0.2 \Tstrut\\
& 180 & \textbf{1.024} & 1.069 & 1.048 & \textbf{1.067} & 1 & 3 & -2.3 & 0.3 \\
& 300 & \textbf{1.019} & 1.061 & 1.040 & \textbf{1.059} & 1 & 3 & -2.0 & 0.3 \\
& 600 & \textbf{1.014} & 1.053 & 1.033 & \textbf{1.050} & 0 & 4 & -1.9 & 0.4 \\\hline
HG-MP-13 & 60 & 1.321 & 1.419 & \textbf{1.214} & \textbf{1.301} & 5 & 5 & 8.0 & 8.1 \Tstrut\\
& 180 & 1.189 & 1.274 & \textbf{1.174} & \textbf{1.221} & 3 & 5 & 1.3 & 4.2 \\
& 300 & 1.174 & 1.247 & \textbf{1.164} & \textbf{1.204} & 3 & 5 & 0.9 & 3.5 \\
& 600 & 1.161 & 1.226 & \textbf{1.155} & \textbf{1.184} & 3 & 5 & 0.6 & 3.5 \\\hline
HG-MP-14 & 60 & 1.166 & 1.257 & \textbf{1.125} & \textbf{1.177} & 3 & 4 & 2.4 & 5.4 \Tstrut\\
& 180 & \textbf{1.089} & 1.141 & 1.093 & \textbf{1.121} & 3 & 3 & -0.6 & 1.5 \\
& 300 & \textbf{1.064} & 1.121 & 1.075 & \textbf{1.106} & 2 & 2 & -1.3 & 1.1 \\
& 600 & \textbf{1.056} & 1.106 & 1.070 & \textbf{1.095} & 2 & 3 & -1.5 & 0.9 \\\hline
HG-MP-15 & 60 & 1.222 & 1.314 & \textbf{1.180} & \textbf{1.240} & 4 & 4 & 3.3 & 5.5 \Tstrut\\
& 180 & 1.162 & 1.220 & \textbf{1.132} & \textbf{1.186} & 3 & 4 & 2.3 & 2.6 \\
& 300 & 1.149 & 1.196 & \textbf{1.127} & \textbf{1.177} & 3 & 4 & 1.7 & 1.2 \\
& 600 & 1.139 & 1.176 & \textbf{1.118} & \textbf{1.164} & 3 & 4 & 1.6 & 0.6 \\\hline
Total &  &  &  &  &  & 156/300 & 227/300 &  & \Tstrut\\
Average (\%) &  &  &  &  &  &  &  & 0.5 & 3.6 \\
\hline
\end{tabular}}
\end{table}
%\end{threeparttable}\end{adjustwidth}

%\begin{adjustwidth}{-2.5 cm}{-2.5 cm}\centering\begin{threeparttable}[!htb]
\begin{table}[!htp]\centering
\caption{Results obtained by ILS and ILS+ for instances with more than 500 pipes.}\label{tab:overview2}
%\scriptsize
\tiny
\resizebox{\textwidth}{!}{%
\begin{tabular}{lc|cc|ccccccc}\hline
& &\multicolumn{2}{c|}{ILS} &\multicolumn{6}{c}{ILS+}  \\
Instance group & Time limit  & Best & Avg & Best & Avg & \#impr & \#impr & Gain (\%) & Gain (\%) \\
 &  (seconds) & ({${10^{6}}$})  & ({${10^{6}}$}) & ({${10^{6}}$}) & ({${10^{6}}$}) & (Best) & (Avg) & (Best) & (Avg) \Bstrut\\\hline
HG-MP-16 & 60 & 1.821 & 2.007 & \textbf{1.619} & \textbf{1.701} & 5 & 5 & 11.3 & 15.3 \Tstrut\\
& 180 & \textbf{1.523} & 1.666 & 1.542 & \textbf{1.610} & 3 & 3 & -1.0 & 3.7 \\
& 300 & \textbf{1.487} & 1.610 & 1.527 & \textbf{1.584} & 2 & 3 & -2.4 & 2.0 \\
& 600 & \textbf{1.463} & 1.575 & 1.499 & \textbf{1.560} & 2 & 3 & -2.2 & 1.4 \\\hline
HG-MP-17 & 60 & 1.913 & 2.096 & \textbf{1.718} & \textbf{1.903} & 4 & 4 & 9.6 & 8.7 \Tstrut\\
& 180 & \textbf{1.623} & \textbf{1.764} & 1.667 & 1.802 & 4 & 4 & -3.5 & -2.5 \\
& 300 & \textbf{1.595} & \textbf{1.717} & 1.655 & 1.766 & 3 & 4 & -4.6 & -3.4 \\
& 600 & \textbf{1.573} & \textbf{1.689} & 1.632 & 1.715 & 4 & 4 & -4.5 & -2.1 \\\hline
HG-MP-18 & 60 & 1.770 & 1.964 & \textbf{1.621} & \textbf{1.759} & 5 & 5 & 8.3 & 10.6 \Tstrut\\
& 180 & \textbf{1.507} & 1.637 & 1.560 & \textbf{1.627} & 1 & 3 & -3.9 & 0.4 \\
& 300 & \textbf{1.480} & \textbf{1.597} & 1.537 & 1.601 & 1 & 2 & -4.3 & -0.4 \\
& 600 & \textbf{1.465} & \textbf{1.567} & 1.514 & 1.572 & 2 & 2 & -3.7 & -0.5 \\\hline
HG-MP-19 & 60 & 2.999 & 3.296 & \textbf{2.408} & \textbf{2.565} & 5 & 5 & 19.4 & 21.9 \Tstrut\\
& 180 & 2.253 & 2.473 & \textbf{2.193} & \textbf{2.354} & 3 & 5 & 2.8 & 4.6 \\
& 300 & 2.184 & 2.359 & \textbf{2.145} & \textbf{2.293} & 3 & 4 & 1.9 & 2.6 \\
& 600 & 2.144 & 2.297 & \textbf{2.093} & \textbf{2.220} & 3 & 4 & 2.5 & 3.1 \\\hline
HG-MP-20 & 60 & 2.928 & 3.211 & \textbf{2.215} & \textbf{2.409} & 5 & 5 & 22.7 & 23.7 \Tstrut\\
& 180 & 2.196 & 2.365 & \textbf{2.114} & \textbf{2.242} & 4 & 4 & 3.9 & 5.5 \\
& 300 & 2.095 & 2.232 & \textbf{2.046} & \textbf{2.156} & 4 & 4 & 2.5 & 3.7 \\
& 600 & 2.031 & 2.150 & \textbf{2.010} & \textbf{2.084} & 2 & 3 & 1.2 & 3.3 \\\hline
HG-MP-21 & 60 & 2.786 & 2.967 & \textbf{2.225} & \textbf{2.356} & 5 & 5 & 20.3 & 20.8 \Tstrut\\
& 180 & 2.243 & 2.385 & \textbf{2.134} & \textbf{2.207} & 5 & 5 & 4.9 & 7.5 \\
& 300 & 2.192 & 2.298 & \textbf{2.125} & \textbf{2.180} & 4 & 4 & 3.1 & 5.2 \\
& 600 & 2.157 & 2.246 & \textbf{2.108} & \textbf{2.151} & 4 & 4 & 2.3 & 4.3 \\\hline
HG-MP-22 & 60 & 4.358 & 4.758 & \textbf{3.144} & \textbf{3.401} & 5 & 5 & 27.4 & 28.0 \Tstrut\\
& 180 & 2.957 & 3.196 & \textbf{2.815} & \textbf{2.997} & 5 & 5 & 4.5 & 5.9 \\
& 300 & 2.798 & 2.965 & \textbf{2.707} & \textbf{2.885} & 5 & 5 & 3.2 & 2.5 \\
& 600 & 2.657 & 2.830 & \textbf{2.620} & \textbf{2.769} & 4 & 5 & 1.4 & 2.1 \\\hline
HG-MP-23 & 60 & 4.129 & 4.516 & \textbf{3.228} & \textbf{3.521} & 5 & 5 & 21.7 & 21.5 \Tstrut\\
& 180 & 2.950 & 3.263 & \textbf{2.883} & \textbf{3.121} & 4 & 4 & 2.2 & 4.1 \\
& 300 & \textbf{2.808} & \textbf{3.000} & 2.826 & 3.040 & 2 & 2 & -0.8 & -1.6 \\
& 600 & 2.734 & \textbf{2.856} & \textbf{2.683} & 2.917 & 2 & 2 & 1.7 & -2.4 \\\hline
HG-MP-24 & 60 & 3.765 & 4.062 & \textbf{2.825} & \textbf{2.975} & 5 & 5 & 25.1 & 27.0 \Tstrut\\
& 180 & 2.759 & 2.923 & \textbf{2.649} & \textbf{2.792} & 4 & 3 & 4.0 & 4.7 \\
& 300 & 2.628 & 2.784 & \textbf{2.594} & \textbf{2.717} & 3 & 3 & 1.4 & 2.4 \\
& 600 & 2.573 & 2.690 & \textbf{2.517} & \textbf{2.618} & 3 & 3 & 2.1 & 2.6 \\\hline
HG-MP-25 & 60 & 7.440 & 7.827 & \textbf{4.856} & \textbf{5.306} & 5 & 5 & 34.6 & 32.3 \Tstrut\\
& 180 & 4.661 & 5.054 & \textbf{4.206} & \textbf{4.577} & 4 & 5 & 8.6 & 8.7 \\
& 300 & 4.211 & 4.507 & \textbf{4.012} & \textbf{4.347} & 3 & 3 & 3.6 & 2.8 \\
& 600 & 3.956 & 4.173 & \textbf{3.792} & \textbf{4.145} & 3 & 2 & 3.1 & -0.1 \\\hline
HG-MP-26 & 60 & 6.836 & 7.257 & \textbf{4.635} & \textbf{4.952} & 5 & 5 & 32.0 & 31.4 \Tstrut\\
& 180 & 4.204 & 4.667 & \textbf{4.070} & \textbf{4.381} & 4 & 5 & 3.0 & 6.1 \\
& 300 & \textbf{3.846} & \textbf{4.225} & 3.889 & 4.242 & 1 & 2 & -1.1 & -0.5 \\
& 600 & \textbf{3.697} & \textbf{3.965} & 3.743 & 4.079 & 2 & 2 & -1.2 & -3.0 \\\hline
HG-MP-27 & 60 & 6.649 & 7.027 & \textbf{4.206} & \textbf{4.477} & 5 & 5 & 36.6 & 36.1 \Tstrut\\
& 180 & 4.164 & 4.471 & \textbf{3.791} & \textbf{4.015} & 5 & 5 & 8.5 & 9.8 \\
& 300 & 3.799 & 4.140 & \textbf{3.720} & \textbf{3.905} & 3 & 3 & 1.8 & 5.5 \\
& 600 & \textbf{3.665} & 3.926 & 3.679 & \textbf{3.808} & 3 & 3 & -0.6 & 3.0 \\\hline
HG-MP-28 & 60 & 10.518 & 11.365 & \textbf{6.917} & \textbf{7.472} & 5 & 5 & 33.9 & 33.9 \Tstrut\\
& 180 & 6.212 & 6.598 & \textbf{5.102} & \textbf{5.563} & 5 & 5 & 17.9 & 15.5 \\
& 300 & 5.356 & 5.705 & \textbf{4.768} & \textbf{5.260} & 5 & 5 & 11.0 & 7.6 \\
& 600 & 4.838 & 5.141 & \textbf{4.670} & \textbf{5.047} & 5 & 4 & 3.5 & 1.9 \\\hline
HG-MP-29 & 60 & 11.624 & 12.385 & \textbf{7.162} & \textbf{7.679} & 5 & 5 & 38.4 & 38.1 \Tstrut\\
& 180 & 6.229 & 6.974 & \textbf{5.555} & \textbf{6.091} & 5 & 5 & 10.7 & 12.5 \\
& 300 & 5.444 & 5.845 & \textbf{5.304} & \textbf{5.710} & 3 & 3 & 2.5 & 2.3 \\
& 600 & \textbf{5.023} & \textbf{5.325} & 5.104 & 5.447 & 2 & 2 & -1.6 & -2.2 \\\hline
HG-MP-30 & 60 & 9.557 & 10.011 & \textbf{5.764} & \textbf{6.068} & 5 & 5 & 39.7 & 39.3 \Tstrut\\
& 180 & 5.613 & 6.097 & \textbf{4.562} & \textbf{5.096} & 5 & 5 & 18.5 & 16.0 \\
& 300 & 4.999 & 5.463 & \textbf{4.334} & \textbf{4.776} & 5 & 5 & 12.9 & 12.1 \\
& 600 & 4.557 & 4.992 & \textbf{4.215} & \textbf{4.474} & 5 & 5 & 7.4 & 10.0 \\\hline
Total &  &  &  &  &  & 228/300 & 240/300 &  & \Tstrut\\
Average (\%) &  &  &  &  &  &  &  & 8.4 & 9.3 \\
\hline
\end{tabular}}
\end{table}
%\end{threeparttable}\end{adjustwidth}

\newpage

\subsection{Graphical summary}
\label{sec:overviewresult}
\begin{figure}[H]
     \centering
     \begin{subfigure}[b]{0.45\textwidth}
        \centering
        \includegraphics[width=\textwidth]{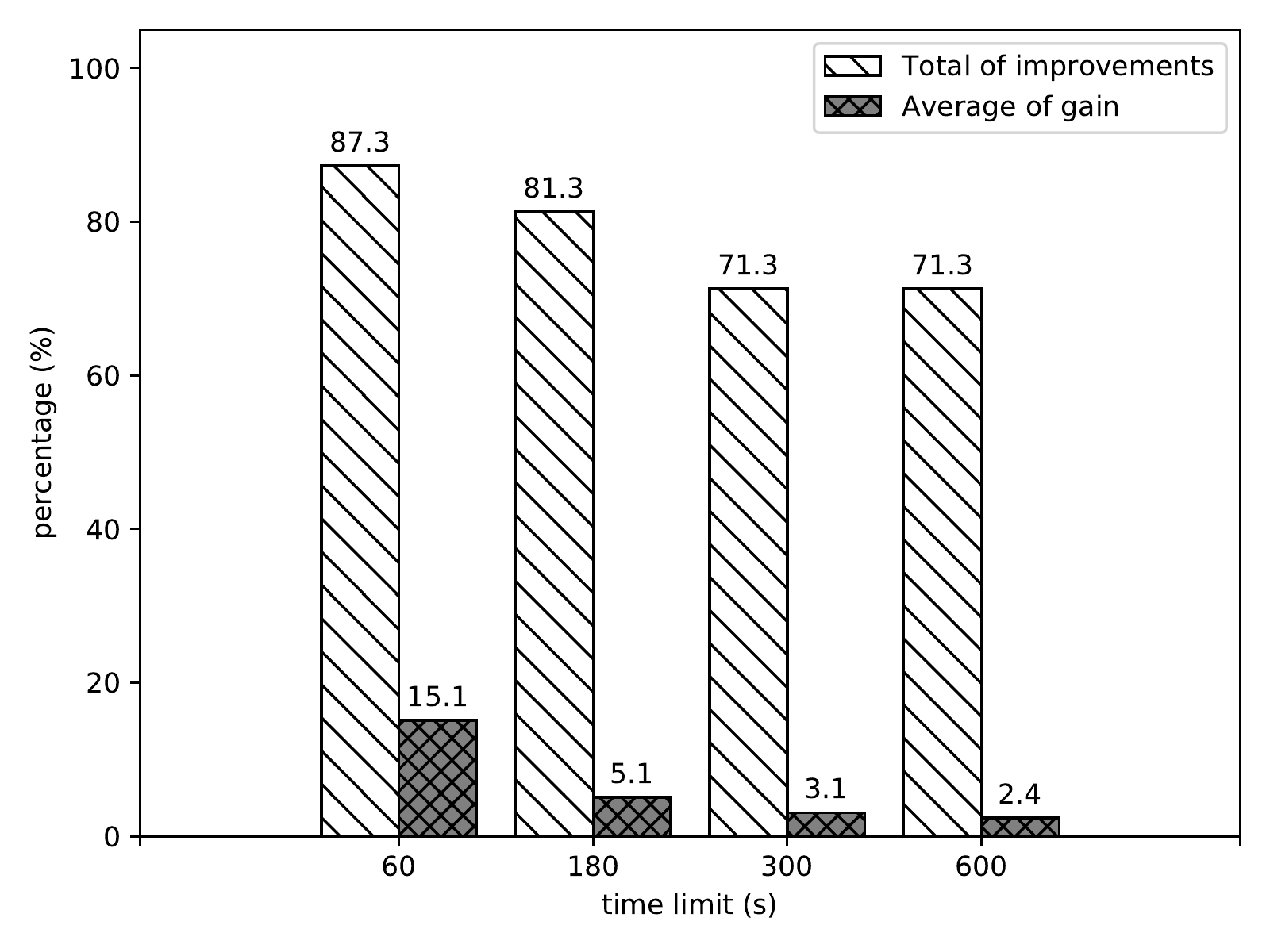}
        %\caption{}
        \caption{Average cost considering all instances}
        \label{fig:averageOverall}
    \end{subfigure}
    \hfill
    \begin{subfigure}[b]{0.45\textwidth}
        \centering
        \includegraphics[width=\textwidth]{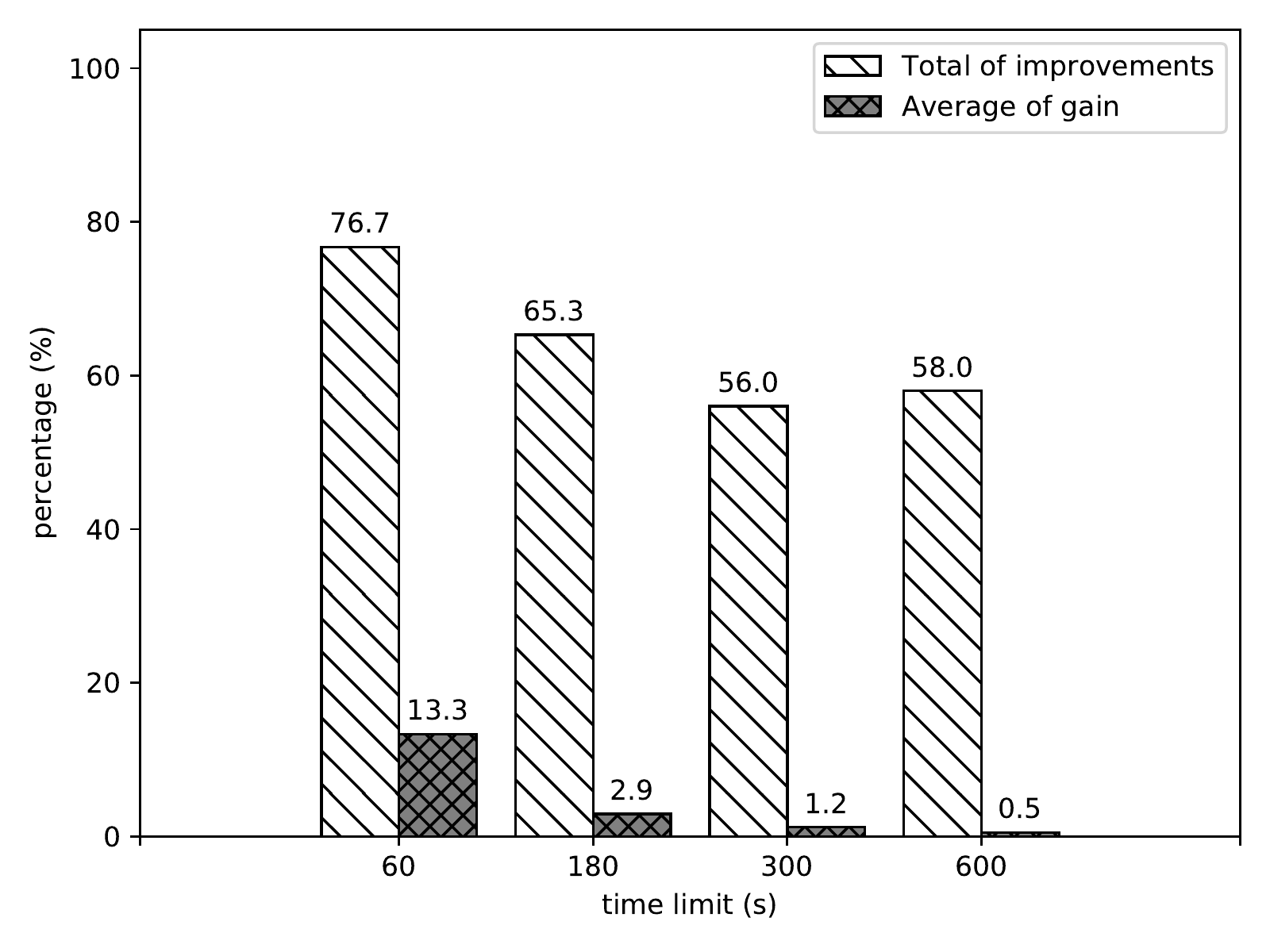}
        %\caption{}
        \caption{Best cost considering all instances}
        \label{fig:bestOverall}
     \end{subfigure}
     \hfill
     \begin{subfigure}[b]{0.45\textwidth}
        \centering
        \includegraphics[width=\textwidth]{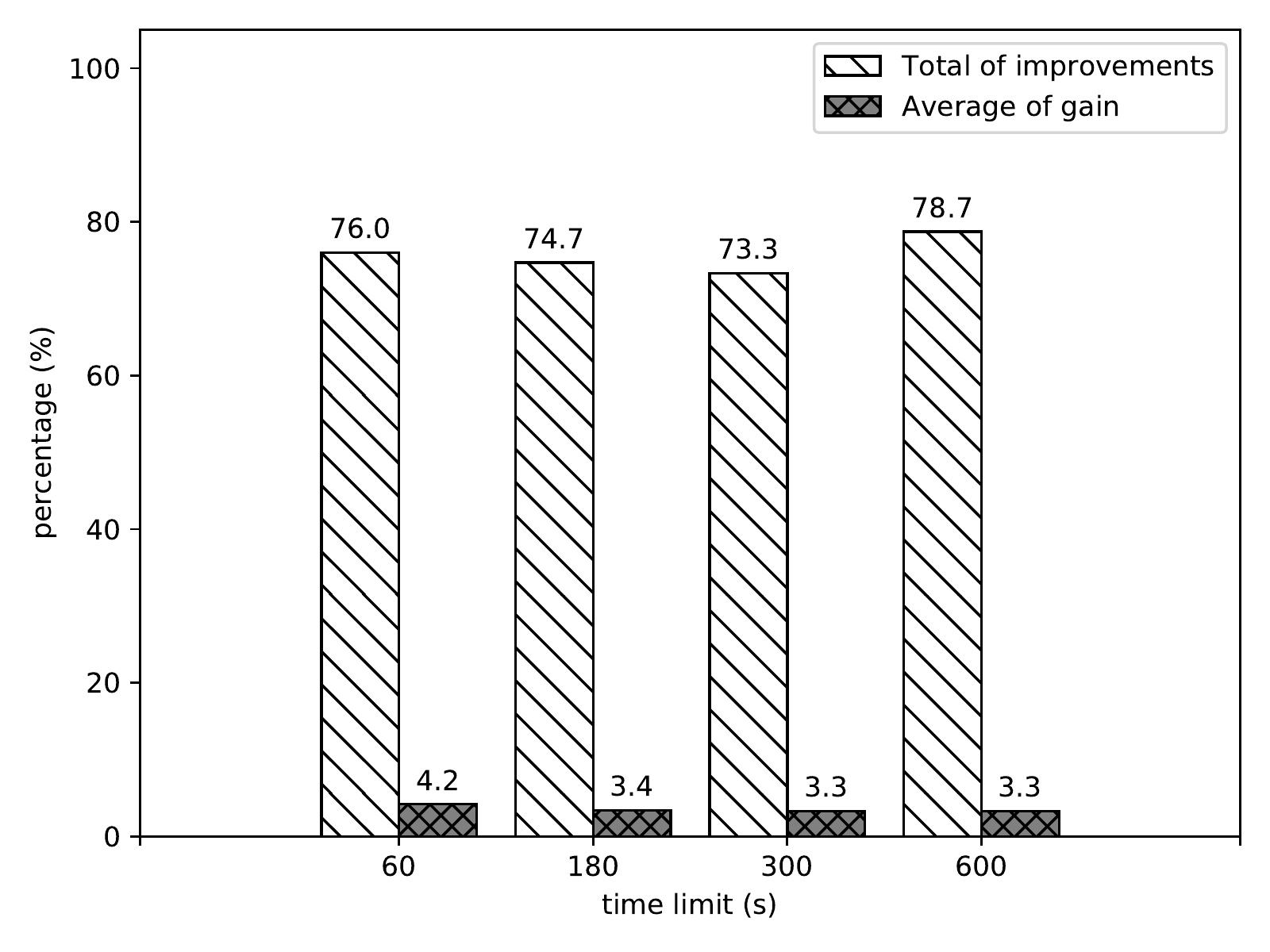}
        %\caption{}
        \caption{Average cost considering the instances with less than 500 pipes}
        \label{fig:averageSmallInstances}
    \end{subfigure}
    \hfill
    \begin{subfigure}[b]{0.45\textwidth}
        \centering
        \includegraphics[width=\textwidth]{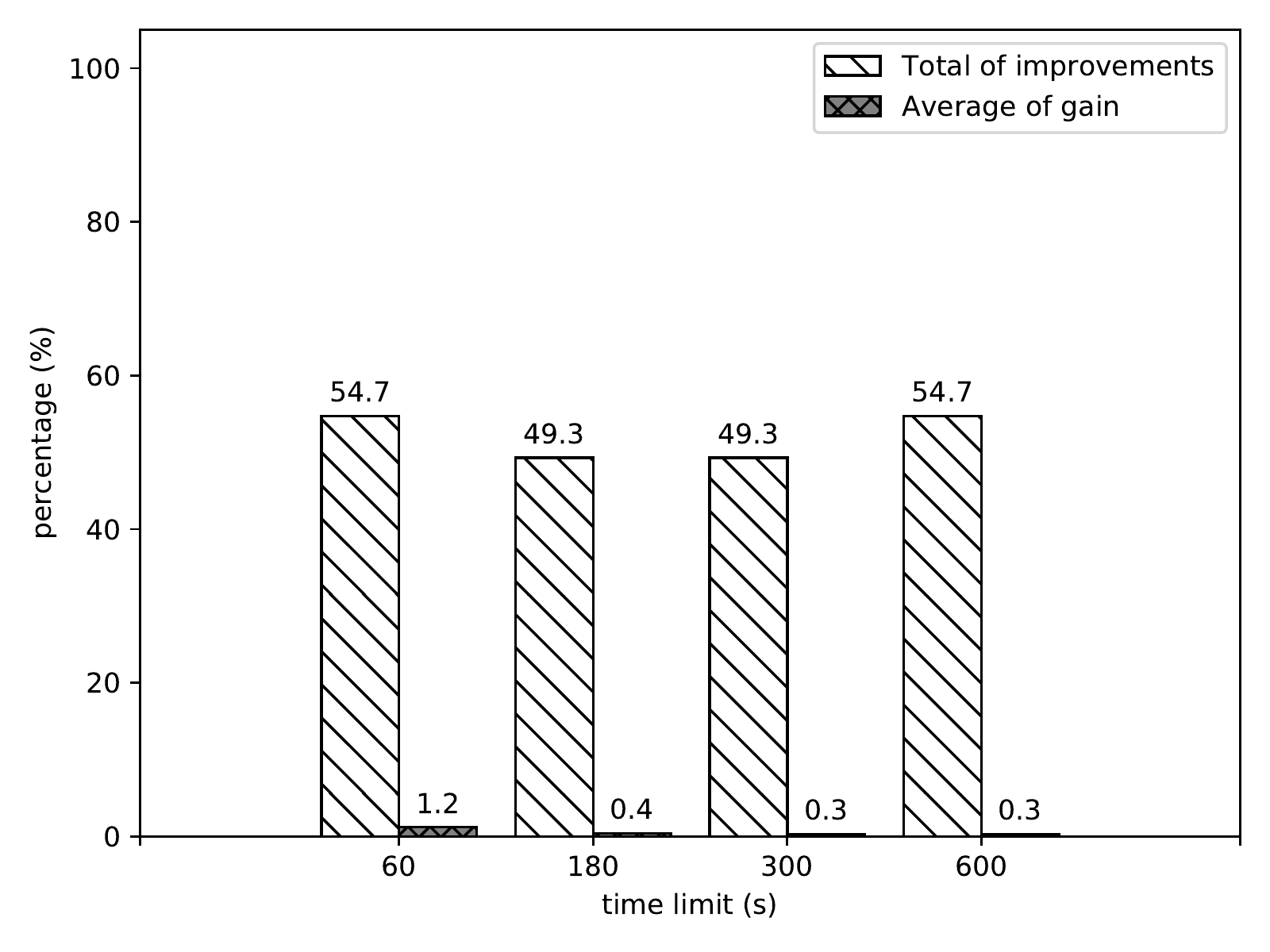}
        %\caption{}
        \caption{Best cost considering the instances with less than 500 pipes}
        \label{fig:bestSmallInstances}
    \end{subfigure}
   \hfill
     \begin{subfigure}[b]{0.45\textwidth}
        \centering
        \includegraphics[width=\textwidth]{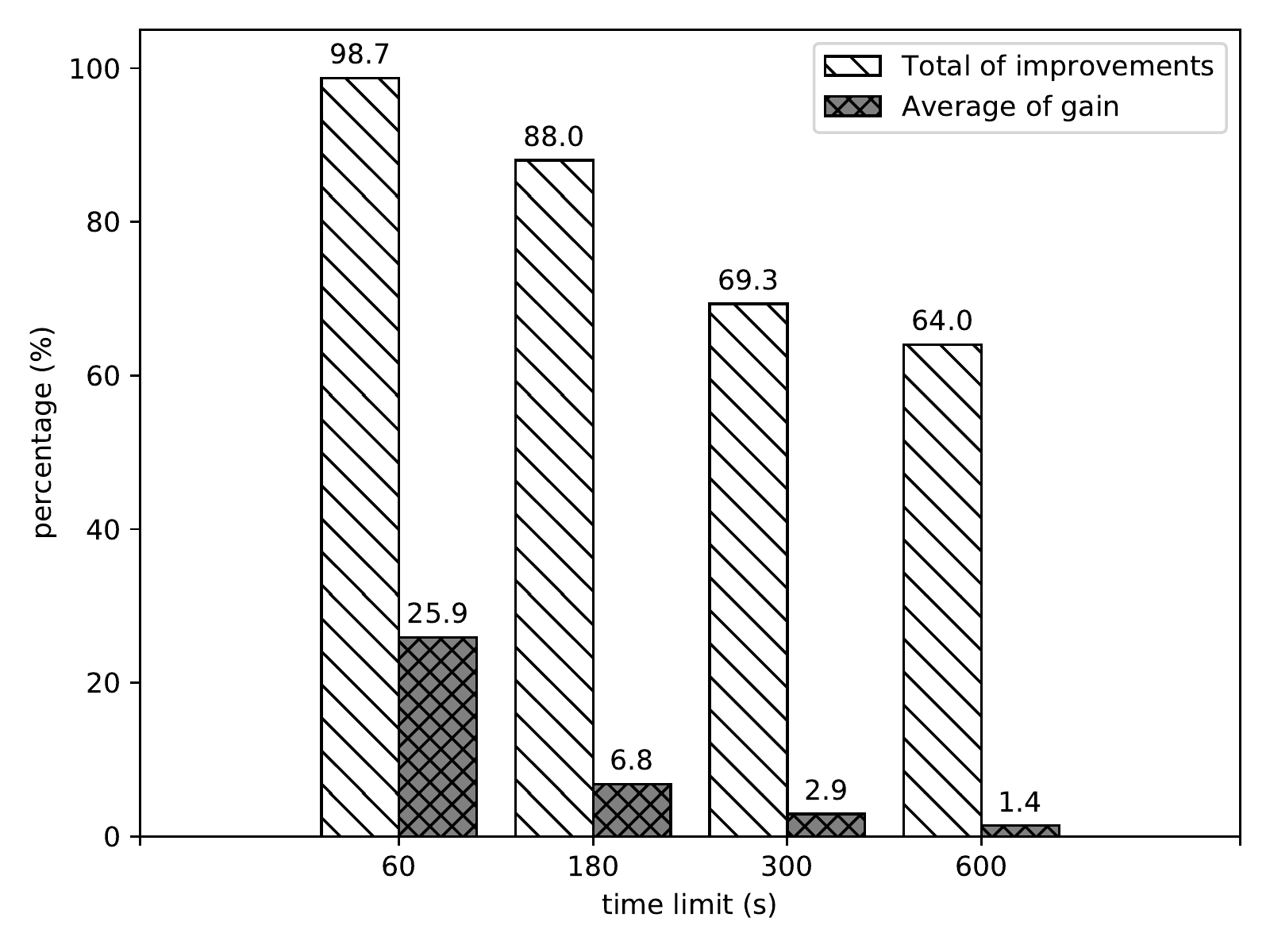}
        %\caption{}
        \caption{Average cost considering the instances with more than 500 pipes}
        \label{fig:averageBigInstances}
    \end{subfigure}
    \hfill
    \begin{subfigure}[b]{0.45\textwidth}
        \centering
        \includegraphics[width=\textwidth]{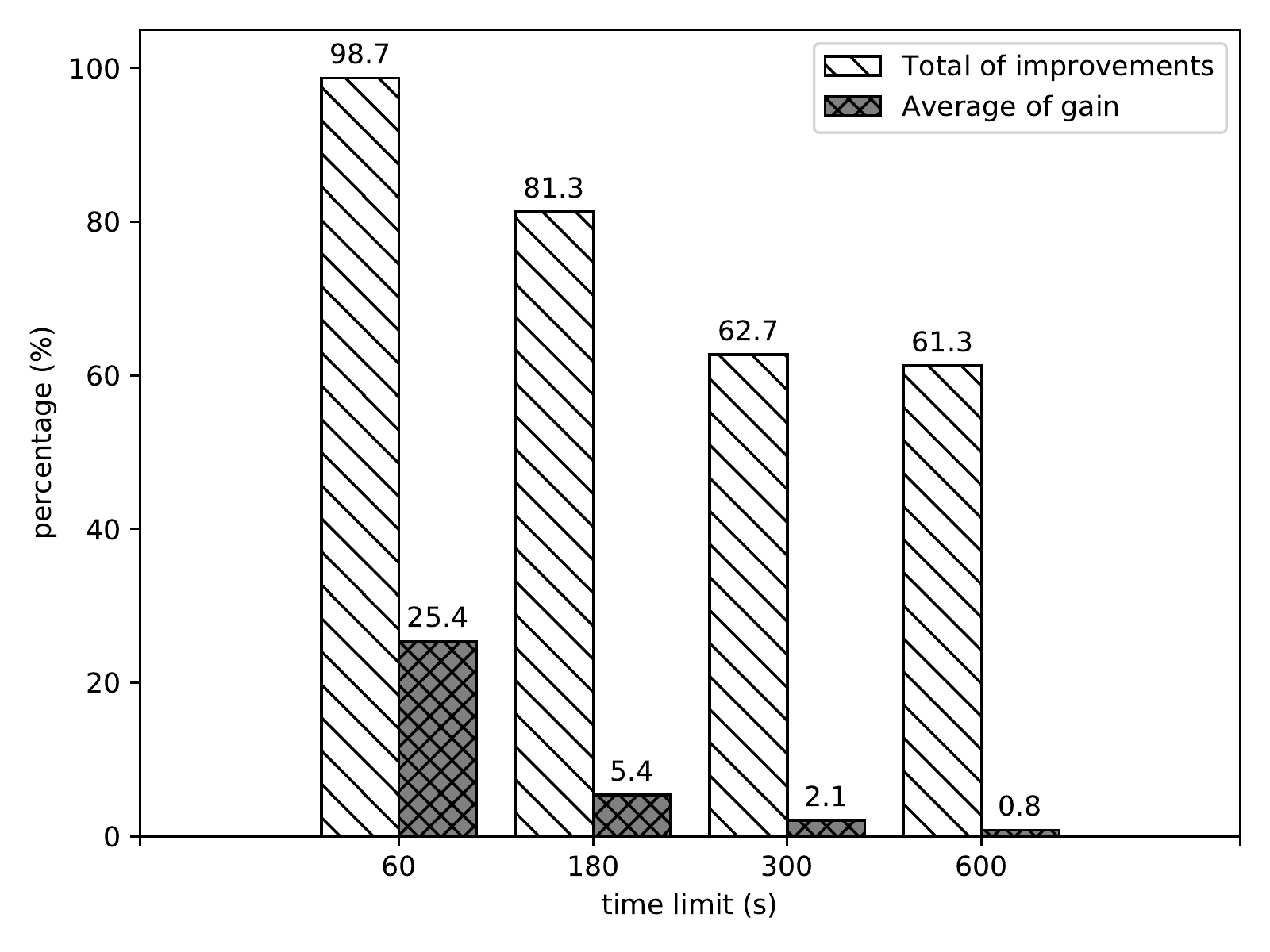}
        %\caption{}
        \caption{Best cost considering the instances with more than 500 pipes}
        \label{fig:bestBigInstances}
    \end{subfigure}
   \hfill
    \caption{Graphical summary of the improvements achieved by ILS+ over ILS. The subfigures show the percentage of the instances for which ILS+ improved ILS, as well as the average gains of ILS+ over ILS. 
    Subfigures (a) and (b) consider the average and best costs for all instances; Subfigures (c) and (d) examine the average and best costs for instances with less than 500 pipes; Subfigures (e) and (f) contemplate instances with more than 500 pipes.}
    \label{fig:summary}
\end{figure}

A graphical summary of the computational results is depicted in Figure~\ref{fig:summary}. The figure illustrates the improvements achieved by ILS+ over ILS, considering the individual instances rather than instance groups. 

Considering all the 150 instances, Figure \ref{fig:averageOverall} shows that ILS+ improved the average solutions achieved by ILS for over 71.3\% of them for all the tested time limits, and such improvement is more evidenced for the time limit of 60 seconds, for which improved solutions were obtained for 87.3\% of the instances. 
The figure also shows that the average gain is considerably larger for the more restricted time limit of 60 seconds (15.1\%) and progressively decreases as the time limit increases.
Figure \ref{fig:bestOverall} shows that ILS+ encountered improved best solutions when compared to ILS for over 56.0\% for all the considered time limits. ILS+ encountered improved best solutions for 76.7\% of the instances when the time limit was set to 60 seconds. Similarly to what was observed for the average solutions, it can be seen that the average gain is considerably larger for the more restricted time limit (13.3\%) and progressively decreases with the increase of the time limit.

Considering the 75 instances with less than 500 pipes, Figure \ref{fig:averageSmallInstances} shows that ILS+ improved the average solutions achieved by ILS for over 73.3\% of them with a similar behavior for all the tested time limits, for which improved solutions were obtained for percentages of the instances varying between 73.3\% and 78.7\%.
The figure also shows that the obtained average gains were between 3.3\% and 4.2\% for all the tested time limits.
Figure \ref{fig:bestSmallInstances} shows that ILS+ encountered a very similar performance when compared to ILS. ILS+ encountered improved best solutions for 54.7\% of the instances for the time limits of 60 and 600 seconds. However, for the time limits of 180 and 300 seconds ILS+ encountered improved best solutions for just 49.3\%. Further, there is no considerably gain for any of the time limits, for which the average gains were obtained around 0.55\%.

Considering the 75 instances with more than 500 pipes,  Figure \ref{fig:averageBigInstances} shows that ILS+ improved the average solutions achieved by ILS for over 64.0\% of them for all the tested time limits, and such improvement is more evident for the time limit of 60 seconds, for which improved solutions were obtained for 98.7\% of the instances. 
The figure also shows that the average gain is considerably larger for the more restricted time limit of 60 seconds (25.9\%) and progressively decreases as the time limit increases.
Figure \ref{fig:bestBigInstances} shows that ILS+ encountered improved best solutions when compared to ILS for over 61.3\% for all the considered time limits. It is remarkable the fact that ILS+ obtained improved best solutions for 98.7\% of the instances when the time limit was set to 60 seconds. Similarly to what was observed for the average solutions, it can be seen that the average gain is considerably larger for the more restricted time limit (25.4\%) and progressively decreases with the increase of the time limit.

\subsection{Analyzing the impact of each individual novelty}
\label{sec:performanceindividualimprovements}

In this section, we analyze the impact of each of the proposed novelties individually on the obtained solution values and compare these with ILS and ILS+. 
Each of the tested novelties is defined as using a single improvement on top of ILS, namely: the aggressive reduction (ILS-REDU), the pool of solutions (ILS-POOL), the new perturbations (ILS-PERT), and the shortest path-based local search (ILS-SPT). 
We consider a subset of the instances described in Subsection~\ref{sec:benchmark}, containing 10\% of them, chosen randomly from varying instance groups. The settings are defined as in Subsection~\ref{sec:test}.

\begin{figure}[H]
\centering
\begin{subfigure}[b]{0.45\textwidth}
\centering
\includegraphics[width=\textwidth]{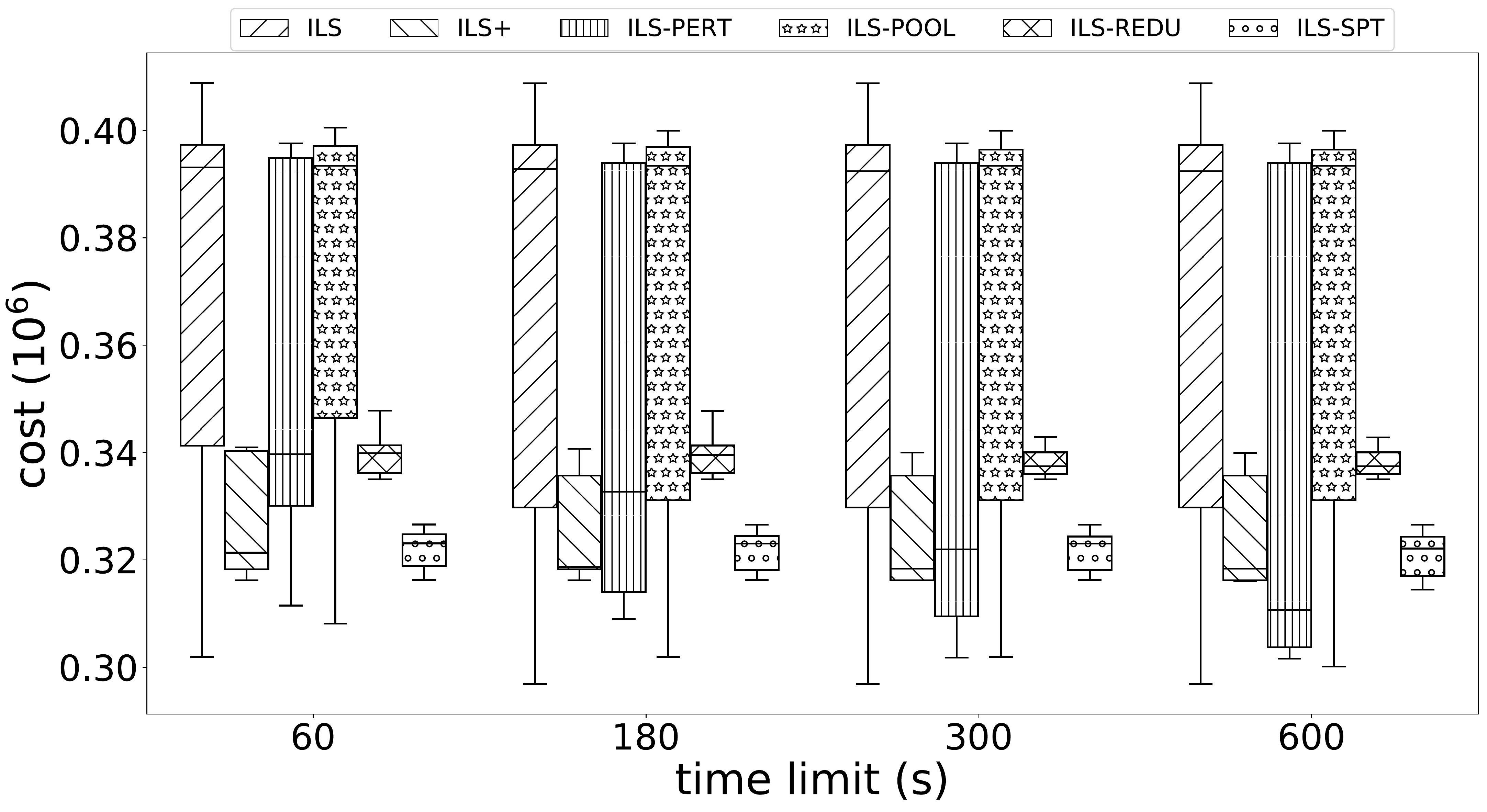}
\caption{HG-MP-2-1}
%\label{fig:LABEL}
\end{subfigure}
\hfill
\centering
\begin{subfigure}[b]{0.45\textwidth}
\centering
\includegraphics[width=\textwidth]{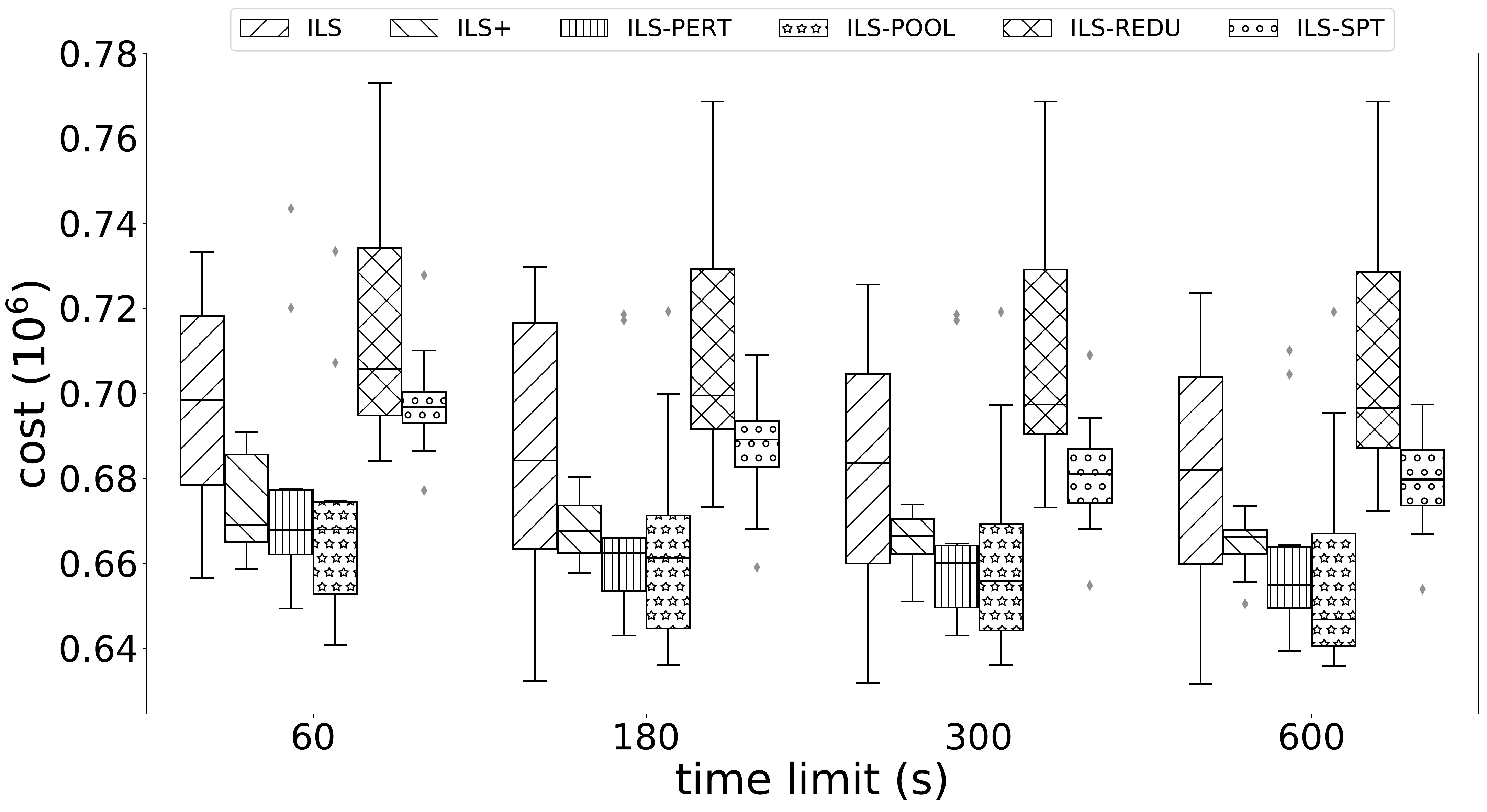}
\caption{HG-MP-4-3}
%\label{fig:LABEL}
\end{subfigure}
\hfill
\centering
\begin{subfigure}[b]{0.45\textwidth}
\centering
\includegraphics[width=\textwidth]{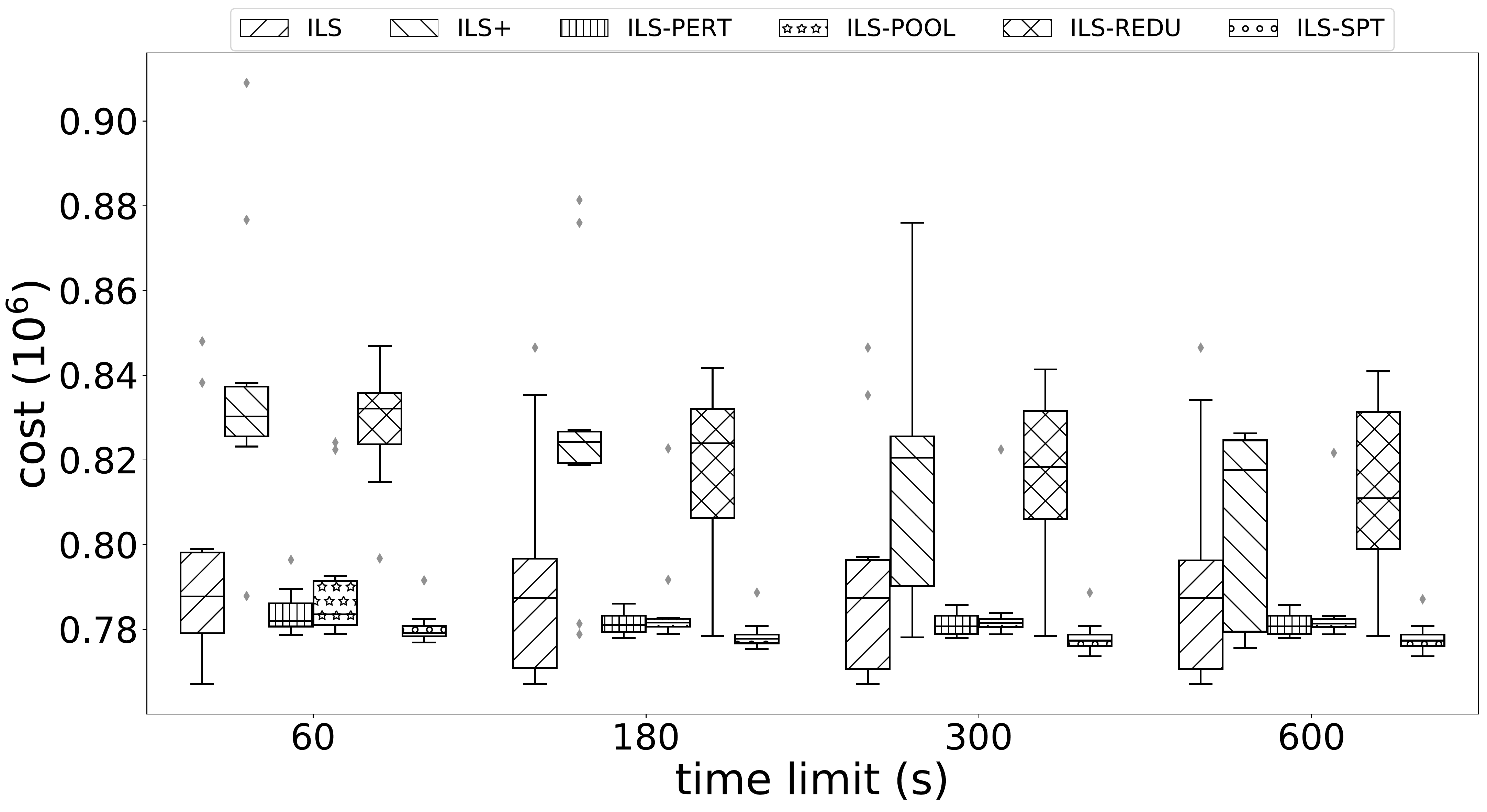}
\caption{HG-MP-6-5}
%\label{fig:LABEL}
\end{subfigure}
\hfill
\centering
\begin{subfigure}[b]{0.45\textwidth}
\centering
\includegraphics[width=\textwidth]{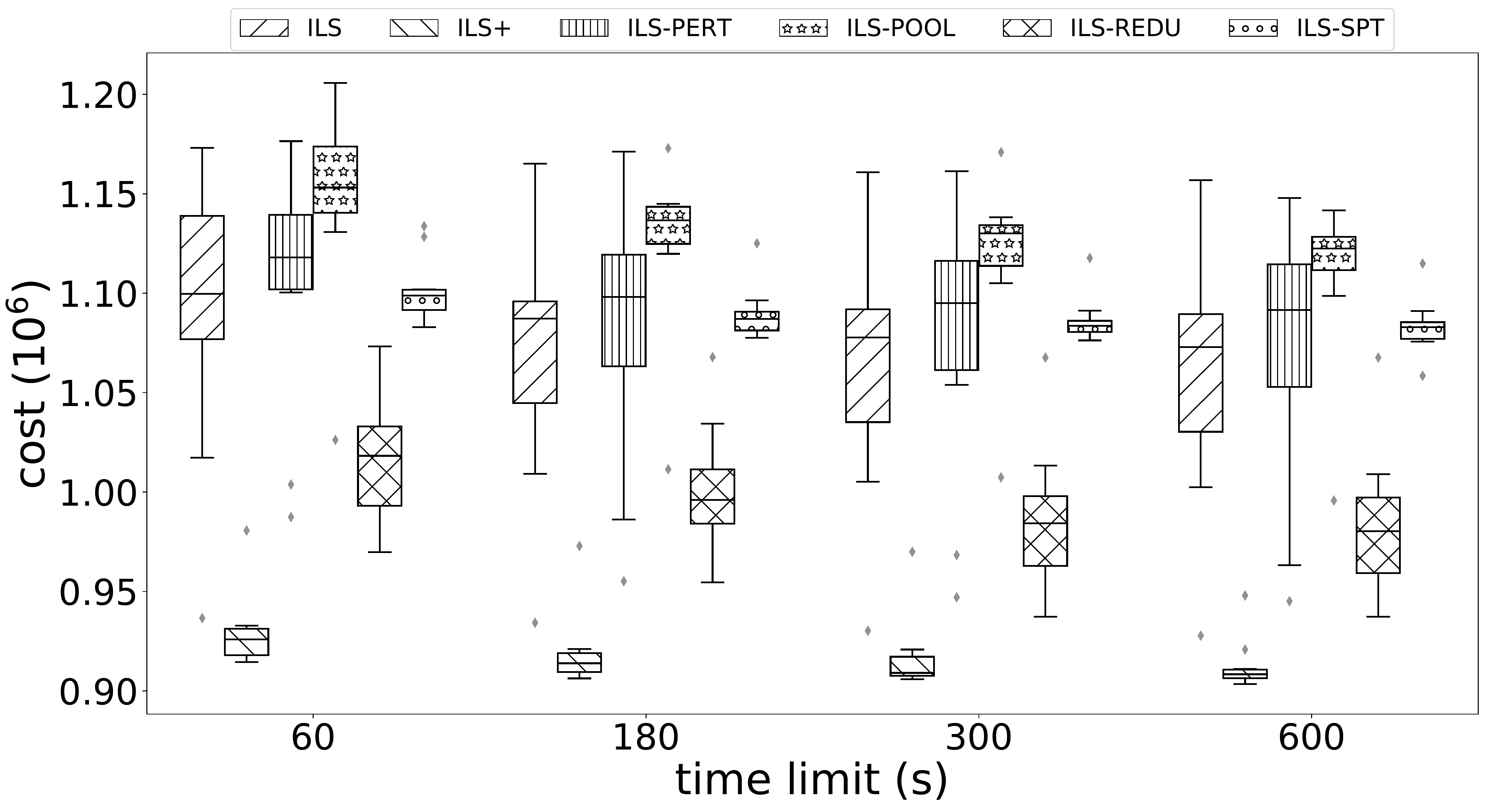}
\caption{HG-MP-8-5}
%\label{fig:LABEL}
\end{subfigure}
\hfill
\centering
\begin{subfigure}[b]{0.45\textwidth}
\centering
\includegraphics[width=\textwidth]{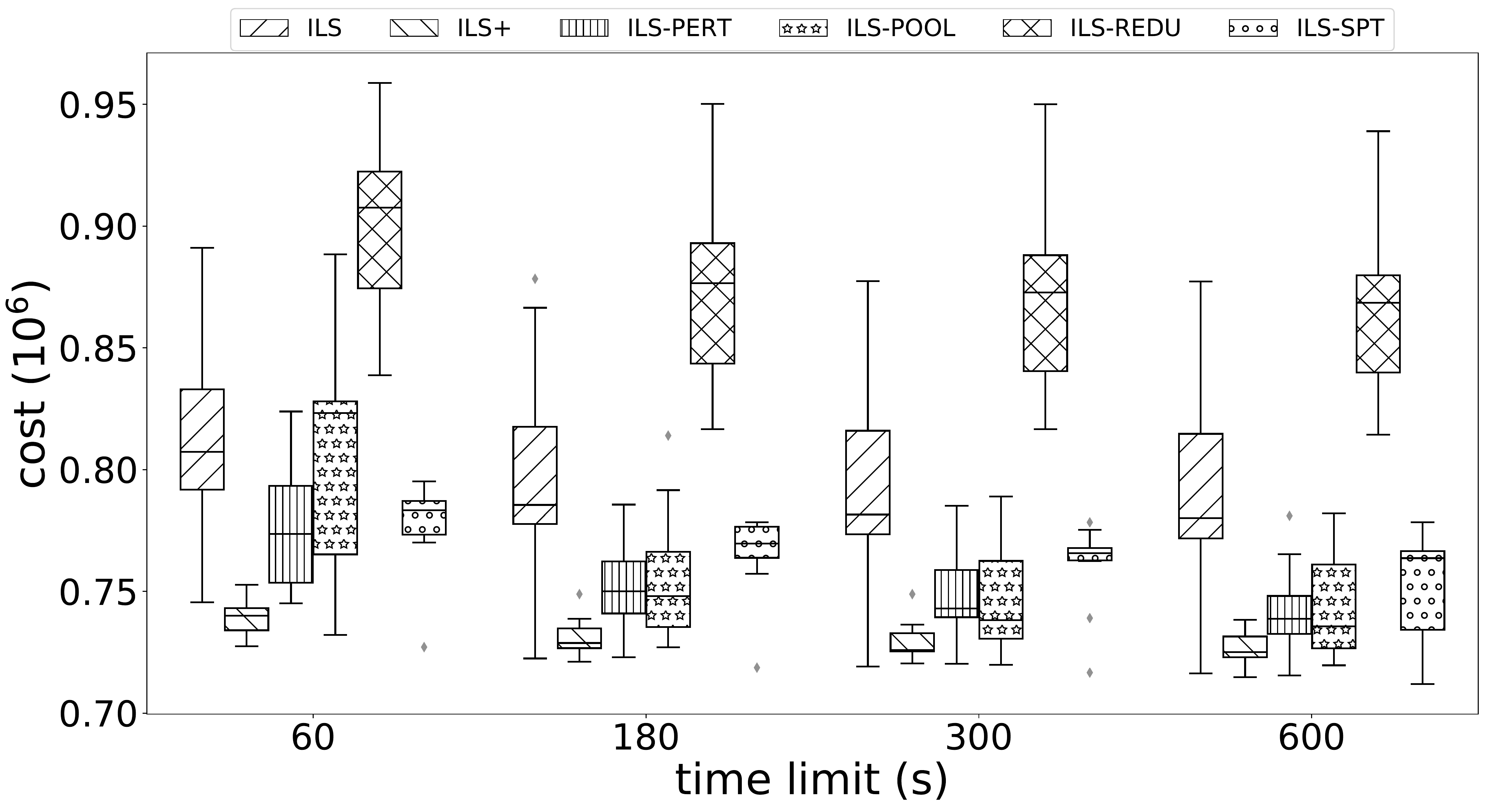}
\caption{HG-MP-10-4}
%\label{fig:LABEL}
\end{subfigure}
\hfill
\centering
\begin{subfigure}[b]{0.45\textwidth}
\centering
\includegraphics[width=\textwidth]{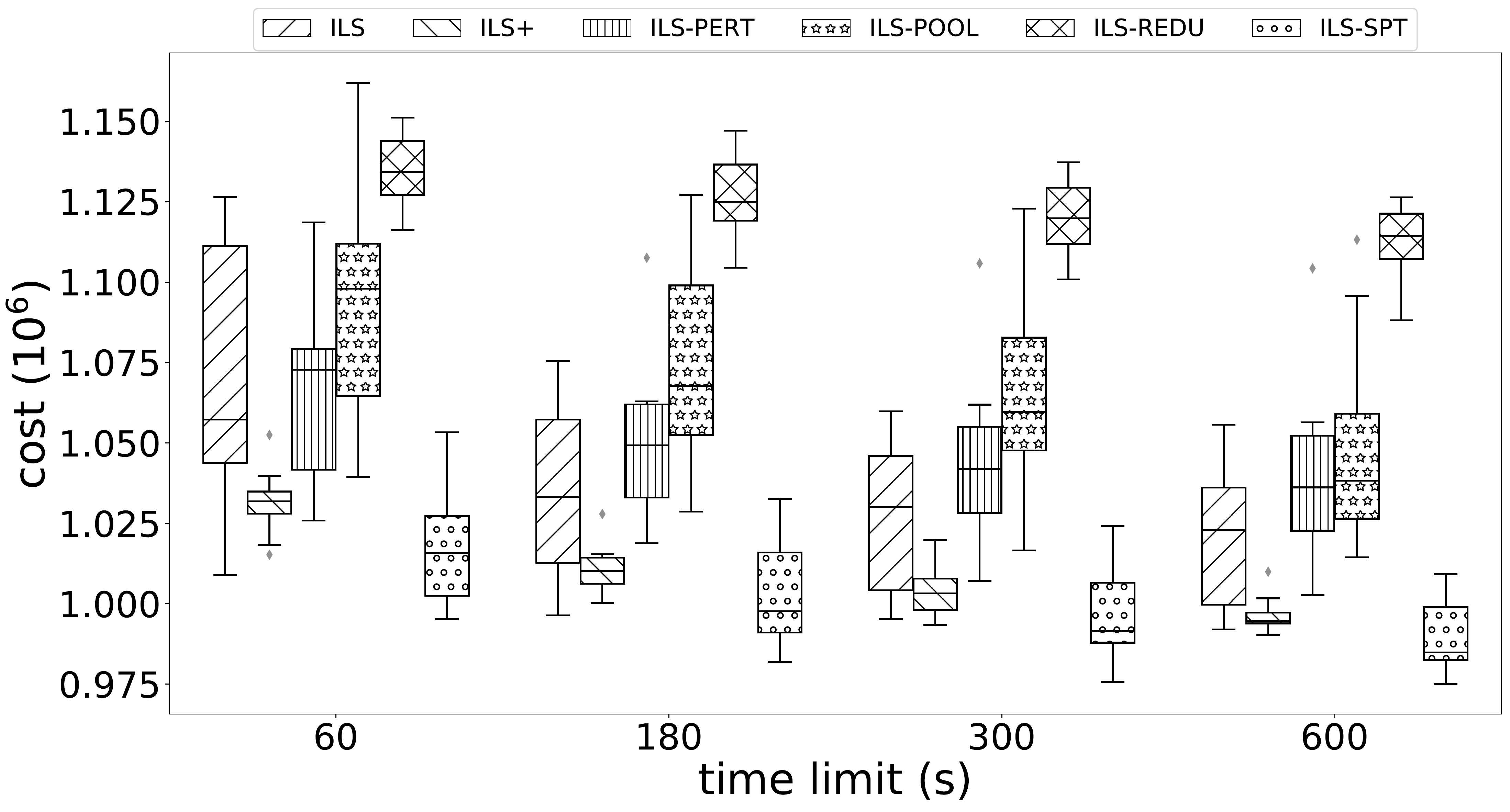}
\caption{HG-MP-12-4}
%\label{fig:LABEL}
\end{subfigure}
\hfill
\centering
\begin{subfigure}[b]{0.45\textwidth}
\centering
\includegraphics[width=\textwidth]{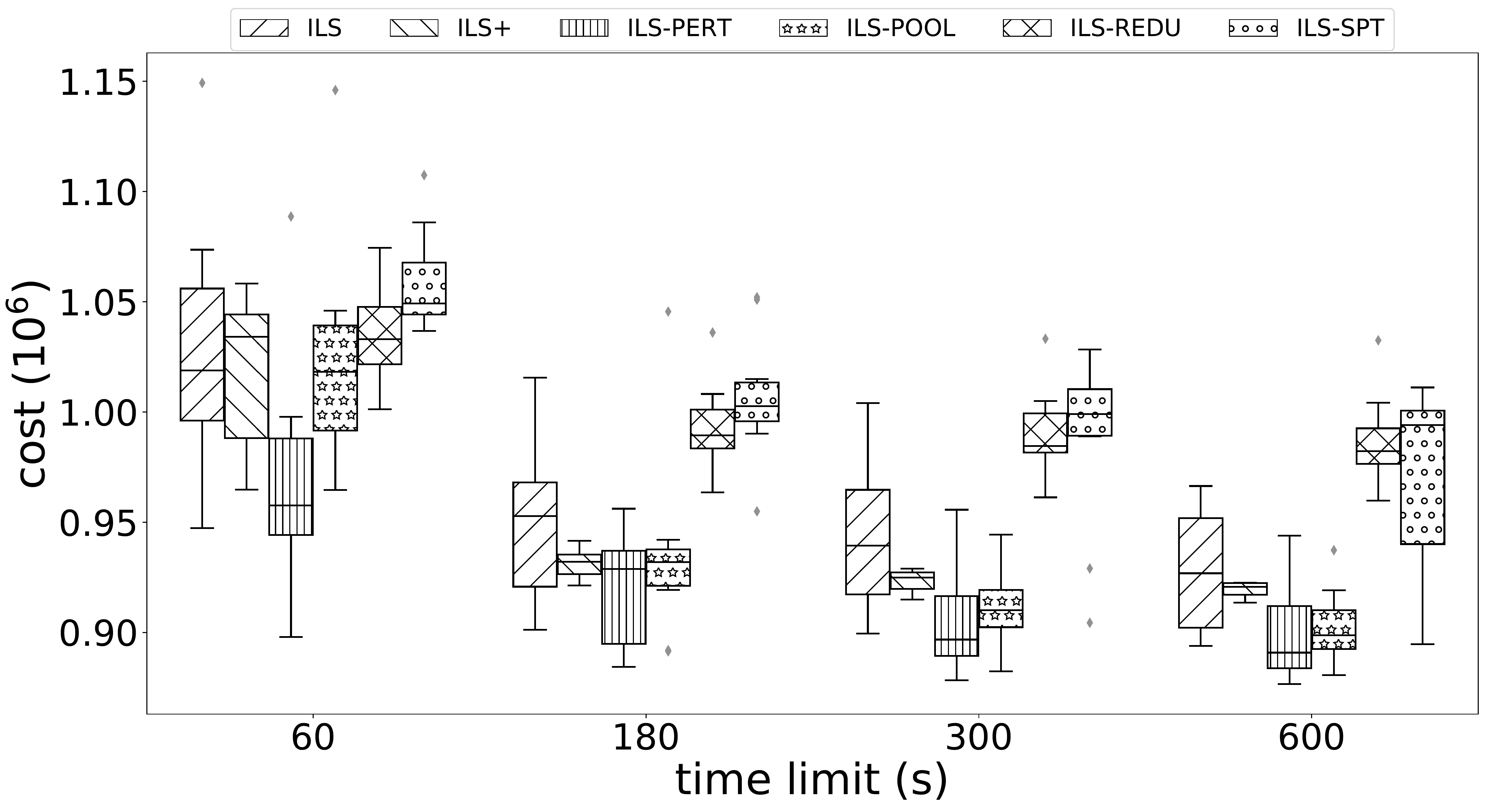}
\caption{HG-MP-14-5}
%\label{fig:LABEL}
\end{subfigure}
\hfill
    \caption{\rafaelC{Impact of the individual novelties on the obtained solution costs considering instances with less than 500 pipes.}}
    \label{fig:new_tests1}
\end{figure}

\begin{figure}[H]
\centering
\begin{subfigure}[b]{0.45\textwidth}
\centering
\includegraphics[width=\textwidth]{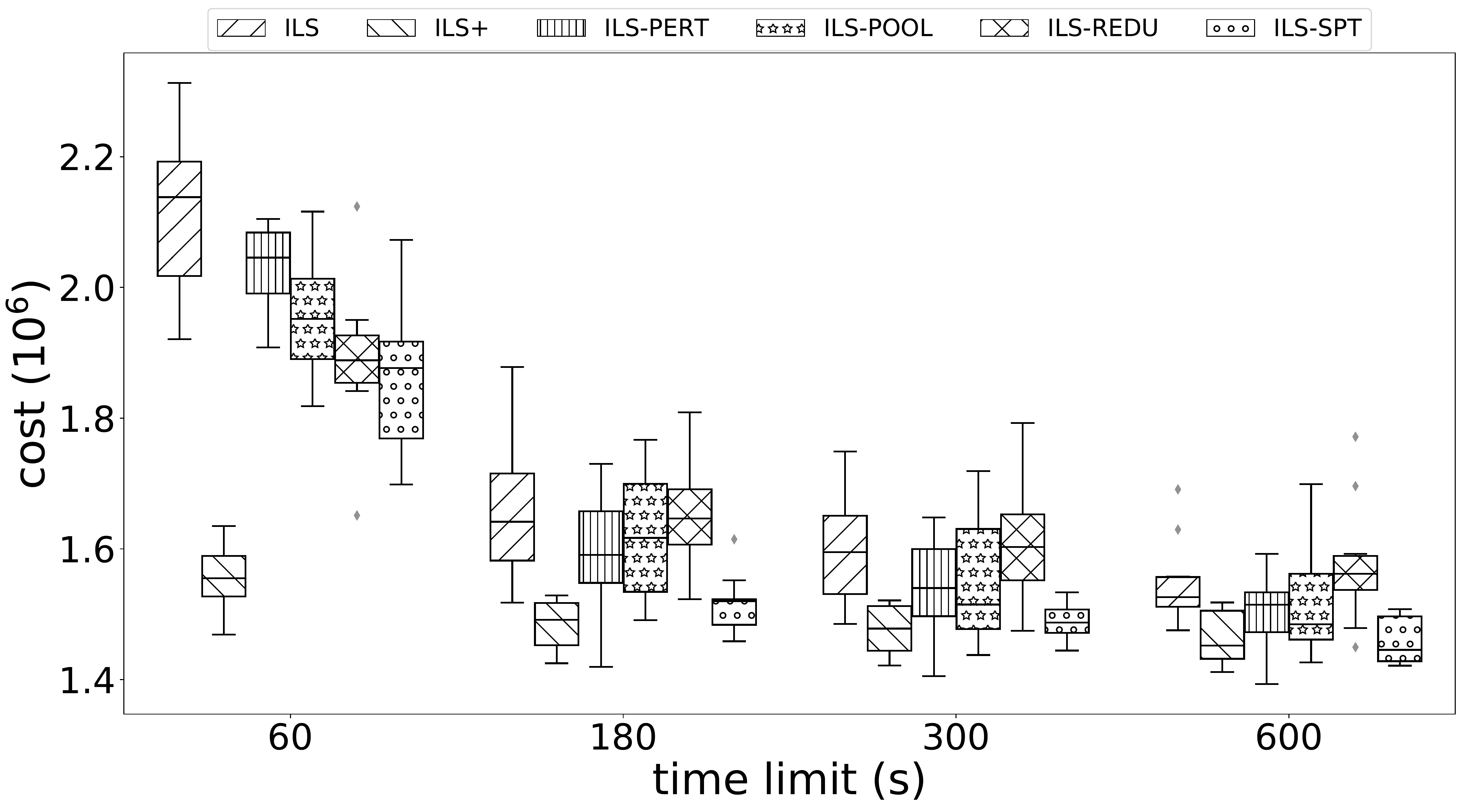}
\caption{HG-MP-16-1}
%\label{fig:LABEL}
\end{subfigure}
\hfill
\centering
\begin{subfigure}[b]{0.45\textwidth}
\centering
\includegraphics[width=\textwidth]{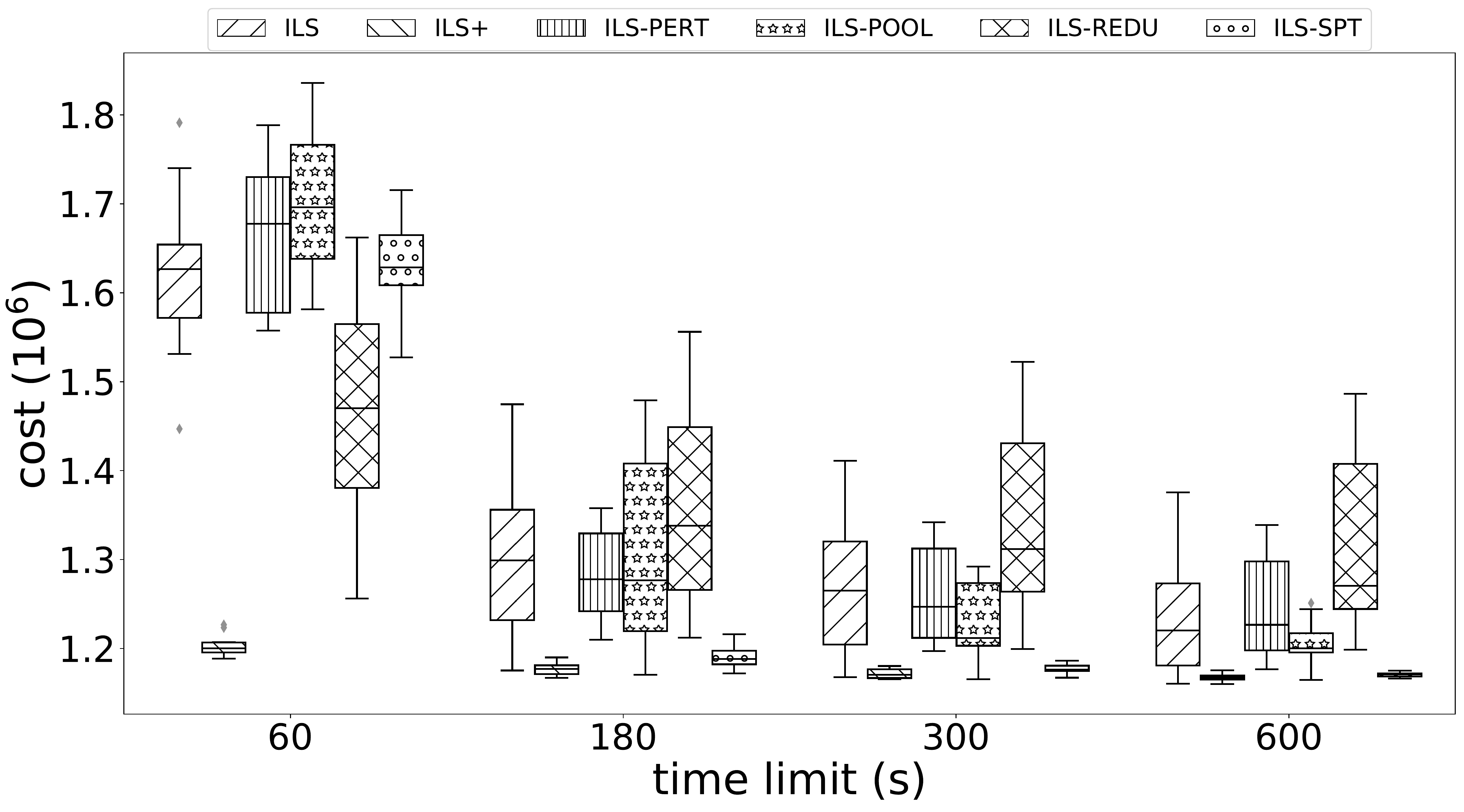}
\caption{HG-MP-18-2}
%\label{fig:LABEL}
\end{subfigure}
\hfill
\centering
\begin{subfigure}[b]{0.45\textwidth}
\centering
\includegraphics[width=\textwidth]{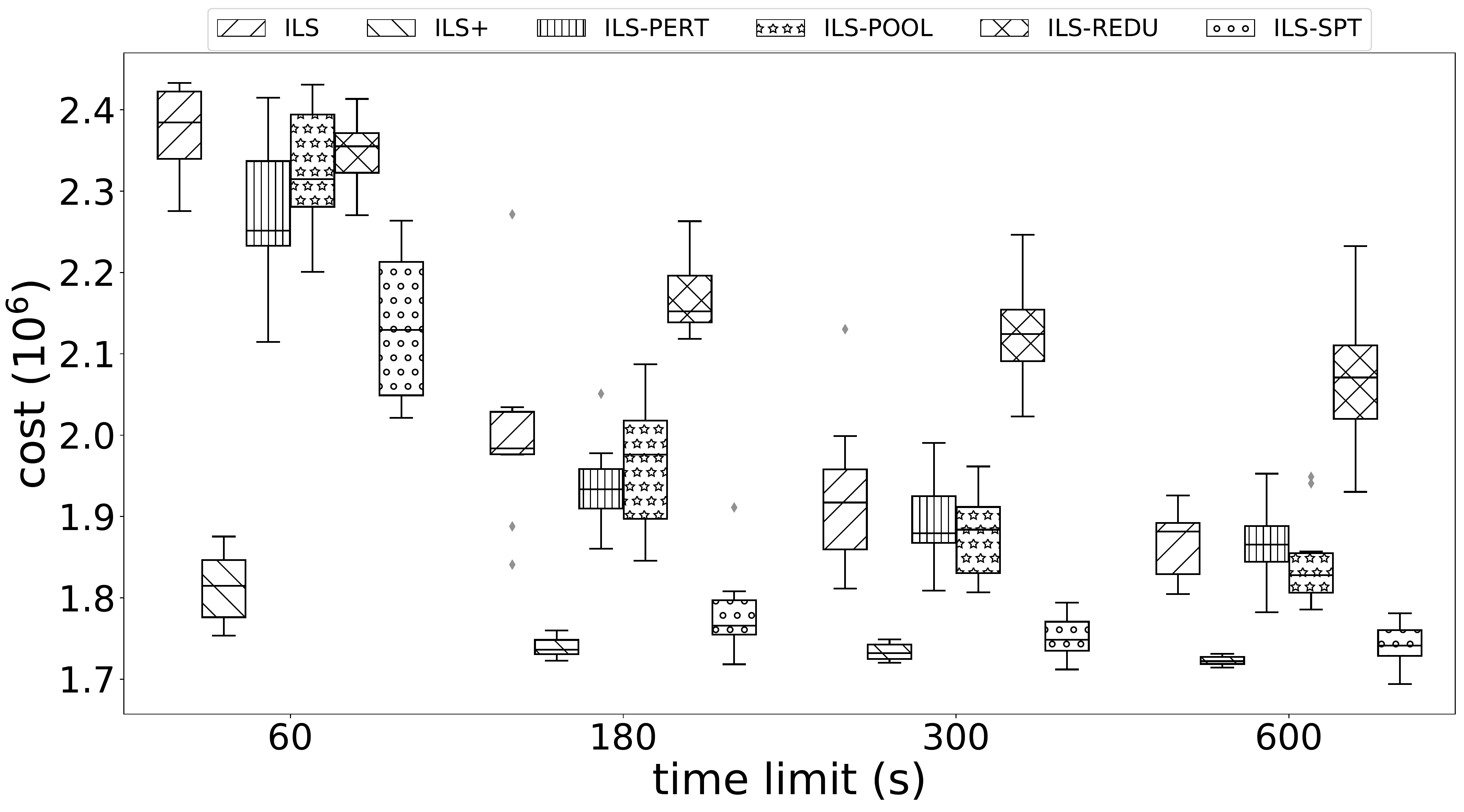}
\caption{HG-MP-20-2}
%\label{fig:LABEL}
\end{subfigure}
\hfill
\centering
\begin{subfigure}[b]{0.45\textwidth}
\centering
\includegraphics[width=\textwidth]{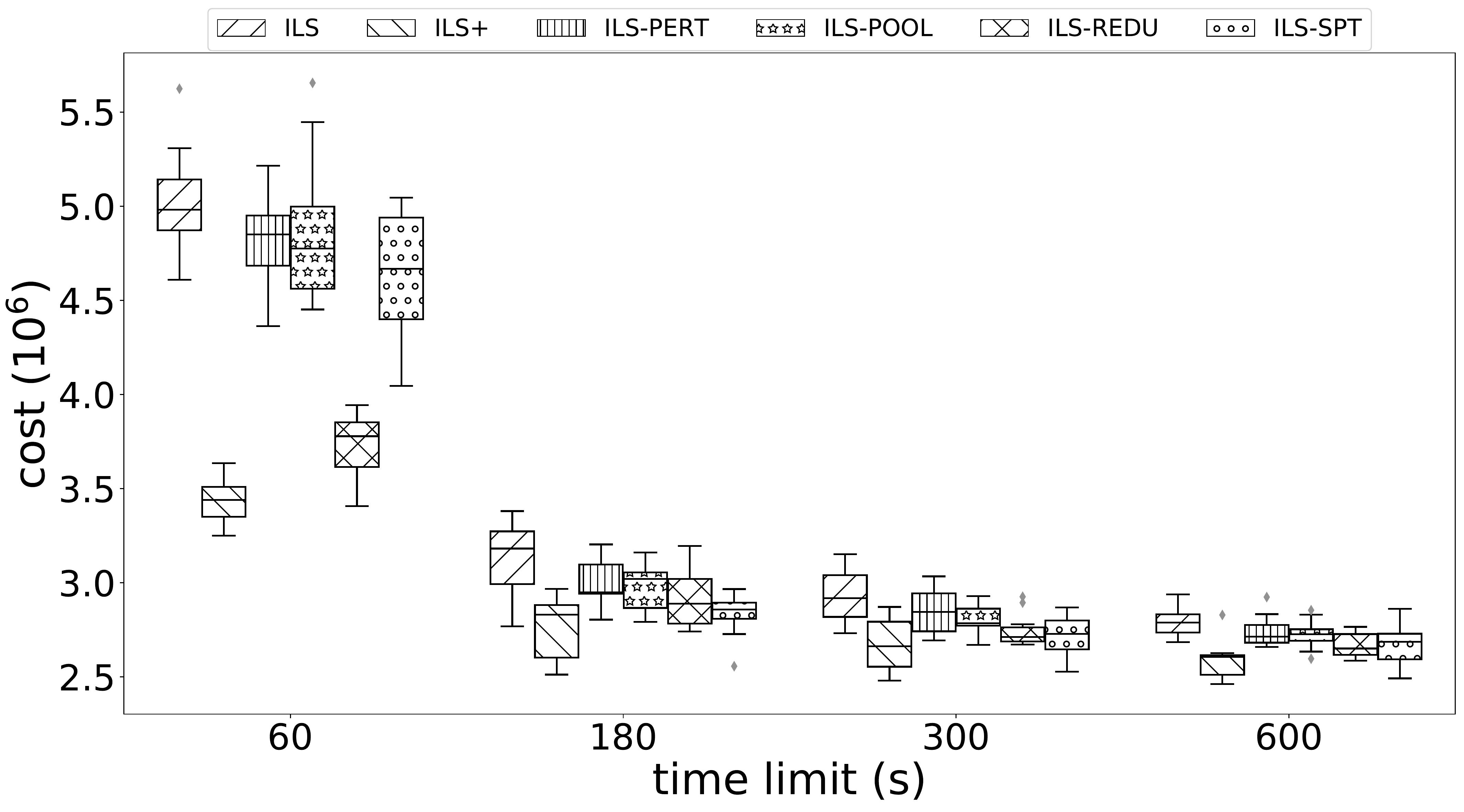}
\caption{HG-MP-22-2}
%\label{fig:LABEL}
\end{subfigure}
\hfill
\centering
\begin{subfigure}[b]{0.45\textwidth}
\centering
\includegraphics[width=\textwidth]{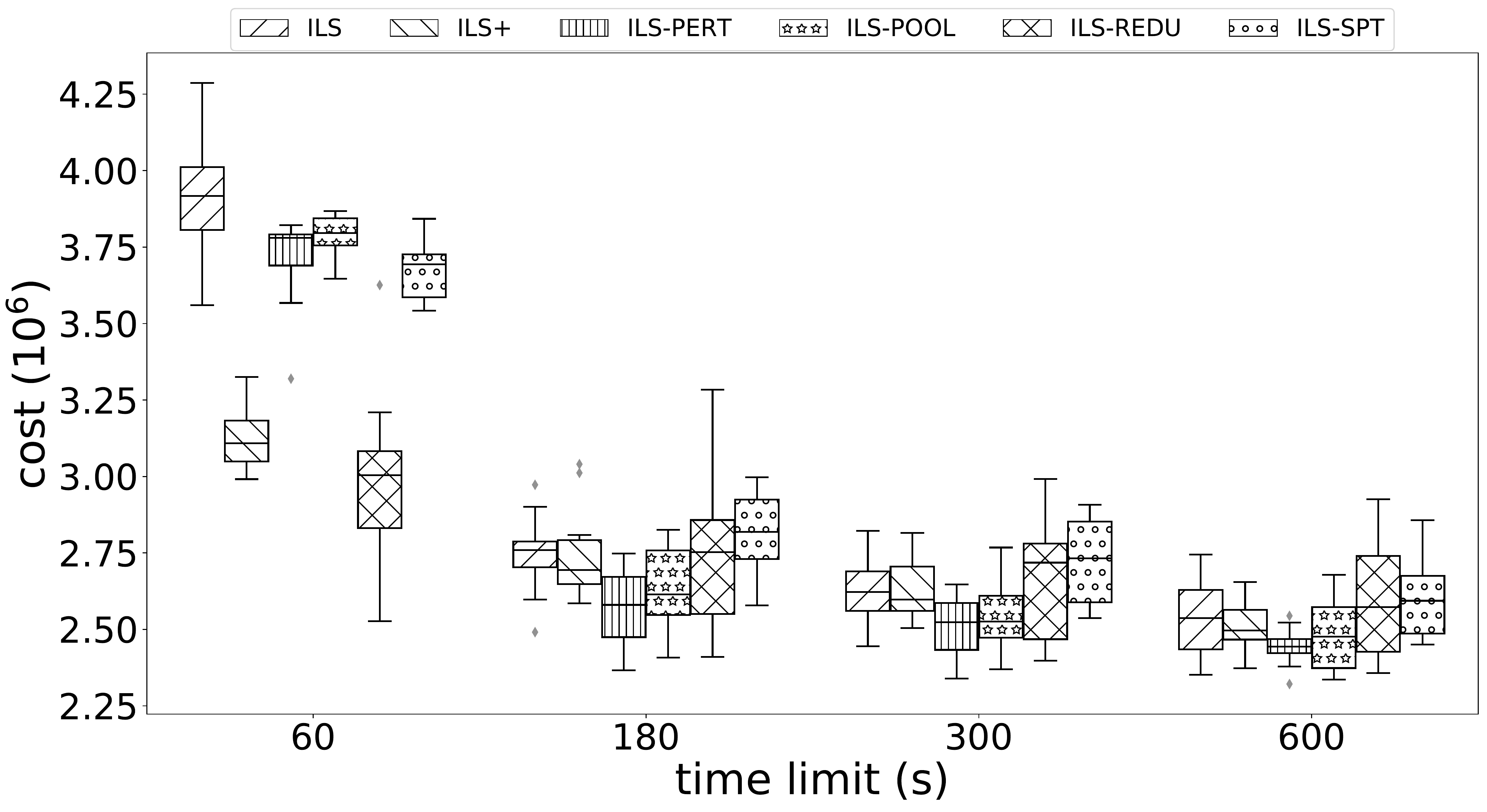}
\caption{HG-MP-24-4}
%\label{fig:LABEL}
\end{subfigure}
\hfill
\centering
\begin{subfigure}[b]{0.45\textwidth}
\centering
\includegraphics[width=\textwidth]{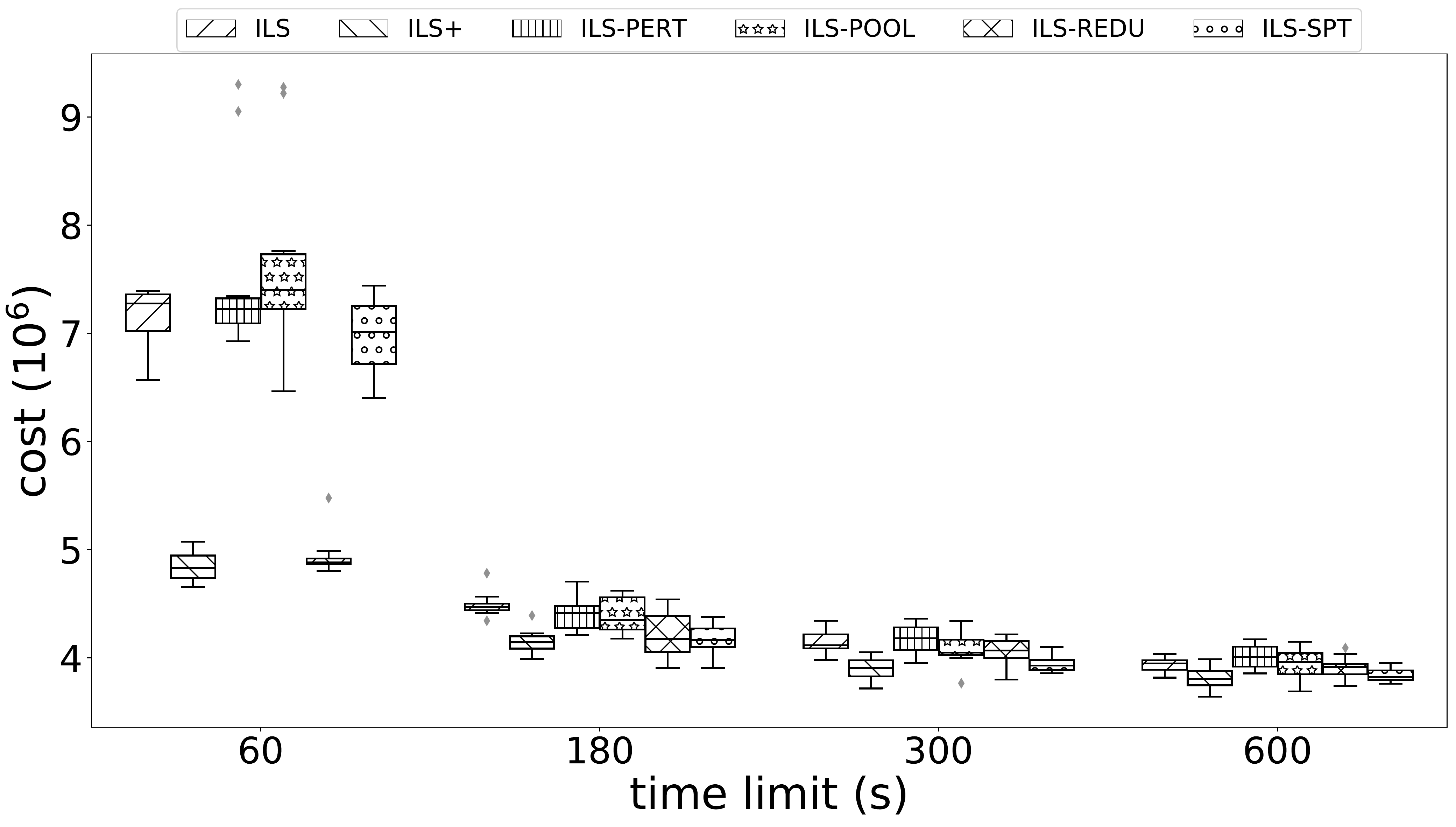}
\caption{HG-MP-26-1}
%\label{fig:LABEL}
\end{subfigure}
\hfill
\centering
\begin{subfigure}[b]{0.45\textwidth}
\centering
\includegraphics[width=\textwidth]{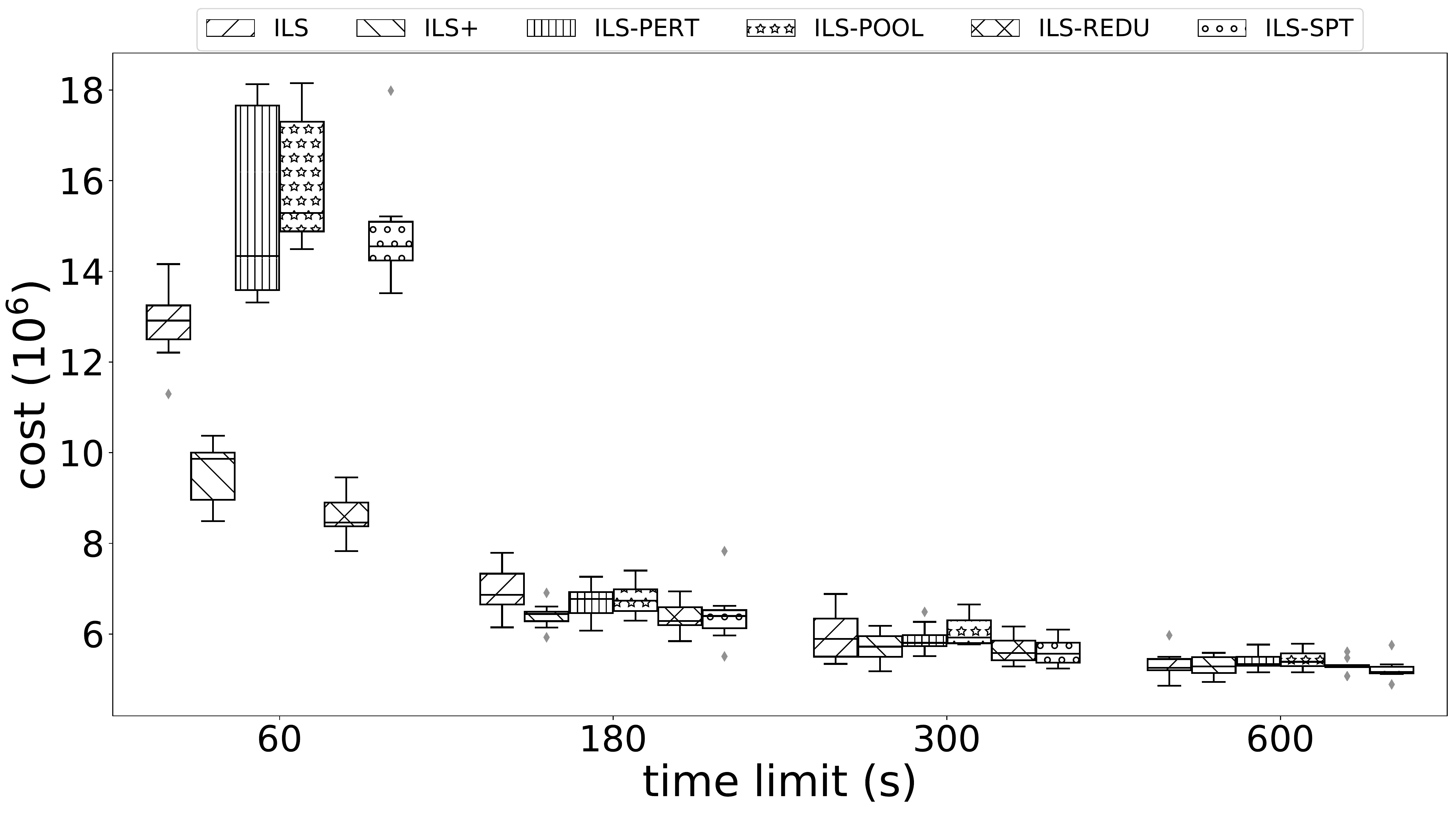}
\caption{HG-MP-28-1}
%\label{fig:LABEL}
\end{subfigure}
\hfill
\centering
\begin{subfigure}[b]{0.45\textwidth}
\centering
\includegraphics[width=\textwidth]{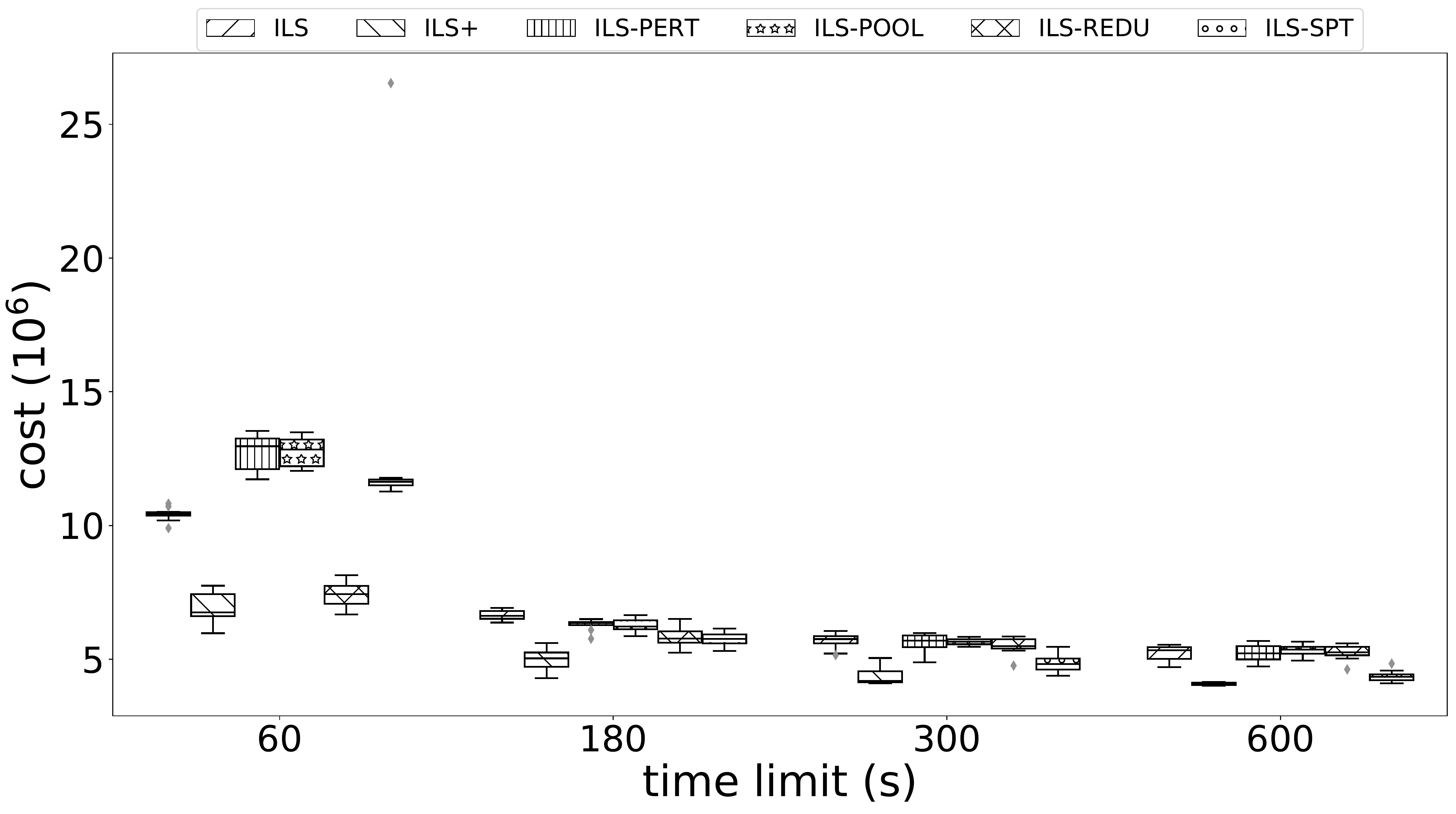}
\caption{HG-MP-30-3}
%\label{fig:LABEL}
\end{subfigure}
\hfill

    \caption{\rafaelC{Impact of the individual novelties on the obtained solution costs considering instances with more than 500 pipes.}}
    \label{fig:new_tests2}
\end{figure}

Figures~\ref{fig:new_tests1}-\ref{fig:new_tests2} summarize the results. The values composing the boxplots for each instance and time limit represent ten runs of the corresponding approach. 
It is noteworthy that the enhancements achieved by the individual novelties on the obtained solutions are very diverse. ILS-PERT was able to reach a good performance in certain cases, and that behavior is possibly related to its ability to overcome local minima solutions encountered by the algorithm. 
ILS-REDU \rafaelC{seldomly} allows considerable gains over ILS, especially when the allowed running time is increased. This is probably due to the nature of ILS-REDU, as it attempts to improve the convergence of the algorithm, quickly reaching hard-to-escape-from local minima solutions. It can allow, however, much faster convergence in some situations.
ILS-POOL has a good performance, especially when considering instances with less than 500 pipes, showing the benefits of diversification. This is most likely related to the fact that it increases the possibility of exploring more diverse solutions. 
ILS-SPT has a great performance and seems to be the individual novelty that allows the most considerable reduction in the costs of the obtained solutions, corroborating our claim that the pattern compelled by the shortest path-based local search indeed allows reaching good quality solutions.

Most importantly, the figures allow us to observe that ILS+ presented the best overall performance, showing that the combination of the individual novelties is indeed valuable. Together they \rafaelC{achieve more robustness, by allowing to obtain smaller deviations from the best known solutions in most cases,} and reduce the possible deficiency that each of them could present individually.
\rafaelC{These observations are further confirmed in Table~\ref{tab:summarydeviations}.
The table presents, for each approach and each time limit, separated by the sizes of the instances, the average deviation (in percent) from the best solution, calculated for each execution of each approach for each instance as $100 \times \frac{z_{app}-z_{best}}{z_{best}}$. In this formula, $z_{app}$ gives the solution value obtained by the approach and $z_{best}$ represents the best solution encountered for that instance by any of the approaches for the specified time limit. The best value in each column is highlighted in bold.
}

\begin{table}[!htp]\centering
\caption{\rafaelC{Average deviation in percent from the best obtained solutions for each time limit, separated by the sizes of the instances.}}
\scriptsize
\begin{tabular}{l|rrrr|rrrrr}\hline
&\multicolumn{4}{c|}{Less than 500 pipes} &\multicolumn{4}{c}{More than 500 pipes} \\
&60s &180s &300s &600s &60s &180s &300s &600s \\\hline
ILS &12.7 &11.2 &11.0 &10.7 &52.3 &22.8 &16.4 &12.4 \\
ILS+ &\textbf{6.2} &\textbf{4.9} &\textbf{4.5} & \textbf{4.2} & \textbf{10.3} & \textbf{9.0} & \textbf{6.1} & \textbf{4.5} \\
ILS-PERT &9.7 &8.6 &8.2 &7.7 &59.4 &17.9 &14.2 &11.7 \\
ILS-POOL &13.2 &11.0 &10.3 &9.8 &61.3 &19.0 &14.1 &11.6 \\
ILS-REDU &13.9 &13.1 &12.9 &12.7 &20.0 &19.1 &16.3 &14.5 \\
ILS-SPT &9.3 &8.7 &8.3 &7.8 &54.1 &12.7 &8.6 &6.4 \\
\hline
\end{tabular}
\label{tab:summarydeviations}
\end{table}

\subsection{Analyzing additional indicators for ILS and ILS+}
\label{sec:experimentsadditionalindicators}

In this section, we analyze the behavior of ILS and ILS+  when considering indicators such as the number of iterations, the number of calls to the hydraulic simulator, and the number of tested solutions.
We consider the same set of instances used in Subsection~\ref{sec:performanceindividualimprovements}. Besides, the settings are defined as in Subsection~\ref{sec:test}.

Table~\ref{tab:individual} summarizes the results. 
The first two columns indicate the instances and the allowed running times.
For each combination of instance and allowed running time, ten executions of each approach were performed.
The next columns report, for ILS and ILS+, the average number of iterations (i.e., number of executions of the while loop of lines~\ref{algo:ils:while:init}-\ref{algo:ils:while:end} in Algorithm~\ref{algo:ils}), the average number of calls to the hydraulic simulator, and the amount of tested solutions (including the average percentage of feasible solutions).
Table~\ref{tab:individual} shows that both ILS and ILS+ present reasonably high numbers of iterations, hydraulic simulation calls, and tested solutions, and these values tend to decrease as the size of the instances increase, as expected. This indicates that the simulations do not seem to be a real bottleneck, and the main challenge lies in the search for good quality solutions.
The table also shows that ILS+ consistently achieved higher values for all indicators but the \rafaelC{feasibility rate}. A possible explanation for \rafaelC{this} behavior is the faster convergence of ILS+\rafaelC{, which allows the approach to explore lower cost solutions in more restrained neighborhoods near local optima solutions, where infeasibilities are easier to obtain}. 
We remark that\rafaelC{, overall, the} indicators suggest that ILS+ presents a better efficiency in the search for good quality solutions when compared to ILS.

\begin{table}[H]\centering
\caption{Additional indicators for analyzing the behavior of ILS and ILS+.}\label{tab:individual}
%\scriptsize
\scriptsize
%\resizebox{\textwidth}{!}{%
%\begin{tabular}{lp{1cm}|p{1.2cm}p{1cm}p{1.9cm}|p{1.2cm}p{1cm}p{1.9cm}}\hline
% & &\multicolumn{3}{c}{ILS} &\multicolumn{3}{|c}{ILS+} \\\Tstrut
%Instance &Time (s) & \#iter ($10^3$) & \#hydr ($10^6$) & \#testedsol ($10^3$) \{Feas. rate (\%)\} &\#iter ($10^3$) & \#hydr ($10^6$) & \#testedsol %($10^3$) \{Feas. rate (\%)\} \Tstrut\\\hline
\begin{tabular}{lc|ccc|ccc}\hline
 & &\multicolumn{3}{c}{ILS} &\multicolumn{3}{|c}{ILS+} \\\Tstrut
Instance &Time & \#iter & \#hydr & \#testedsol ($10^3$) &\#iter & \#hydr & \#testedsol ($10^3$) \\
 & (s) & ($10^3$) & ($10^6$) & \{Feas. rate (\%)\} & ($10^3$) &  ($10^6$) & \{Feas. rate (\%)\} \\\hline
HG-MP-2-1 &60 &1.98 &1.03 &93.58 \{21.05\} &2.94 &1.43 &130.67 \{14.01\} \Tstrut\\
&180 &6.05 &3.08 &281.25 \{20.62\} &8.89 &4.26 &390.09 \{13.87\} \\
&300 &10.13 &5.12 &469.44 \{20.44\} &14.88 &7.08 &649.22 \{13.91\} \\
&600 &20.36 &10.23 &940.06 \{20.30\} &30.08 &14.11 &1301.81 \{13.99\} \\
\hline
HG-MP-4-3 &60 &0.48 &0.55 &45.17 \{25.52\} &0.86 &0.73 &67.01 \{16.24\} \Tstrut\\
&180 &1.54 &1.59 &135.79 \{23.47\} &2.71 &2.16 &204.17 \{15.58\} \\
&300 &2.60 &2.63 &226.54 \{23.19\} &4.58 &3.59 &341.91 \{15.48\} \\
&600 &5.26 &5.24 &453.77 \{22.95\} &9.27 &7.17 &687.18 \{15.40\} \\
\hline
HG-MP-6-5 &60 &0.48 &0.60 &55.61 \{19.55\} &0.74 &0.75 &82.88 \{9.93\} \Tstrut\\
&180 &1.51 &1.75 &168.40 \{17.77\} &2.30 &2.24 &251.07 \{9.37\} \\
&300 &2.54 &2.90 &281.20 \{17.43\} &3.87 &3.73 &419.44 \{9.28\} \\
&600 &5.12 &5.78 &563.18 \{17.20\} &7.86 &7.47 &839.93 \{9.31\} \\
\hline
HG-MP-8-5 &60 &0.19 &0.38 &30.44 \{28.05\} &0.41 &0.48 &45.95 \{16.93\} \Tstrut\\
&180 &0.66 &1.07 &92.29 \{23.73\} &1.28 &1.42 &140.89 \{15.11\} \\
&300 &1.13 &1.76 &154.01 \{22.86\} &2.13 &2.35 &235.53 \{14.66\} \\
&600 &2.31 &3.48 &309.22 \{22.13\} &4.26 &4.68 &472.42 \{14.27\} \\
\hline
HG-MP-10-4 &60 &0.09 &0.30 &20.42 \{33.53\} &0.16 &0.36 &26.46 \{20.31\} \Tstrut\\
&180 &0.35 &0.84 &59.75 \{28.09\} &0.67 &1.05 &81.18 \{18.81\} \\
&300 &0.62 &1.38 &99.01 \{27.00\} &1.19 &1.74 &136.39 \{18.63\} \\
&600 &1.30 &2.70 &197.49 \{26.17\} &2.51 &3.46 &275.76 \{18.53\} \\
\hline
HG-MP-12-4 &60 &0.10 &0.32 &21.52 \{33.00\} &0.16 &0.38 &29.40 \{16.40\} \Tstrut\\
&180 &0.34 &0.90 &64.42 \{25.78\} &0.56 &1.15 &90.33 \{13.88\} \\
&300 &0.60 &1.47 &107.36 \{24.35\} &0.97 &1.91 &151.51 \{13.40\} \\
&600 &1.24 &2.88 &214.53 \{23.23\} &1.98 &3.81 &305.25 \{13.05\} \\
\hline
HG-MP-14-5 &60 &0.05 &0.23 &15.85 \{39.98\} &0.10 &0.28 &24.27 \{20.60\} \Tstrut\\
&180 &0.21 &0.63 &47.56 \{30.75\} &0.41 &0.84 &72.70 \{18.15\} \\
&300 &0.38 &1.02 &79.10 \{28.74\} &0.73 &1.40 &121.74 \{17.74\} \\
&600 &0.81 &1.99 &157.66 \{27.52\} &1.59 &2.80 &245.95 \{17.57\} \\
\hline
HG-MP-16-1 &60 &0.02 &0.19 &12.48 \{40.48\} &0.05 &0.22 &15.90 \{26.35\} \Tstrut\\
&180 &0.11 &0.51 &36.53 \{31.65\} &0.23 &0.64 &50.66 \{19.54\} \\
&300 &0.21 &0.81 &59.63 \{29.31\} &0.42 &1.06 &86.09 \{18.04\} \\
&600 &0.46 &1.55 &117.08 \{27.07\} &0.90 &2.10 &175.24 \{17.05\} \\
\hline
HG-MP-18-2 &60 &0.02 &0.22 &12.07 \{57.20\} &0.06 &0.24 &16.09 \{25.68\} \Tstrut\\
&180 &0.13 &0.57 &38.34 \{36.48\} &0.24 &0.73 &52.23 \{18.42\} \\
&300 &0.24 &0.93 &64.03 \{32.35\} &0.43 &1.21 &89.46 \{17.14\} \\
&600 &0.53 &1.80 &126.79 \{29.51\} &0.93 &2.41 &184.15 \{16.28\} \\
\hline
HG-MP-20-2 &60 &0.02 &0.19 &11.58 \{48.46\} &0.05 &0.20 &15.34 \{28.77\} \Tstrut\\
&180 &0.10 &0.49 &36.59 \{32.93\} &0.22 &0.59 &50.05 \{21.10\} \\
&300 &0.18 &0.79 &60.80 \{30.07\} &0.40 &0.97 &84.98 \{19.65\} \\
&600 &0.41 &1.51 &120.49 \{28.01\} &0.84 &1.93 &172.46 \{18.60\} \\
\hline
HG-MP-22-2 &60 &0.01 &0.15 &8.99 \{49.60\} &0.02 &0.15 &10.97 \{34.34\} \Tstrut\\
&180 &0.05 &0.39 &26.66 \{38.72\} &0.10 &0.46 &36.12 \{22.89\} \\
&300 &0.11 &0.61 &43.85 \{34.11\} &0.20 &0.76 &61.01 \{19.65\} \\
&600 &0.25 &1.16 &87.01 \{29.09\} &0.42 &1.53 &122.73 \{17.09\} \\
\hline
HG-MP-24-4 &60 &0.01 &0.19 &9.55 \{67.53\} &0.02 &0.18 &12.67 \{31.75\} \Tstrut\\
&180 &0.06 &0.47 &30.88 \{40.93\} &0.11 &0.53 &41.71 \{20.45\} \\
&300 &0.12 &0.74 &52.41 \{33.76\} &0.20 &0.88 &70.80 \{17.96\} \\
&600 &0.28 &1.43 &105.26 \{28.82\} &0.45 &1.77 &144.00 \{15.90\} \\
\hline
HG-MP-26-1 &60 &0.01 &0.15 &7.95 \{62.49\} &0.01 &0.15 &10.25 \{34.44\} \Tstrut\\
&180 &0.04 &0.40 &25.02 \{41.21\} &0.07 &0.45 &33.37 \{22.07\} \\
&300 &0.08 &0.63 &42.85 \{33.58\} &0.13 &0.75 &56.64 \{18.85\} \\
&600 &0.19 &1.20 &86.08 \{28.35\} &0.30 &1.49 &115.44 \{15.95\} \\
\hline
HG-MP-28-1 &60 &0.01 &0.14 &8.09 \{49.06\} &0.01 &0.14 &9.25 \{40.30\} \Tstrut\\
&180 &0.03 &0.36 &23.54 \{41.13\} &0.05 &0.39 &29.84 \{26.57\} \\
&300 &0.06 &0.57 &38.96 \{36.31\} &0.10 &0.64 &50.15 \{22.54\} \\
&600 &0.15 &1.05 &75.97 \{31.12\} &0.23 &1.26 &99.87 \{18.83\} \\
\hline
HG-MP-30-3 &60 &0.01 &0.15 &8.23 \{56.68\} &0.01 &0.14 &10.01 \{35.84\} \Tstrut\\
&180 &0.03 &0.39 &25.00 \{42.29\} &0.05 &0.42 &31.77 \{25.74\} \\
&300 &0.06 &0.61 &42.42 \{35.52\} &0.10 &0.71 &52.62 \{21.29\} \\
&600 &0.16 &1.16 &85.48 \{29.18\} &0.22 &1.43 &104.89 \{16.93\} \\
\hline
\end{tabular}
\end{table}

\section{Final comments} \label{sec:final}

In this paper, we addressed the water distribution network design optimization problem and proposed a new enhanced simulation-based iterated local search (ILS) metaheuristic. Grounded on the structure of the problem, the proposed simulation-based ILS presents four main novelties: a {local search strategy for smart dimensioning the} pipes in shortest paths between reservoirs and high demand nodes, an aggressive pipe diameter reduction scheme based on varying factors {to speed up convergence to good quality solutions}, a concentrated perturbation mechanism {to permit escaping from very restrained local optima regions}, and a pool of solutions {to allow a good balance between intensification and diversification}.

The performed computational experiments have shown that the novelties embedded in our newly proposed simulation-based ILS allow improved performances when compared to a state-of-the-art simulation-based ILS. Furthermore, the results showed that our new approach presents a much faster convergence to good quality solutions and more robustness, as it obtained low costs solutions when allowing stricter time limits and achieved less deviation from the encountered best solutions when multiple executions were performed. 
Our approach found improved best solutions for 64.0\% (384 out of 600) of the tests for the instance groups and improved average solutions for 77.8\% (467 out of 600) of them. Remarkable improvements were achieved for the larger instance groups, for which these values were 76.0\% (228 out of 300) for best solutions and 80.0\% (240 out of 300) for average solutions. Furthermore, the average gains for the larger instances were in the order of 8.4\% for best solutions and 9.3\% for average solutions.

\vspace{0.8cm}

{
\noindent \small 
\textbf{Acknowledgments:} Work of Willian C. S. Martinho was supported by a State of Bahia Research Foundation (FAPESB) scholarship.   Work of Rafael A. Melo was supported by Universidade Federal da Bahia; the Brazilian Ministry of Science, Technology, Innovation and Communication (MCTIC); the State of Bahia Research Foundation (FAPESB); and the Brazilian National Council for Scientific and Technological Development (CNPq).
The authors are thankful to the anonymous reviewers for the comments which helped to improve the quality of the paper.
}

\bibliographystyle{apacite}
\bibliography{WDND_ILS_R2}

\end{document}